\documentclass{article}
\usepackage{iclr2027_conference,times}

\usepackage[utf8]{inputenc}
\usepackage[T1]{fontenc}
\usepackage{hyperref}
\hypersetup{hidelinks,
  pdftitle={OmniVR: Audio-Video Conditional Generation for Archival Footage Restoration},
  pdfauthor={OmniVR Authors}}
\usepackage{url}
\usepackage{booktabs}
\usepackage{amsfonts}
\usepackage{amsmath}
\usepackage{amssymb}
\DeclareMathSizes{7.6}{7.6}{6}{5}
\usepackage{amsthm}
\theoremstyle{plain}
\newtheorem{ovtheorem}{Theorem}
\newtheorem{ovlemma}[ovtheorem]{Lemma}
\newtheorem{ovproposition}[ovtheorem]{Proposition}
\newtheorem{ovcorollary}[ovtheorem]{Corollary}
\theoremstyle{definition}
\newtheorem{ovassumption}[ovtheorem]{Assumption}
\theoremstyle{plain}
\usepackage{bm}
\usepackage{mathtools}
\usepackage{nicefrac}
\usepackage{microtype}
\usepackage{xcolor}
\usepackage{graphicx}
\usepackage{caption}
\usepackage{subcaption}
\usepackage{algorithm}
\usepackage{float} 
\usepackage{algorithmic}
\usepackage{multirow}
\usepackage{array}
\usepackage{tabularx}
\usepackage{adjustbox}
\usepackage{colortbl}
\usepackage{enumitem}
\usepackage{pifont}
\usepackage{marvosym} 
\usepackage{tikz}
\usetikzlibrary{calc}

\newcommand{\fref}[1]{Figure~\ref{#1}}
\newcommand{\tref}[1]{Table~\ref{#1}}
\newcommand{\eref}[1]{Eq.~\ref{#1}}
\newcommand{\sref}[1]{Sec.~\ref{#1}}
\newcommand{\R}{\mathbb{R}}
\newcolumntype{Y}{>{\centering\arraybackslash}X}

\definecolor{ovL}{HTML}{DB2777} 
\definecolor{ovR}{HTML}{6D28D9} 
\definecolor{oursrow}{HTML}{FFF1F5} 
\definecolor{stripbg}{HTML}{F7F8FC}
\definecolor{stripink}{HTML}{172033}
\definecolor{avblue}{HTML}{2563EB}
\definecolor{avteal}{HTML}{0F766E}

\newcommand{\ovTeaserDataIcon}{%
  \begin{tikzpicture}[baseline=-0.58ex,x=0.01\linewidth,y=0.01\linewidth,
    line cap=round,line join=round,scale=0.78]
      \path[fill=avteal!22,draw=avteal!70,line width=0.45pt,rounded corners=0.28mm]
        (-1.55,-1.15) rectangle (0.95,1.2);
      \path[fill=white,draw=avteal,line width=0.48pt,rounded corners=0.18mm]
        (-1.3,-0.85) rectangle (0.7,0.9);
      \fill[avteal] (-0.76,-0.46)--(-0.76,0.5)--(0.14,0.02)--cycle;
      \foreach \yy in {-0.75,0,0.75}{\fill[avteal] (-1.45,\yy) circle (0.065);}
      \path[fill=white,draw=avteal,line width=0.45pt]
        (0.28,-0.5)--(1.95,-0.5)--(1.95,0.95)--(1.38,1.5)--(0.28,1.5)--cycle;
      \path[fill=avteal!24,draw=avteal,line width=0.35pt]
        (1.38,1.5)--(1.38,0.95)--(1.95,0.95);
      \draw[avteal,line width=0.35pt] (0.62,0.68)--(1.1,0.68) (0.62,0.3)--(1.57,0.3);
      \foreach \xx/\hh in {0.55/0.25,0.92/0.55,1.29/0.85,1.66/0.42}{
        \draw[avteal,line width=0.55pt] (\xx,-1.55+\hh/2)--(\xx,-1.55-\hh/2);
      }
  \end{tikzpicture}%
}
\newcommand{\ovTeaserAdaptIcon}{%
  \begin{tikzpicture}[baseline=-0.58ex,x=0.01\linewidth,y=0.01\linewidth,
    line cap=round,line join=round,scale=0.72]
      \foreach \xx/\col in {-2.4/avblue,0/avblue!70!ovR,2.4/ovR}{
        \path[fill=\col!10,draw=\col,line width=0.45pt,rounded corners=0.25mm]
          (\xx-0.78,-0.55) rectangle +(1.56,1.25);
        \path[fill=\col!20] (\xx-0.78,-0.55)--(\xx-0.38,-0.18)
          --(\xx+1.18,-0.18)--(\xx+0.78,-0.55)--cycle;
      }
      \draw[avblue!80,line width=0.65pt,-stealth] (-1.45,0.05)--(-0.95,0.05);
      \draw[ovR!80,line width=0.65pt,-stealth] (0.95,0.05)--(1.45,0.05);
      \draw[ovR!70,line width=0.4pt] (-2.4,-0.84)--(2.4,-0.84);
  \end{tikzpicture}%
}
\newcommand{\ovTeaserFlashIcon}{%
  \begin{tikzpicture}[baseline=-0.58ex,x=0.01\linewidth,y=0.01\linewidth,
    line cap=round,line join=round,scale=0.72]
      \foreach \xx in {-0.8,0,0.8}{\draw[ovR!75,line width=0.48pt] (\xx,1)--(\xx,1.5) (\xx,-1)--(\xx,-1.5);}
      \foreach \yy in {-0.6,0.15,0.8}{\draw[ovR!75,line width=0.48pt] (-1.15,\yy)--(-1.65,\yy) (1.15,\yy)--(1.65,\yy);}
      \path[fill=ovR!18,draw=ovR,line width=0.5pt,rounded corners=0.3mm]
        (-1.15,-1) rectangle (1.15,1.12);
      \path[fill=white,draw=ovR!70,line width=0.35pt,rounded corners=0.16mm]
        (-0.85,-0.7) rectangle (0.85,0.82);
      \path[fill=ovR,rounded corners=0.06mm]
        (0.22,0.78)--(-0.6,-0.08)--(-0.05,-0.08)--(-0.24,-0.8)
        --(0.63,0.15)--(0.08,0.15)--cycle;
  \end{tikzpicture}%
}
\newcommand{\ovTeaserEvalIcon}{%
  \begin{tikzpicture}[baseline=-0.58ex,x=0.01\linewidth,y=0.01\linewidth,
    line cap=round,line join=round,scale=0.72]
      \path[fill=ovL!20,draw=ovL!70,line width=0.45pt,rounded corners=0.2mm]
        (-1.22,-1.6) rectangle (1.4,1.4);
      \path[fill=white,draw=ovL,line width=0.45pt,rounded corners=0.2mm]
        (-1.4,-1.4) rectangle (1.15,1.6);
      \path[fill=ovL!25,draw=ovL,line width=0.35pt,rounded corners=0.2mm]
        (-0.65,1.3) rectangle (0.4,1.85);
      \foreach \xx/\yy in {-0.77/0.65,0.27/0.65,-0.77/-0.55,0.27/-0.55}{
        \path[fill=ovL!7,draw=ovL!60,line width=0.25pt,rounded corners=0.1mm]
          (\xx-0.35,\yy-0.35) rectangle (\xx+0.35,\yy+0.35);
        \draw[ovL,line width=0.4pt] (\xx-0.21,\yy)--(\xx-0.04,\yy-0.15)--(\xx+0.22,\yy+0.17);
      }
  \end{tikzpicture}%
}

\newcommand{\teasertechstrip}{%
  \begin{tikzpicture}[x=0.01\linewidth,y=0.01\linewidth,
      line cap=round,line join=round,
      ovhead/.style={anchor=base,align=center,inner sep=0pt,text=stripink,
        font=\sffamily\bfseries\fontsize{7.8}{8.5}\selectfont},
      ovmetric/.style={anchor=base,align=center,inner sep=0pt,text=stripink,
        font=\bfseries\fontsize{7.6}{8.3}\selectfont},
      ovdetail/.style={anchor=base,align=center,inner sep=0pt,text=stripink!86,
        font=\fontsize{7.2}{8.0}\selectfont}]
    \useasboundingbox (0,0) rectangle (100,14.2);
    \foreach \xx/\col in {0/avteal,25.33/avblue,50.66/ovR,75.99/ovL}{
      \path[fill=\col!16,rounded corners=1.15mm]
        (\xx+0.30,0.20) rectangle +(23.7,13.42);
      \path[fill=\col!3,draw=\col!28,line width=0.38pt,rounded corners=1.15mm]
        (\xx,0.66) rectangle +(23.7,13.42);
      \draw[white,line width=0.5pt] (\xx+1.4,13.55)--(\xx+22.3,13.55);
    }

    \node[ovhead] (ovDataHead) at (11.85,10.72) {Data pipeline};
    \node[anchor=base east,inner sep=0pt,outer sep=0pt] at ([xshift=-2.4pt]ovDataHead.base west) {\ovTeaserDataIcon};
    \node[ovmetric] at (11.85,7.70) {100K high-quality shots};
    \node[ovdetail] at (11.85,5.25) {Detailed AV captions};
    \node[ovdetail] at (11.85,3.05) {Joint AV degradation};

    \node[ovhead] (ovAdaptHead) at (37.18,10.72) {Adaptation};
    \node[anchor=base east,inner sep=0pt,outer sep=0pt] at ([xshift=-2.4pt]ovAdaptHead.base west) {\ovTeaserAdaptIcon};
    \node[anchor=base,inner sep=0pt] at (37.18,7.70) {%
      \begin{tikzpicture}[baseline=0pt,x=0.01\linewidth,y=0.01\linewidth,
          line cap=round,line join=round]
        \node[ovmetric,anchor=base west,font=\bfseries\fontsize{6.8}{8.3}\selectfont]
          at (1.45,0) {T2AV};
        \node[ovmetric,anchor=base west,font=\bfseries\fontsize{6.8}{8.3}\selectfont]
          at (7.35,0) {TAV2AV};
        \node[ovmetric,anchor=base west,font=\bfseries\fontsize{6.8}{8.3}\selectfont]
          at (15.75,0) {AV2AV};
        \draw[avblue!85,line width=0.42pt,-stealth] (6.05,0.55)--(7.02,0.55);
        \draw[ovR!85,line width=0.42pt,-stealth] (14.48,0.55)--(15.46,0.55);
      \end{tikzpicture}%
    };
    \node[ovdetail] at (37.18,5.25) {Gated condition annealing};
    \node[ovdetail] at (37.18,3.05) {AV input; no user text};

    \node[ovhead] (ovFlashHead) at (62.51,10.72) {OmniVR-Flash};
    \node[anchor=base east,inner sep=0pt,outer sep=0pt] at ([xshift=-2.4pt]ovFlashHead.base west) {\ovTeaserFlashIcon};
    \node[ovmetric] at (62.51,7.70) {\textasciitilde38 / \textasciitilde18 fps (1K/2K)};
    \node[ovdetail] at (62.51,5.25) {MeanFlow + Turbo VAE};
    \node[ovdetail] at (62.51,3.05) {One step, single GPU};

    \node[ovhead] (ovEvalHead) at (87.84,10.72) {Evaluation};
    \node[anchor=base east,inner sep=0pt,outer sep=0pt] at ([xshift=-2.4pt]ovEvalHead.base west) {\ovTeaserEvalIcon};
    \node[ovmetric] at (87.84,7.70) {200 clips, four axes};
    \node[ovdetail] at (87.84,5.25) {OmniVRBench evaluation};
    \node[ovdetail] at (87.84,3.05) {Video, audio, time, sync};
  \end{tikzpicture}%
}

\newcommand{\preferencebar}[4]{%
  \begin{tikzpicture}[baseline=-0.55ex,x=0.017cm,y=0.11cm]
    \path[fill=black!7,rounded corners=0.55mm] (0,0) rectangle (100,1.3);
    \draw[black!26,line width=0.28pt] (50,-0.25) -- (50,1.55);
    \shade[left color=#2,right color=#3,rounded corners=0.55mm]
      (0,0) rectangle (#1,1.3);
    \fill[#3] (#1,0.65) circle[radius=0.55mm];
    \node[anchor=west,text=#4,font=\scriptsize\bfseries] at (105,0.65) {#1};
  \end{tikzpicture}%
}
\newcommand{\basepref}[1]{\preferencebar{#1}{black!24}{black!52}{stripink}}
\newcommand{\ourspref}[1]{\preferencebar{#1}{ovL}{ovR}{ovR}}

\title{%
  {\color{ovL}O}%
  {\color{ovL!80!ovR}m}%
  {\color{ovL!60!ovR}n}%
  {\color{ovL!40!ovR}i}%
  {\color{ovL!20!ovR}V}%
  {\color{ovR}R}%
  : Joint Audio-Video Conditional\\Generation for Archival Footage Restoration%
}

\author{
Xin Lu\textsuperscript{\dag},
Zihao Fan\textsuperscript{\dag},
Jie Huang\textsuperscript{\ddag},
Mingchen Zhong,
Hexin Zhang,
Xueyang Fu\textsuperscript{\Letter},
Zheng-Jun Zha\\[4pt]
University of Science and Technology of China\quad
{\normalfont\small\textsuperscript{\ddag}Project leader\quad\textsuperscript{\Letter}Corresponding author}
}

\iclrfinalcopy  

\begin{document}
\renewcommand{\headrulewidth}{0pt}
\lhead{Preprint}              

\maketitle
\addtocontents{toc}{\protect\setcounter{tocdepth}{-10}}  

\vspace{-0.4in}
\begin{figure}[H]
\centering
\begingroup
\setlength{\tabcolsep}{0pt}
\begin{tabular}{@{}p{0.294\linewidth}p{0.706\linewidth}@{}}
  {\rmfamily\bfseries\fontsize{7.6}{8.6}\selectfont\color{stripink}(a) Archival inputs} &
  {\rmfamily\bfseries\fontsize{7.6}{8.6}\selectfont\color{stripink}(b) Joint audio-video restoration}\\[0.5pt]
\end{tabular}
\endgroup
\includegraphics[width=\textwidth]{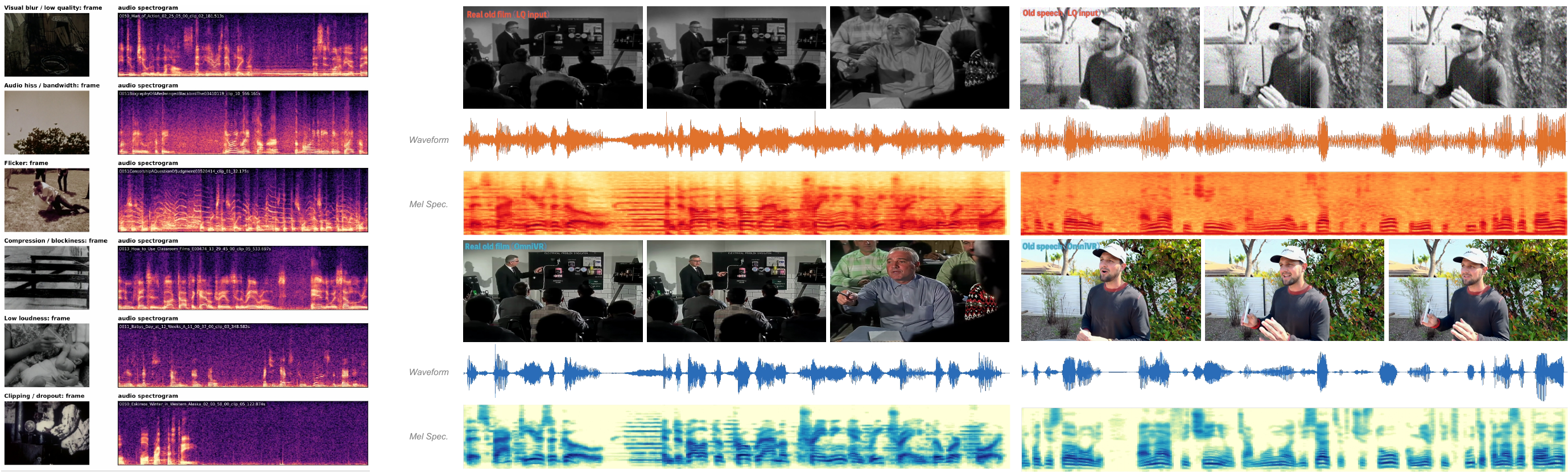}
\par\vspace{1pt}
\teasertechstrip
\captionsetup{skip=3pt}
\caption{\textbf{OmniVR: the first systematic framework for joint audio-video restoration.}
(a) Degraded archival footage and (b) paired restoration examples with video frames, waveforms, and mel spectrograms (AV denotes audio-video). The system connects AV captioning and paired degradation, T2AV-to-TAV2AV-to-AV2AV adaptation, one-step 1K/2K inference, and joint evaluation.
AV2AV requires no user text; a fixed restoration prompt is retained internally.}
\label{fig:teaser}
\end{figure}

\begin{abstract}
Archival footage often suffers from coupled visual and acoustic degradations, yet most restoration systems process the two modalities separately. To address this problem, we present \textbf{OmniVR}, the first systematic framework for joint audio-video restoration, covering data construction, model adaptation, efficient inference, and evaluation. We construct a high-quality audio-video corpus with detailed captions and use a joint degradation pipeline to produce aligned clean and degraded pairs. Using these pairs, we adapt a pretrained text-to-audio-video model (T2AV) by introducing degraded audio-video conditions (TAV2AV), then progressively replace sample captions with a fixed restoration prompt while retaining caption/null rehearsal. The resulting AV2AV model requires no user-provided text. Under a compatible residual-learning model, we prove that this condition-annealing schedule reduces gradient variance and expected restoration risk relative to direct fixed-prompt adaptation at the same training budget. For efficient deployment, \textbf{OmniVR-Flash} combines reduced-resolution video conditioning, MeanFlow-based one-step distillation, and Turbo VAE, achieving approximately 38 fps at 1K and 18 fps at 2K on a single B200 GPU. We further introduce \textbf{OmniVRBench} to evaluate four complementary dimensions: visual quality, audio quality, temporal consistency, and audio-visual synchrony. OmniVR achieves state-of-the-art results on public benchmarks and OmniVRBench. Data, code, and model weights will be released.
\paragraph{Project Page:} \url{https://xin1u.github.io/OminiVR_PAGE/} 
\end{abstract}


\section{Introduction}
\label{sec:intro}

\begin{figure*}[t]
\centering
\includegraphics[width=\textwidth]{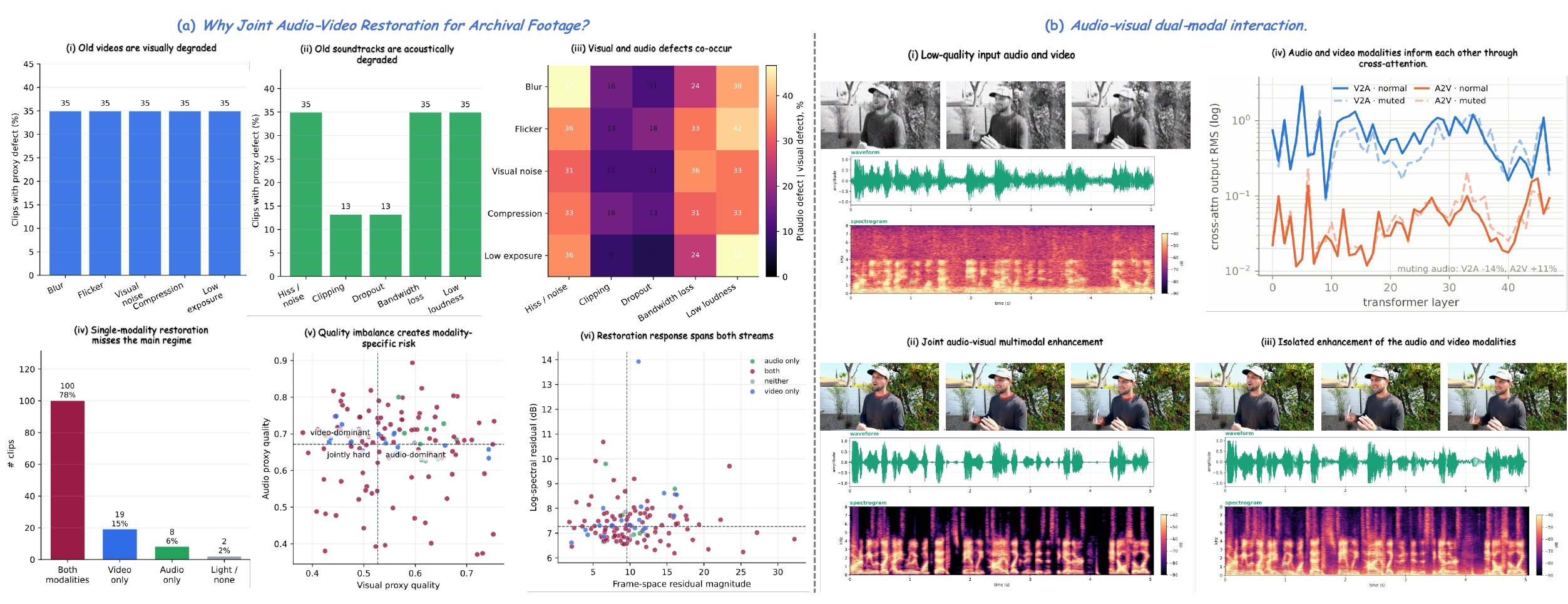}
\caption{\textbf{Why joint audio-video restoration?} \textbf{(a)} Diagnostics show pervasive visual and acoustic degradation (i, ii), cross-modal co-occurrence (iii), predominantly joint degradation (iv), quality imbalance (v), and restoration needs in both streams (vi). \textbf{(b)} Audio and video attend to each other (iv); muting audio alters the video response (i--iii), showing cross-modal interaction.}
\label{fig:motivation}
\end{figure*}

Archival footage is a joint record: visual defects such as blur, fading, scratches, and instability often occur together with hiss, clipping, and missing acoustic bands. Restoring it therefore requires recovering both streams while preserving the events, identities, and timing in the source. Existing visual restoration methods address archival-footage artifacts and temporal degradation~\citep{wan2022oldfilms,mao2025makingoldfilm}, while audio restoration systems recover speech and waveform detail with neural synthesis~\citep{kong2020hifigan,kong2021diffwave,liu2022voicefixer}. Most existing systems still optimize video and audio separately, so they cannot use visible articulation to resolve acoustic ambiguity or soundtrack evidence to constrain visual reconstruction. Our diagnostics show that the two degradations co-occur in collected material and that the audio and video pathways interact inside a joint generator; muting one modality changes the response of the other (\fref{fig:motivation}). Restoration is thus a coupled conditional-generation problem rather than two independent enhancement tasks.

A pretrained audio-video generator supplies a strong prior for this task, as demonstrated by recent joint audiovisual generation models~\citep{kondratyuk2023videopoet,polyak2024moviegen,hacohen2026ltx2}, but three constraints must be addressed together. Real clean/degraded archival pairs are scarce, so restoration supervision must be synthesized from aligned audio-video material. The generator is trained to follow captions, whereas restoration should be controlled by the observed low-quality pair and should not require a new prompt at inference. Meanwhile, repeated denoising and high-resolution encoding and decoding make a large joint model costly on long recordings; few-step flow formulations target this bottleneck~\citep{liu2022rectifiedflow,geng2025meanflows}. These constraints couple the data interface, conditioning transition, and deployment path; addressing only one leaves the restoration pipeline incomplete.

We present \textbf{OmniVR}, the first systematic framework for joint audio-video restoration (\fref{fig:teaser}). Our data pipeline (\fref{fig:data_pipeline}) constructs a corpus of 100K high-quality, synchronized 3--20\,s audio-video shots from Internet recordings using shot segmentation and quality and safety filtering. Gemini combines each shot with audio, ASR, and metadata to produce a structured audiovisual caption. Online degradation yields aligned clean/LQ pairs with severity metadata; shared shot identifiers also support high-resolution refiner supervision. These records connect captioning, degraded conditions, and restoration targets.

Starting from the 22B-parameter LTX-2 backbone~\citep{hacohen2026ltx2}, we follow the staged path in \fref{fig:framework}: T2AV preserves the pretrained caption/null CFG interface, TAV2AV introduces degraded audio-video conditions, and AV2AV gradually replaces sample captions with a fixed restoration prompt. Independent 90\% retention masks for text and LQ conditions preserve caption/null rehearsal as gated annealing shifts control toward the observed pair. Inference requires no user text. Under a compatible residual-learning model, this schedule lowers gradient variance and strictly reduces expected restoration risk relative to direct fixed-prompt adaptation at a finite training budget; our analysis states the compatibility, orthogonality, and curvature assumptions. First-frame anchoring, degradation-aware weights, and waveform supervision support long-video restoration.

For efficient deployment, \textbf{OmniVR-Flash} combines reduced-resolution video conditioning, MeanFlow one-step distillation, and Turbo VAE (\fref{fig:framework}). These components reduce conditioning, denoising, and encoding/decoding costs while retaining the joint restoration interface. On a single B200 GPU, throughput reaches approximately 38\,fps at 1K and 18\,fps at 2K, exceeding real-time ($>$24\,fps) at 1K. \tref{tab:flash_timing} reports per-component wall time.

We evaluate OmniVR on the RTN benchmark~\citep{wan2022oldfilms} and \textbf{OmniVRBench}, covering visual and audio quality, temporal consistency, and audio-video synchrony across controlled degradations and real archival footage. OmniVR achieves the best reported no-reference visual scores on RTN and the strongest joint-quality results in the main OmniVRBench comparisons. On the real archival track, MUSIQ reaches $61.87$ versus $52.38$ for the strongest visual baseline, while DNSMOS improves from $2.04$ for the degraded input to $2.43$. The data, code, and model weights will be released.

\section{Related Work}
\label{sec:related}

\paragraph{Archival images, films, and video restoration.}
Restoring archival media combines inverse-problem reconstruction with the need to preserve historically meaningful content. Early image systems focused on scratches, fading, blur, and color loss in old photographs, while later methods introduced learned priors and temporal propagation for moving footage~\citep{wan2020oldphoto,wan2022oldfilms}. DeepRemaster treats restoration and color transfer as a temporally coherent process, and more recent work models spatially varying and time-varying artifacts in old films~\citep{iizuka2020deepremaster,mao2025makingoldfilm}. These systems established two principles that also matter for audiovisual restoration: degradations should be modeled as structured processes rather than independent pixel noise, and temporal consistency is as important as single-frame perceptual quality.

General video restoration has developed complementary architectures for propagation and alignment. EDVR aggregates aligned neighboring frames, BasicVSR++ uses recurrent propagation, and transformer-based models such as VRT and RVRT model temporal dependencies with attention~\citep{wang2019edvr,chan2022basicvsrpp,liang2022vrt,liang2022rvrt}. RealBasicVSR studies the gap between synthetic and real degradations, and VideoGigaGAN uses a generative adversarial prior to recover high-frequency details~\citep{chan2022realbasicvsr,xu2024videogigagan}. Colorization systems including DDColor, BiSTNet, and ColorMNet show that semantic priors can restore plausible chroma when observations are grayscale or severely faded~\citep{kang2023ddcolor,yang2024bistnet,yang2024colormnet}. However, these visual systems generally treat the soundtrack as external; they cannot use speech articulation, music rhythm, or sound events to disambiguate visual reconstruction.

\paragraph{Audio, speech, and waveform restoration.}
Audio restoration covers denoising, dereverberation, bandwidth extension, declipping, and recovery of missing harmonic structure. HiFi-GAN demonstrates efficient adversarial waveform generation, DiffWave formulates waveform synthesis with diffusion, and VoiceFixer combines a spectral restoration front end with a HiFi-GAN neural vocoder for heterogeneous corruptions, operating in a spectrogram analysis--resynthesis paradigm~\citep{kong2020hifigan,kong2021diffwave,liu2022voicefixer}. These methods are strong at modality-specific fidelity, but normally receive only audio and have no mechanism to enforce agreement with visible mouth motion, scene dynamics, or visual event timing. For archival recordings, this limitation matters because hiss, clipping, missing bands, and frame defects often arise from the same capture or transcoding process.

\paragraph{Cross-modal representation and audiovisual synchronization.}
A separate line of work learns shared representations or explicit mappings between speech and visual evidence. VisualVoice uses visual context to separate and synthesize speech, AV-HuBERT learns audiovisual speech representations, and Wav2Lip aligns generated lip motion with a target waveform~\citep{gao2021visualvoice,shi2022avhubert,prajwal2020wav2lip}. These models demonstrate that the two streams carry complementary information, but they are specialized for speech-centric understanding or lip synchronization rather than open-domain restoration. Broader audiovisual benchmarks evaluate quality and synchronization jointly, making clear that a system can produce plausible video and plausible audio while still violating cross-modal timing~\citep{yang2026avbench,zhou2026avsyncbench}. OmniVR uses the same interaction principle for restoration: audio and video are encoded as a joint condition and reconstructed together, while temporal and synchrony metrics are evaluated explicitly.

\paragraph{Audiovisual generation and multimodal priors.}
Large generative models provide a useful prior for recovering details absent from a degraded observation. Diffusion models and latent diffusion establish score-based reconstruction, rectified flow gives a deterministic transport view, and diffusion transformers scale this paradigm to long token sequences~\citep{ho2020denoising,rombach2022high,liu2022rectifiedflow,peebles2023dit}. In visual restoration, Real-ESRGAN learns from synthetic high-order corruptions, whereas StableSR, DiffBIR, and SUPIR adapt pretrained generative priors to restoration conditions~\citep{wang2021realesrgan,wang2023stablesr,lin2024diffbir,yu2024supir}. Their success also exposes a central restoration trade-off: an unconstrained prior can improve perceptual realism while changing the observed identity, geometry, or event, so the condition must carry enough evidence to prevent hallucination.

Recent foundation models extend these ideas from images or silent video to joint audio-video generation. VideoPoet and Movie Gen model synchronized visual and acoustic content at scale, while LTX-2 provides a large joint audiovisual backbone with a text-conditioned interface~\citep{kondratyuk2023videopoet,polyak2024moviegen,hacohen2026ltx2}. These models are trained primarily for synthesis from captions; restoration instead requires fidelity to a particular observed pair and eventually operation without a user-supplied caption. OmniVR therefore keeps the pretrained caption/null interface during the initial T2AV stage, introduces degraded audio-video conditions in TAV2AV, and anneals toward an AV2AV interface whose fixed restoration prompt is internal. This staged transition connects the synthesis prior to a condition-faithful restoration objective.

\paragraph{Conditioning, flow matching, and efficient inference.}
Conditioning strategies determine whether a generative model follows semantic intent, observed evidence, or both. Text conditioning provides a broad semantic prior, while low-quality audiovisual conditioning supplies instance-specific information; using only one source can either underconstrain reconstruction or discard the pretrained generative capability. Our condition-annealing schedule is specialized to the T2AV-to-TAV2AV-to-AV2AV transition and is accompanied by a finite-budget risk analysis. It retains caption/null rehearsal while progressively increasing reliance on the degraded pair, so the endpoint can use an internal fixed prompt without requiring new text from the user.

For deployment, flow matching represents generation as integration of a learned velocity field, and rectified-flow formulations reduce the number of solver evaluations~\citep{liu2022rectifiedflow}. MeanFlow models an interval-averaged velocity and supports one-step prediction, which makes it suitable for distilling a multi-step restoration model~\citep{geng2025meanflows}. In parallel, lower-resolution condition encoding and efficient VAE modules reduce the cost of processing long high-resolution clips. OmniVR-Flash combines these ideas: the restoration model is distilled with an interval objective, the video condition is encoded at reduced resolution, and Turbo VAE lowers latent encoding and decoding overhead while preserving high-resolution joint outputs.

\paragraph{Evaluation protocols and the position of OmniVR.}
Video-generation evaluation has expanded from frame fidelity to perceptual quality, motion stability, semantic alignment, and long-range consistency. DOVER, EvalCrafter, and VBench++ provide complementary no-reference and generation-oriented measurements~\citep{wu2023dover,liu2023evalcrafter,huang2024vbenchpp}. Audiovisual evaluation adds waveform quality and synchronization, while restoration requires a further distinction between perceptual plausibility and source fidelity. OmniVRBench therefore separates visual quality, audio quality, temporal consistency, and audio-visual synchrony, and includes both a controlled track with clean references and a real archival track without references. Taken together, the prior literature supplies strong modality-specific components and generative priors, but does not provide a complete open-domain system that constructs paired audiovisual supervision, transitions a text-to-audiovisual generator to an observation-only restoration interface, and accelerates the resulting model to real-time high-resolution inference. OmniVR is designed to fill this systems gap.

\section{Methodology}
\label{sec:method}

OmniVR constructs paired supervision, adapts a joint generator through condition annealing, and accelerates inference with OmniVR-Flash. Figures~\ref{fig:data_pipeline} and~\ref{fig:framework} summarize this pipeline.

\subsection{Data Construction and Joint Degradation}
\label{sec:degradation}

\begin{figure*}[t]
\centering
\includegraphics[width=\textwidth]{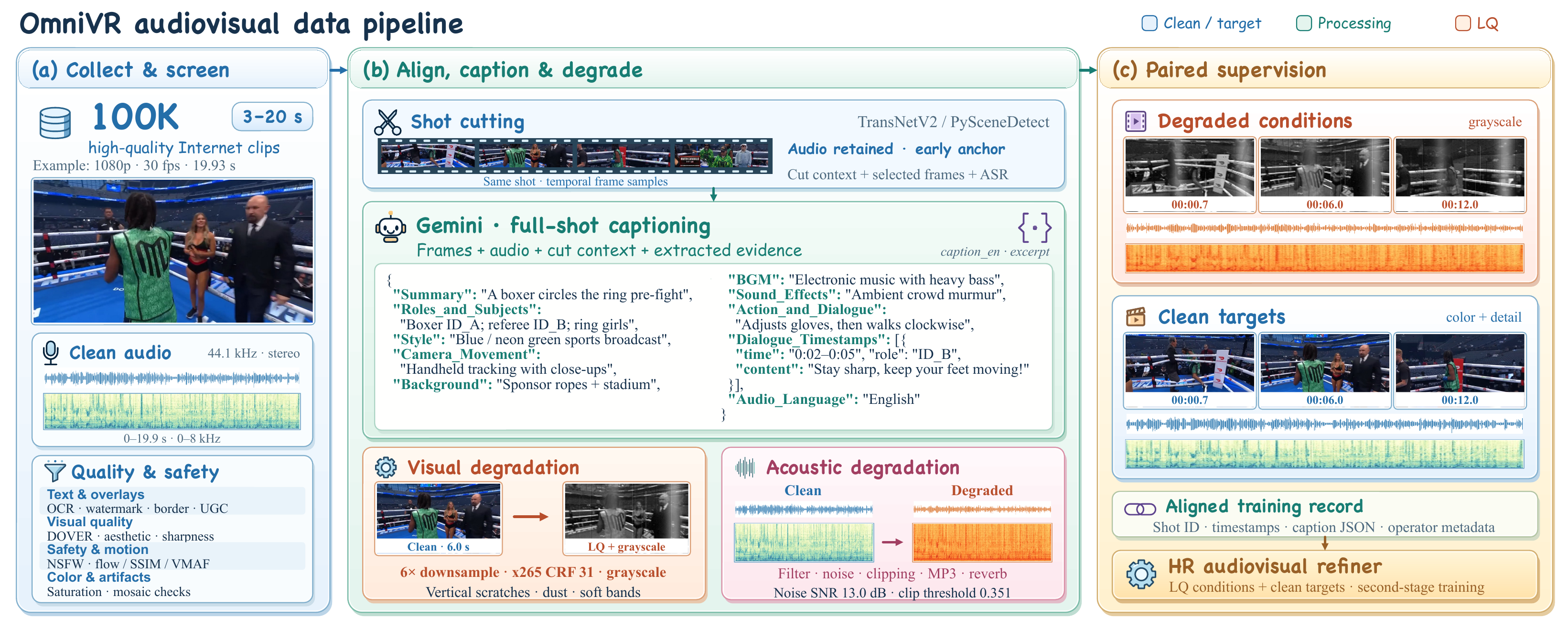}
\caption{\textbf{Shot-aligned audiovisual data pipeline.} Synchronized 3--20\,s shots pass the listed quality/safety operators; Gemini combines shot context, frames, and audio evidence into one structured caption. Caption JSON, severity metadata, and clean/LQ pairs remain aligned for TAV2AV training and high-resolution refiner supervision. The boxing example uses archived frames and measured waveforms/spectra; the degraded video is additionally grayscaled for visualization.}
\label{fig:data_pipeline}
\end{figure*}

\paragraph{Captioned training corpus.}
We construct a corpus of 100K synchronized 3--20\,s shots from high-quality Internet audio-video recordings using TransNetV2/PySceneDetect, retaining each shot's audio and a clear annotation frame from its first two seconds. After quality and safety filtering (\fref{fig:data_pipeline}), Gemini uses the shot, ASR evidence, and metadata to describe entities, actions, camera motion, dialogue, music, and sound effects. Each record contains clean video $\mathbf V^h$, audio $\mathbf A^h$, caption $y$, and shot metadata. The identifiers also support high-resolution refiner supervision; details appear in \sref{sec:s6}.

\paragraph{Paired degradation.}
An online operator synthesizes the degraded conditions from each clean pair:
\begin{equation}
    (\mathbf{V}^{l}, \mathbf{A}^{l}) = \mathcal{D}(\mathbf{V}^{h}, \mathbf{A}^{h}).
    \label{eq:degradation}
\end{equation}
Independent stochastic video and audio branches produce visual-dominant, audio-dominant, and jointly degraded samples while preserving alignment. The visual branch extends Real-ESRGAN-style high-order degradation~\citep{wang2021realesrgan} with temporally coherent flicker, scratches, dust, abrasion bands, and grayscale conversion. The audio branch models hiss, rumble, clicks, clipping, bandwidth loss, spectral over-smoothing, harmonic thinning, and phase discontinuities. Each record is $(y,\mathbf V^l,\mathbf A^l;\mathbf V^h,\mathbf A^h)$, with severity scalars $d_v,d_a\in[0,1]$ used in \eref{eq:loss_reweight}.

\subsection{Joint Restoration Model and Conditional Interface}
\label{sec:av2av}

\paragraph{Latent encoding and flow matching.}
We adapt the pretrained joint DiT to predict clean audio-video latents from degraded observations. For modality $m\in\{v,a\}$, let $\mathbf X_v=\mathbf V$ and $\mathbf X_a=\mathbf A$. Frozen VAE encoders $E_m$ and patchifiers $P_m$ produce target and condition tokens:
\begin{equation}
    \mathbf{x}_m=P_m\bigl(E_m(\mathbf X_m^h)\bigr),\qquad
    \mathbf{c}_m=P_m\bigl(E_m(\mathbf X_m^l)\bigr).
    \label{eq:encode_patchify}
\end{equation}
In the base model, $\mathbf x_m,\mathbf c_m\in\R^{N_m\times128}$. Video uses $N_v=T_zH_zW_z$ latent-grid tokens, where $T_z$, $H_z$, $W_z$ are the temporal depth, height, and width of the latent grid. Audio produces $N_a$ tokens by packing 8 channels and 16 mel bins into each 128-dimensional token. Rectified-flow training~\citep{liu2022rectifiedflow} uses a shared $\sigma\sim\mathcal U(0,1)$ and independent noise $\boldsymbol\epsilon_m\sim\mathcal N(\mathbf0,\mathbf I)$:
\begin{equation}
    \tilde{\mathbf{x}}_m=(1-\sigma)\mathbf{x}_m+\sigma\boldsymbol\epsilon_m,\qquad
    \mathbf{u}_m=\boldsymbol\epsilon_m-\mathbf{x}_m.
    \label{eq:flow_target}
\end{equation}

\paragraph{Channel-concat condition injection.}
We perturb the LQ tokens to discourage direct copying, then concatenate each with its target token before projection:
\begin{equation}
    \tilde{\mathbf c}_m=\mathbf c_m+\rho_m\boldsymbol\eta_m,
    \quad\boldsymbol\eta_m\sim\mathcal N(\mathbf0,\mathbf I),\qquad
    \mathbf h_{m,i}=W_m[\tilde{\mathbf x}_{m,i};\tilde{\mathbf c}_{m,i}]+\mathbf b_m.
    \label{eq:concat_projection}
\end{equation}
Here $i$ indexes tokens, $[\cdot;\cdot]$ denotes channel concatenation, $W_m\in\R^{d\times256}$ maps each concatenated token to the DiT hidden dimension $d$, and $\mathbf b_m$ is the corresponding bias. The first 128 columns of $W_m$ copy the pretrained projection and the new condition columns are zero-initialized, preserving the initial T2AV mapping. We sample $\rho_m\sim\mathcal U(0.4,0.6)$ during training and use $\rho=0.5$ at inference.

\paragraph{Joint prediction.}
The multimodal DiT with parameters $\theta$ predicts both velocity fields using the projected tokens, text embedding $\mathbf e$, and shared noise level:
\begin{equation}
    (\hat{\mathbf u}_v,\hat{\mathbf u}_a)
    =G_\theta(\mathbf h_v,\mathbf h_a,\mathbf e,\sigma).
    \label{eq:joint_transformer}
\end{equation}
Cross-modal interactions let each stream use evidence from the other. With full condition $\mathbf q=(\mathbf c_v,\mathbf c_a,\mathbf e)$, we transfer control from captions to the LQ pair while retaining the text pathway.

\subsection{Condition Annealing from T2AV to AV2AV Inference}
\label{sec:annealing}

\paragraph{Three-stage condition schedule.}
The transition starts from the pretrained T2AV caption/null interface, introduces LQ conditions during caption-assisted TAV2AV adaptation, and anneals captions to a fixed restoration prompt for AV2AV inference. Throughout adaptation, independent masks $A_k\sim\operatorname{Bernoulli}(p)$ and $T_k\sim\operatorname{Bernoulli}(t)$ retain the LQ pair and non-null text, respectively, with $p=t=0.9$. The LQ mask gates the new condition projection. Initially, the (caption, LQ), (caption, no-LQ), (null, LQ), and (null, no-LQ) regimes have probabilities $0.81$, $0.09$, $0.09$, and $0.01$.

We anneal non-null text only when LQ is present. Let $\mathbf e_y$, $\mathbf e_{\rm rest}$, and $\mathbf e_{\emptyset}$ encode the caption, fixed restoration prompt, and null text. An independent selector $M_k\sim\operatorname{Bernoulli}(\lambda_k)$ gives:
\begin{equation}
 \lambda_k=\max\!\left(1-\frac{k}{K_{\rm ann}},0\right),\qquad
 \mathbf e_k=\begin{cases}
 \mathbf e_{\emptyset}, & T_k=0,\\
 \mathbf e_{\rm rest}, & T_k=1,\ A_k=1,\ M_k=0,\\
 \mathbf e_y, & \text{otherwise}.
 \end{cases}
 \label{eq:dual_condition_annealing}
\end{equation}
Here $k$ indexes adaptation updates and $K_{\rm ann}$ spans the first 30\% of training. The 10\% no-LQ branch continues caption/null rehearsal throughout training. At $\lambda_k=0$, the LQ-present branch uses the fixed prompt with probability $0.9$ and null text with probability $0.1$. At inference, the user supplies only the degraded pair, and the model uses $\mathbf q_{\rm inf}=(\mathbf c_v,\mathbf c_a,\mathbf e_{\rm rest})$ internally.

\begin{figure*}[t]
\centering
\includegraphics[width=0.999\textwidth]{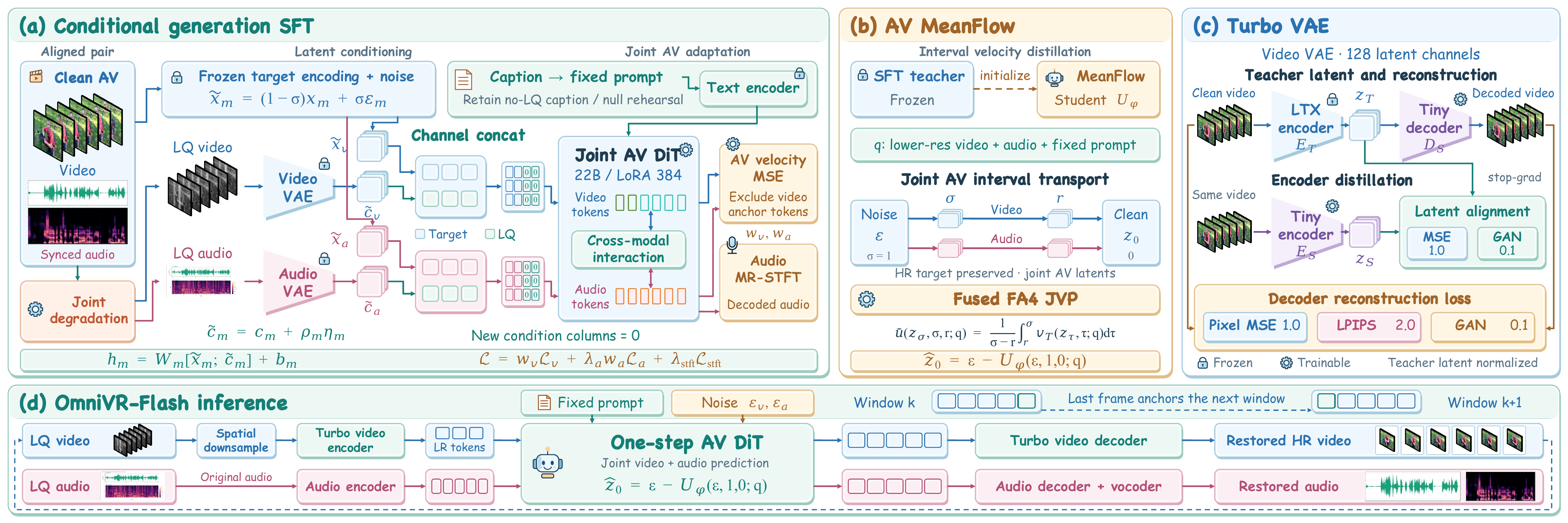}
\caption{
\textbf{OmniVR and OmniVR-Flash framework.} (a) Conditional SFT adapts a joint audiovisual DiT with degraded inputs and caption annealing to a fixed prompt. (b) AV MeanFlow enables one-step prediction. (c) Turbo VAE accelerates video encoding and decoding. (d) OmniVR-Flash downsamples LQ video before condition encoding and DiT injection, reducing conditioning cost while preserving high-resolution outputs. The last restored frame anchors the next window.
}
\label{fig:framework}
\end{figure*}

\paragraph{Training objective and finite-budget gain.}
Let $J_{\rm cap}^{0},J_{\emptyset}^{0}$ be expected no-LQ losses, and $J_{\rm cap}^{\rm LQ},J_{\rm rest}^{\rm LQ},J_{\emptyset}^{\rm LQ}$ their LQ-present counterparts. Each uses the severity-aware objective in \eref{eq:total_loss}. Define $P=tJ_{\rm cap}^{0}+(1-t)J_{\emptyset}^{0}$ for T2AV rehearsal, $C=tJ_{\rm cap}^{\rm LQ}+(1-t)J_{\emptyset}^{\rm LQ}$ for caption-assisted restoration, and $R=tJ_{\rm rest}^{\rm LQ}+(1-t)J_{\emptyset}^{\rm LQ}$ for fixed-prompt restoration. The exact expected loss is:
\begin{equation}
 F_\lambda=(1-p)P+p\bigl[\lambda C+(1-\lambda)R\bigr].
 \label{eq:dual_condition_objective}
\end{equation}
The degradation weights depend on observed severities, independently of the condition masks, so they preserve these mixture coefficients. The constant rehearsal term retains training pressure on the original text interface as the restoration objective shifts from $C$ to $R$.

Under the compatible scalar residual model of Proposition~\ref{prop:dual_risk} (under Assumption~\ref{ass:noise_ortho}, which makes the LQ feature independent of the residual variables and requires the noise to have zero conditional mean given the residual and its caption estimate), caption assistance lowers gradient variance at the same mean contraction rate. At fixed initialization, step size $0<\eta<2$, retention probabilities, and $N$ updates, the expected risk gain over direct fixed-prompt training is:
\begin{equation}
 \mathbb E\!\left[\mathcal R_R(w_N^{\rm dir})-\mathcal R_R(w_N^{\rm ann})\right]
 =\frac{pt\eta^2\kappa}{2}
   \sum_{k=0}^{N-1}\lambda_k\rho^{N-1-k}\ge0,
 \qquad \rho=1-p\eta(2-\eta).
 \label{eq:dual_condition_main_gain}
\end{equation}
Here $w_N$ is the learned scalar LQ coefficient, $\mathcal R_R$ is its fixed-prompt restoration risk, $\kappa>0$ measures caption compatibility with residual variation orthogonal to the LQ feature, and $\rho$ denotes the per-step contraction rate (distinct from the condition noise scale $\rho_m$ in \eref{eq:concat_projection}). For $p=t=0.9$, the gain is strict when any $\lambda_k>0$. \sref{sec:s7} gives the assumptions, the severity-weighted extension, and the separate comparison with an always-LQ baseline. It also establishes local prior-retention and stationary-path tracking bounds under explicit curvature assumptions; these bounds describe the stated residual and gradient-descent models.

\paragraph{AV2AV guidance.}
At inference, classifier-free guidance compares the fixed restoration prompt with null text while retaining the same LQ pair $\mathbf c=(\mathbf c_v,\mathbf c_a)$ in both branches. Writing $\mathbf v_\theta$ for the velocity field of $G_\theta$:
\begin{equation}
 \mathbf v_{\rm cfg}(\mathbf e;\mathbf c)
 =\mathbf v_\theta(\mathbf e_{\emptyset};\mathbf c)
 +w\left[\mathbf v_\theta(\mathbf e;\mathbf c)-
             \mathbf v_\theta(\mathbf e_{\emptyset};\mathbf c)\right],
 \quad \mathbf e=\mathbf e_{\rm rest}.
 \label{eq:dual_cfg}
\end{equation}
Here $w$ is the guidance scale and the shared latent and noise-level arguments are omitted. This AV2AV interface requires no user-provided text. The no-LQ rehearsal branch retains the caption/null interface for T2AV generation.

\begin{figure*}[t]
\centering
\includegraphics[width=\textwidth]{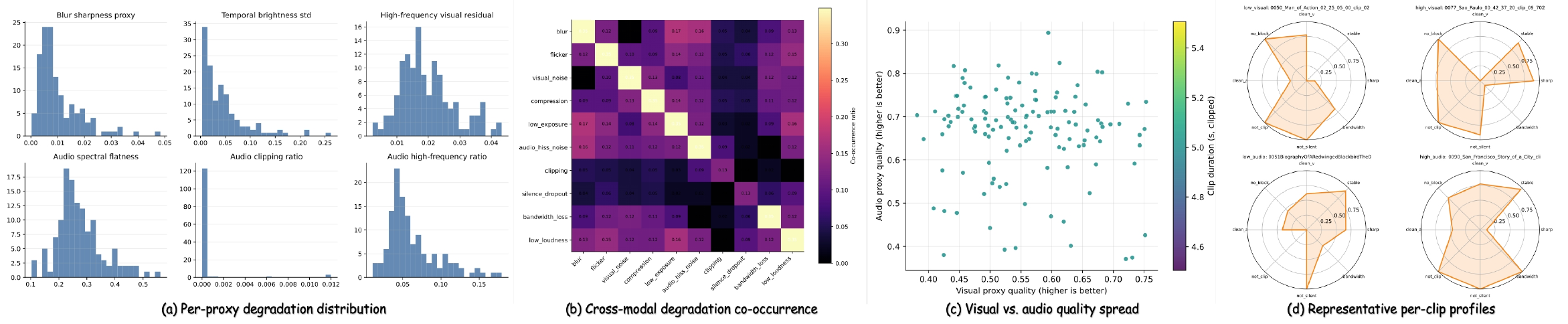}
\caption{\textbf{OmniVRBench degradation characterization.} \textbf{(a)} Distributions of three visual and three audio degradation proxies. \textbf{(b)} Co-occurrence of ten defect tags: $77.5\%$ of clips have both visual and acoustic degradation. \textbf{(c)} Weakly correlated visual/audio quality scores motivate joint evaluation. \textbf{(d)} Eight-axis profiles for four clips with low/high visual or audio quality illustrate degradation diversity.}
\label{fig:omnibench_diag}
\end{figure*}

\subsection{Temporal Consistency and Optimization}
\label{sec:i2av}
The conditional model is trained with a boundary anchor for long-video continuity and multimodal losses that account for uneven degradation severity.

\paragraph{First-frame I2V conditioning.}
OmniVR processes 121-frame windows. Training enables first-frame conditioning with probability $p_{\rm ff}=0.5$. Let $\mathbf M\in\{0,1\}^{N_v}$ select the first $H_zW_z$ video tokens for anchored samples and be zero otherwise. We replace selected tokens with clean targets ($\odot$ denotes element-wise multiplication):
\begin{equation}
    \tilde{\mathbf x}_v\leftarrow
    \mathbf M\odot\mathbf x_v+(1-\mathbf M)\odot\tilde{\mathbf x}_v.
    \label{eq:first_frame_replace}
\end{equation}
Anchor tokens are excluded from the loss. The first inference window runs without anchoring, a mode sampled with probability $1-p_{\rm ff}=0.5$ during training. Subsequent windows use the previous window's restored last frame, enabling bounded-memory processing. \sref{sec:s8} analyzes the shift from clean training anchors to generated inference anchors.

\paragraph{Degradation-aware velocity losses.}
The video loss averages over non-anchor tokens, while the audio loss uses all $N_a$ tokens:
\par\noindent
\begin{minipage}[b]{0.54\linewidth}
\begin{equation}
    \mathcal L_v=
    \frac{\sum_i(1-\mathbf M_i)\|\hat{\mathbf u}_{v,i}-\mathbf u_{v,i}\|_2^2}
    {\sum_i(1-\mathbf M_i)},
    \label{eq:video_loss}
\end{equation}
\end{minipage}\hfill
\begin{minipage}[b]{0.44\linewidth}
\begin{equation}
    \mathcal L_a=\frac1{N_a}\sum_i\|\hat{\mathbf u}_{a,i}-\mathbf u_{a,i}\|_2^2.
    \label{eq:audio_loss}
\end{equation}
\end{minipage}\par
For each sample, $d_v,d_a\in[0,1]$ average the normalized degradation parameters within each modality. We assign modality weights:
\begin{equation}
    w_v=\mathrm{Norm}_B(1+\alpha_vd_v+\beta_vd_a+\gamma_vd_vd_a),\qquad
    w_a=\mathrm{Norm}_B(1+\alpha_ad_a+\beta_ad_v+\gamma_ad_vd_a),
    \label{eq:loss_reweight}
\end{equation}
where $\mathrm{Norm}_B$ normalizes each modality's weights to unit mean over the 64-sample global batch using an all-reduce. We use $\alpha_v=\alpha_a=1.0$, $\beta_v=\beta_a=0.5$, and $\gamma_v=\gamma_a=2.0$; the interaction term emphasizes jointly severe degradation. Global normalization preserves sample-dependent weights despite one-sample GPU micro-batches. Bounds and observed weight ranges appear in \sref{app:implementation}.

\paragraph{Waveform supervision and total loss.}
A random audio sub-batch is decoded through the frozen audio VAE and vocoder. Multi-resolution STFT loss $\mathcal L_{\rm stft}$ compares the reconstructed and clean waveforms using window sizes 512, 1024, and 2048. We combine the modality-weighted velocity losses with this spectral term:
\begin{equation}
    \mathcal L=w_v\mathcal L_v+\lambda_a w_a\mathcal L_a
    +\lambda_{\rm stft}\mathcal L_{\rm stft},
    \label{eq:total_loss}
\end{equation}
with $\lambda_a=1.0$ and $\lambda_{\rm stft}=3\times10^{-3}$. The spectral term supervises decoded acoustic detail and remains additive, without multiplication by $w_a$. This objective covers all condition regimes in \eref{eq:dual_condition_objective}.

\subsection{Efficient Deployment with OmniVR-Flash}
\label{sec:flash}
OmniVR-Flash combines reduced-resolution video conditioning, AV MeanFlow distillation, and Turbo VAE (\fref{fig:framework}). It retains the fixed-prompt AV2AV interface and reaches approximately 38\,fps at 1K and 18\,fps at 2K on one B200 GPU.

\paragraph{Reduced-resolution video conditioning.}
We downsample and encode LQ video, then inject the condition through a single learned DiT layer. The target retains its high-resolution grid and audio conditioning is unchanged. Condition resolution trades source detail for efficiency.

\paragraph{AV MeanFlow distillation.}
Using the conditional restoration model as a frozen teacher, we train a student to predict finite-interval transport of the joint audio-video latent. Let $\mathbf z_\sigma$ denote the joint latent at noise level $\sigma$ (corresponding to $\tilde{\mathbf x}_m$ in \eref{eq:flow_target}). For fixed condition $\mathbf q$, the MeanFlow average velocity~\citep{geng2025meanflows} over $[r,\sigma]$ is:
\begin{equation}
 \overline{\mathbf u}(\mathbf z_\sigma,\sigma,r;\mathbf q)
 =\frac1{\sigma-r}\int_r^\sigma
 \mathbf v_{\rm teacher}(\mathbf z_\tau,\tau;\mathbf q)\,d\tau,
 \qquad 0\le r<\sigma\le1.
 \label{eq:flash_meanflow}
\end{equation}
Here $\mathbf v_{\rm teacher}$ denotes the frozen teacher's classifier-free guided velocity at scale $3.0$, as specified in Algorithm~\ref{alg:ov_imf}. The student $\mathbf U_\phi$, parameterized by $\phi$, learns average velocity with the improved MeanFlow (iMF) velocity objective~\citep{imeanflow}, with a 3D adversarial loss on decoded one-step video outputs (\sref{app:ov_imf}). We compute the Jacobian--vector product with a fused FlashAttention-4 (FA4) kernel~\citep{MLSYS2026_ae8b0b58}. With the data-to-noise convention in \eref{eq:flow_target}, both modalities take one step to yield the predicted clean latent $\widehat{\mathbf z}_0$:
\begin{equation}
 \widehat{\mathbf z}_0=\boldsymbol\epsilon-
 \mathbf U_\phi(\boldsymbol\epsilon,1,0;\mathbf q),
 \label{eq:flash_onestep}
\end{equation}
where $\boldsymbol\epsilon$ is the initial audio-video noise and $\mathbf q$ contains the LQ pair and internal restoration prompt.

\paragraph{Turbo VAE.}
Following Ultra Flash's training procedure and code~\citep{luxury2026ultraflash}, a lightweight encoder aligns with frozen teacher latents, and a lightweight decoder learns from teacher latents using pixel, perceptual, and adversarial losses. Both retain the teacher's 128-channel video latent interface. At inference, they encode the downsampled LQ video and decode the high-resolution output; the audio VAE and vocoder are unchanged.

\subsection{Implementation and Inference Settings}
\label{sec:impl}
We adapt the 22B LTX-2 joint DiT~\citep{hacohen2026ltx2} with rank-384 LoRA~\citep{hu2021lora}, keeping the VAEs, text encoder, and vocoder frozen. Base training uses 64 B200 GPUs for approximately five days. Reported base-model results use the 50K-step checkpoint, 15 Euler steps, CFG 3.0, and condition noise $\rho=0.5$. Training schedules, checkpoint identifiers, optimizer settings, and Flash architecture are specified in \sref{app:implementation}.

\begin{table}[t]
\centering
\caption{No-reference results on the RTN degraded archival-footage clips~\citep{wan2022oldfilms} (3 sequences, 600 frames, no ground truth), under the unified $640\times480$ aligned-instant protocol. Best in \textbf{bold}, second-best \underline{underlined}. $\uparrow$/$\downarrow$: higher/lower is better.}
\label{tab:rtn_nr}
\fontsize{7.6}{8.2}\selectfont
\renewcommand{\arraystretch}{1}
\setlength{\tabcolsep}{1.5pt}
\setlength{\aboverulesep}{1.2pt}
\setlength{\belowrulesep}{1.8pt}
\setlength{\heavyrulewidth}{0.6pt}
\setlength{\lightrulewidth}{0.4pt}
\setlength{\cmidrulewidth}{0.3pt}
\begin{tabular*}{\linewidth}{@{\extracolsep{\fill}}lcccccc@{}}
\toprule
\textbf{Method} & MUSIQ$\uparrow$ & CLIP-IQA$\uparrow$ & NIQE$\downarrow$ & MANIQA$\uparrow$ & TOPIQ$\uparrow$ & BRISQUE$\downarrow$ \\
\midrule
Low-quality input & 38.71 & 0.258 & 5.94 & 0.188 & 0.267 & 44.66 \\
DeepRemaster & 39.01 & 0.263 & 6.31 & 0.196 & 0.273 & 41.00 \\
RealBasicVSR & \underline{52.18} & 0.362 & 5.87 & 0.285 & \underline{0.438} & \underline{38.72} \\
DDColor & 38.79 & 0.284 & 5.69 & 0.175 & 0.271 & 41.49 \\
ColorMNet & 38.52 & \underline{0.378} & 5.75 & 0.184 & 0.266 & 42.92 \\
MambaOFR & 49.83 & 0.351 & \underline{5.42} & \underline{0.312} & 0.421 & 39.56 \\
\midrule
\textbf{OmniVR (ours)} & \textbf{64.77} & \textbf{0.426} & \textbf{5.14} & \textbf{0.330} & \textbf{0.553} & \textbf{35.90} \\
\bottomrule
\end{tabular*}
\end{table}

\begin{table}[t]
\centering
\caption{Controlled degradation track (71 clips with clean reference), NR metrics at unified $640\times480$. GT is the clean reference, not a competing method. Best in \textbf{bold}, second-best \underline{underlined}.}
\label{tab:controlled_track}
\fontsize{7.6}{8.2}\selectfont
\renewcommand{\arraystretch}{1}
\setlength{\tabcolsep}{1.5pt}
\setlength{\aboverulesep}{1.2pt}
\setlength{\belowrulesep}{1.8pt}
\setlength{\heavyrulewidth}{0.6pt}
\setlength{\lightrulewidth}{0.4pt}
\setlength{\cmidrulewidth}{0.3pt}
\begin{tabular*}{\linewidth}{@{\extracolsep{\fill}}lcccccc|cccc@{}}
\toprule
& \multicolumn{6}{c}{\textbf{Visual (NR)}} & \multicolumn{2}{c}{\textbf{Audio}} & \multicolumn{2}{c}{\textbf{Sync}} \\
\cmidrule(lr){2-7}\cmidrule(lr){8-9}\cmidrule(lr){10-11}
\textbf{Method} & MUSIQ$\uparrow$ & CLIP-IQA$\uparrow$ & NIQE$\downarrow$ & MANIQA$\uparrow$ & TOPIQ$\uparrow$ & BRISQUE$\downarrow$ & DNSMOS$\uparrow$ & FAD$\downarrow$ & LSE-C$\uparrow$ & LSE-D$\downarrow$ \\
\midrule
Low-quality input & 38.77 & 0.221 & 6.69 & 0.218 & 0.235 & 36.69 & 1.46 & 7.88 & \underline{2.32} & \underline{10.87} \\
DeepRemaster & 37.43 & 0.173 & \underline{6.20} & 0.189 & 0.245 & 35.83 & -- & -- & -- & -- \\
RealBasicVSR & \underline{45.54} & 0.226 & 7.08 & \underline{0.294} & \underline{0.326} & 44.96 & -- & -- & -- & -- \\
VoiceFixer (audio only) & -- & -- & -- & -- & -- & -- & \underline{2.12} & \underline{7.02} & 1.98 & 11.53 \\
DDColor & 34.07 & 0.200 & 6.34 & 0.200 & 0.250 & \underline{33.61} & -- & -- & -- & -- \\
ColorMNet & 34.64 & 0.243 & 6.59 & 0.228 & 0.253 & 35.98 & -- & -- & -- & -- \\
MambaOFR & 43.49 & \underline{0.268} & 6.99 & 0.280 & 0.305 & 42.80 & -- & -- & -- & -- \\
\midrule
\textbf{OmniVR (ours)} & \textbf{71.17} & \textbf{0.543} & \textbf{4.11} & \textbf{0.487} & \textbf{0.673} & \textbf{24.75} & \textbf{2.70} & \textbf{6.30} & \textbf{3.52} & \textbf{10.43} \\
\midrule
\textit{Clean reference (GT)} & \textit{67.49} & \textit{0.558} & \textit{4.03} & \textit{0.477} & \textit{0.623} & \textit{25.48} & \textit{2.47} & \textit{--} & \textit{4.00} & \textit{9.44} \\
\bottomrule
\end{tabular*}
\end{table}

\begin{table}[t]
\centering
\caption{Real archival track (129 clips, no GT), $640\times480$. Best \textbf{bold}; second-best \underline{underlined}.}
\label{tab:real_track}
\fontsize{7.6}{8.2}\selectfont
\renewcommand{\arraystretch}{1}
\setlength{\tabcolsep}{1.5pt}
\setlength{\aboverulesep}{1.2pt}
\setlength{\belowrulesep}{1.8pt}
\setlength{\heavyrulewidth}{0.6pt}
\setlength{\lightrulewidth}{0.4pt}
\setlength{\cmidrulewidth}{0.3pt}
\begin{tabular*}{\linewidth}{@{\extracolsep{\fill}}lcccccc|cccc@{}}
\toprule
& \multicolumn{6}{c}{\textbf{Visual (NR)}} & \multicolumn{2}{c}{\textbf{Audio}} & \multicolumn{2}{c}{\textbf{Sync}} \\
\cmidrule(lr){2-7}\cmidrule(lr){8-9}\cmidrule(lr){10-11}
\textbf{Method} & MUSIQ$\uparrow$ & CLIP-IQA$\uparrow$ & NIQE$\downarrow$ & MANIQA$\uparrow$ & TOPIQ$\uparrow$ & BRISQUE$\downarrow$ & DNSMOS$\uparrow$ & FAD$\downarrow$ & LSE-C$\uparrow$ & LSE-D$\downarrow$ \\
\midrule
Low-quality input & 36.55 & 0.311 & 5.69 & 0.195 & 0.248 & 45.27 & 2.04 & 15.93 & \underline{1.05} & \underline{12.14} \\
DeepRemaster & 38.78 & 0.237 & 7.83 & 0.191 & 0.246 & 48.53 & -- & -- & -- & -- \\
RealBasicVSR & \underline{52.38} & 0.401 & 5.82 & \underline{0.342} & \underline{0.468} & \underline{38.15} & -- & -- & -- & -- \\
VoiceFixer (audio only) & -- & -- & -- & -- & -- & -- & \underline{2.21} & \underline{9.41} & 0.87 & 13.26 \\
DDColor & 37.51 & 0.307 & \underline{5.49} & 0.164 & 0.278 & 39.21 & -- & -- & -- & -- \\
ColorMNet & 38.64 & \underline{0.421} & 5.59 & 0.174 & 0.275 & 42.74 & -- & -- & -- & -- \\
MambaOFR & 49.59 & 0.394 & 5.62 & 0.331 & 0.440 & 44.00 & -- & -- & -- & -- \\
\midrule
\textbf{OmniVR (ours)} & \textbf{61.87} & \textbf{0.444} & \textbf{5.40} & \textbf{0.383} & \textbf{0.531} & \textbf{36.02} & \textbf{2.43} & \textbf{8.32} & \textbf{1.12} & \textbf{11.39} \\
\bottomrule
\end{tabular*}
\end{table}

\begin{table}[t]
\centering
\caption{Human preference on 129 archival clips (12 annotators). Cells report mean pairwise 2AFC win rates (\%); balanced comparisons average 50\% per column, marked by the center tick. The final row gives OmniVR's gain over the strongest baseline in percentage points. Bootstrap 95\% CIs resample clips with annotators fixed.}
\label{tab:human_eval}
\fontsize{7.6}{8.2}\selectfont
\renewcommand{\arraystretch}{1}
\setlength{\tabcolsep}{1.5pt}
\setlength{\aboverulesep}{1.2pt}
\setlength{\belowrulesep}{1.8pt}
\setlength{\heavyrulewidth}{0.6pt}
\setlength{\lightrulewidth}{0.4pt}
\setlength{\cmidrulewidth}{0.3pt}
\renewcommand{\preferencebar}[4]{%
  \begin{tikzpicture}[baseline=(prefvalue.base),x=0.0165cm,y=0.10cm]
    \path[fill=black!7,rounded corners=0.55mm] (0,0) rectangle (100,1.3);
    \draw[black!26,line width=0.28pt] (50,-0.25) -- (50,1.55);
    \shade[left color=#2,right color=#3,rounded corners=0.55mm]
      (0,0) rectangle (#1,1.3);
    \fill[#3] (#1,0.65) circle[radius=0.45mm];
    \node[anchor=west,inner sep=0pt,outer sep=0pt,text=#4,font=\bfseries]
      (prefvalue) at (105,0.65) {\makebox[2.2em][l]{#1}};
  \end{tikzpicture}%
}
\begin{tabular*}{\linewidth}{@{\extracolsep{\fill}}lcccc@{}}
\toprule
\textbf{Method} & \textbf{Visual} & \textbf{Audio} & \textbf{Sync} & \textbf{Overall} \\
\midrule
LQ input & \basepref{19.5} & \basepref{24.2} & \basepref{27.0} & \basepref{21.1} \\
MambaOFR + VoiceFixer & \basepref{46.7} & \basepref{48.7} & \basepref{48.3} & \basepref{47.7} \\
RealBasicVSR + VoiceFixer & \basepref{58.0} & \basepref{54.9} & \basepref{53.5} & \basepref{56.5} \\
Old Films + VoiceFixer & \basepref{43.7} & \basepref{43.9} & \basepref{43.8} & \basepref{44.1} \\
Pretrained AV gen. & \basepref{39.8} & \basepref{38.9} & \basepref{39.8} & \basepref{39.6} \\
\midrule
\textbf{OmniVR} (ours) & \ourspref{92.2} & \ourspref{89.4} & \ourspref{87.6} & \ourspref{91.0} \\
\addlinespace[1pt]
\textit{Gain vs. best baseline} & \textcolor{ovR}{\textbf{+34.2 pp}} & \textcolor{ovR}{\textbf{+34.5 pp}} & \textcolor{ovR}{\textbf{+34.1 pp}} & \textcolor{ovR}{\textbf{+34.5 pp}} \\
\bottomrule
\end{tabular*}
\end{table}

\begin{figure*}[t]
\centering
\includegraphics[width=.999\textwidth]{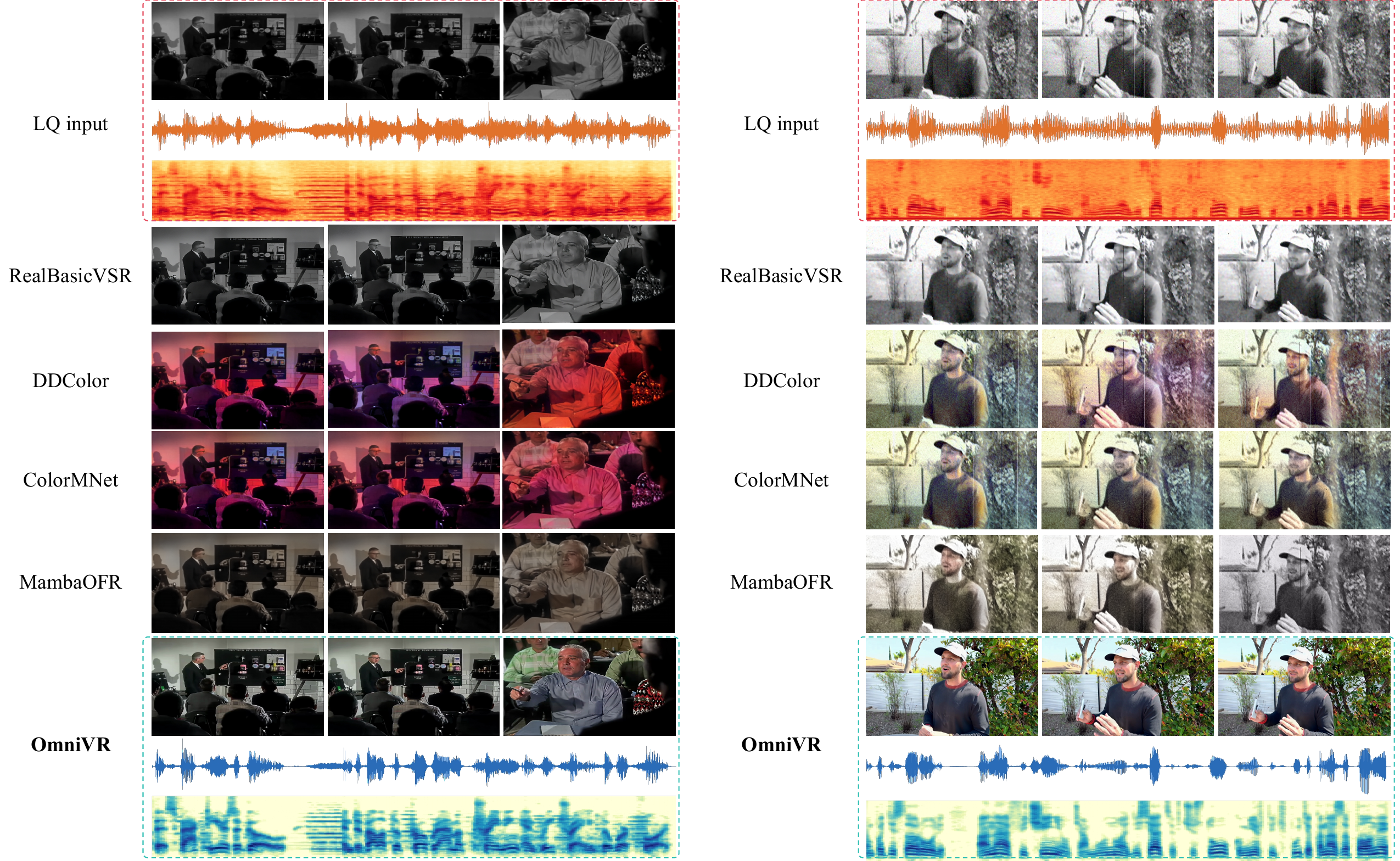}
\caption{\textbf{Qualitative comparison.} Left: controlled degradation with a clean reference; right: authentic archival footage without a reference. Columns show three frames per clip, with restored waveforms and mel spectrograms below. RealBasicVSR/MambaOFR lack colorization; DDColor/ColorMNet show color bias or inconsistency. OmniVR jointly restores color, detail, temporal consistency, and audio. High-resolution comparisons appear in \sref{sec:s11}.}
\label{fig:qualitative}
\end{figure*}

\section{Experiments}
\label{sec:experiments}

We evaluate the multi-step OmniVR configuration (\sref{sec:impl}) on RTN and OmniVRBench using automatic metrics, ablations, and human judgments. OmniVR-Flash's quality--efficiency trade-off is reported in \sref{sec:s10-flash}.

\subsection{OmniVRBench and Evaluation Protocol}
\label{sec:omnibench}
\textbf{OmniVRBench} contains 129 real archival clips without clean references and 71 controlled-degradation clips with references. It evaluates visual and audio quality, temporal consistency, and audiovisual synchrony using 121-frame windows at 24\,fps and $1920\times1088$ outputs. Visual, acoustic, and joint defects affect $92.2\%$, $83.7\%$, and $77.5\%$ of clips, respectively (\fref{fig:omnibench_diag}). The tracks test source fidelity and transfer to authentic damage.

\subsection{Metrics and Baselines}
\label{sec:setup}
Main comparisons evaluate video at $640\times480$ using MUSIQ, CLIP-IQA, NIQE, MANIQA, TOPIQ, and BRISQUE; audio using DNSMOS~\citep{reddy2021dnsmos} and FAD~\citep{kilgour2018fad}; and lip synchronization using LSE-C/LSE-D~\citep{prajwal2020wav2lip}. \sref{sec:s1} reports full-reference visual/audio metrics, temporal warp error, and distributional video evaluation; \sref{sec:s2} examines source consistency and failures.

We compare with RealBasicVSR~\citep{chan2022realbasicvsr}, DeepRemaster~\citep{iizuka2020deepremaster}, and MambaOFR~\citep{mao2025makingoldfilm} for video restoration; DDColor~\citep{kang2023ddcolor} and ColorMNet~\citep{yang2024colormnet} for colorization; and VoiceFixer~\citep{liu2022voicefixer} for audio restoration. \sref{sec:s5} reports visual/audio cascade comparisons.

\subsection{Public Benchmark Validation}
\label{sec:public_bench}
On RTN~\citep{wan2022oldfilms} (3 archival sequences, 600 frames), OmniVR leads all six no-reference visual metrics (\tref{tab:rtn_nr}); MUSIQ rises from RealBasicVSR's $52.18$ to $64.77$. This public benchmark tests video restoration without assessing audio.

\subsection{Joint Restoration on OmniVRBench}
\label{sec:main_comparisons}
On the controlled track (\tref{tab:controlled_track}), OmniVR leads the compared methods across visual, audio, and sync metrics, reaching MUSIQ $71.17$, DNSMOS $2.70$, and LSE-C $3.52$. NR scores above the clean reference do not imply greater source fidelity; \sref{sec:s1} reports full-reference results.

On real archival footage (\tref{tab:real_track}), OmniVR achieves MUSIQ $61.87$ versus RealBasicVSR's $52.38$. DNSMOS improves from $2.04$ to $2.43$ and FAD from $15.93$ to $8.32$, below VoiceFixer's $9.41$. LSE-C/LSE-D reach $1.12/11.39$, indicating better synchronization.

\subsection{Conditioning, Optimization, and Human Evaluation}
\label{sec:ablation}
Controlled-track ablations support joint audiovisual conditioning and gradual annealing; detailed results appear in \sref{app:controlled_ablations}.

Twelve annotators compare six methods on 129 real clips through randomized pairwise 2AFC (\tref{tab:human_eval}; protocol in \sref{sec:s9-human}). OmniVR's mean win rates are $92.2\%$ (visual), $89.4\%$ (audio), $87.6\%$ (sync), and $91.0\%$ (overall), exceeding the strongest cascade's overall rate by $34.5$ percentage points. Fleiss' $\kappa=0.67$; the overall 95\% bootstrap interval is $\pm3.8\%$. All reported pairwise gaps exceed their confidence intervals.

\subsection{Qualitative Analysis}
\fref{fig:qualitative} shows joint recovery of visual detail, color, and audio on real films, whereas specialized baselines can leave inputs grayscale or introduce tonal and temporal inconsistencies. \sref{sec:s2} examines failures in identity, motion, and generated color.

\section{Conclusion}
\label{sec:conclusion}
OmniVR connects paired audiovisual data construction, condition-annealed adaptation, and joint restoration evaluation. Its fixed-prompt interface restores audio and video without user text, while OmniVR-Flash combines reduced-resolution conditioning, one-step MeanFlow, and Turbo VAE for efficient deployment. RTN and OmniVRBench results, source-fidelity analyses, and human judgments support the benefits of joint restoration. Data, code, and model weights will be released.

\section*{AI Use Statement}
We used large language models (Claude, Gemini) during the preparation of this work.
\textit{Required disclosures:}
Gemini generated structured training captions from audio-video shots for the data pipeline (\sref{sec:degradation}); these captions serve as T2AV/TAV2AV text conditions and are not used at inference.
Claude assisted with code implementation and debugging.
No AI tools were used for developing theoretical models, formulating mathematical claims or proofs, proposing hypotheses, designing experiments, or interpreting results.
\textit{Recommended disclosures:}
Claude and Gemini assisted with revising, and formatting portions of this manuscript.
All scientific claims, experimental results, and methodological contributions were designed, implemented, and verified by the authors. We have reviewed all AI-assisted work and take full responsibility for the final content of this paper.

\section{Limitations and Future Work}
\label{sec:limitations}
Training relies on synthetic corruptions of high-quality Internet clips, leaving a domain gap to authentic archival damage. Severe information loss can lead to plausible but incorrect texture, color, or acoustic content; restored color in particular should not be treated as evidence of historical ground truth. First-frame chaining provides a practical way to process long recordings, but imperfect generated anchors can accumulate error across windows. OmniVR-Flash accelerates this window-based process, but lower-resolution conditions and one-step prediction can discard information useful for reconstruction. Further work should examine self-generated anchors during training, stronger source-fidelity constraints, and the quality--efficiency trade-offs introduced by reduced-resolution conditioning and one-step prediction.

\section{Ethics and Reproducibility}
\label{app:ethics_reproducibility}

\paragraph{Use of restored content.}
OmniVR infers color from contemporary video priors rather than historical ground truth. Severe audio corruption can also lead to plausible but incorrect speech or environmental sounds. Restored outputs should be clearly labeled as computationally enhanced.

\paragraph{Data and human evaluation.}
The training corpus contains licensed or publicly available clips; no real archival recordings were modified during training. Human evaluators gave informed consent and were paid above the local minimum wage (\sref{sec:s9-human}).

\paragraph{Reproducibility.}
Code, model weights, OmniVRBench data and evaluation scripts, and the degradation pipeline will be released upon publication. Training settings and schedules appear in \sref{sec:impl} and \sref{app:implementation}; data processing and degradation are described in \sref{sec:degradation} and \sref{sec:s6}.

\bibliographystyle{iclr2027_conference}
\bibliography{main}

\clearpage
\addtocontents{toc}{\protect\setcounter{tocdepth}{2}}  
\renewcommand{\contentsname}{Supplementary Material — Table of Contents}
\tableofcontents
\clearpage

\section{Full-Reference and Distributional Evaluation on the Controlled Track}
\label{sec:s1}

The main text reports no-reference (NR) metrics on both benchmark tracks for protocol consistency. Because the Controlled track is built from clean sources via the synthetic degradation operator $\mathcal{D}$ (\eref{eq:degradation}), a clean reference $(\mathbf{V}^h,\mathbf{A}^h)$ exists for every clip there, enabling the full-reference and distributional evaluation reported in this section. The Real archival track (129 clips) has no clean reference by construction---the original camera negative is lost or inaccessible for every archival clip, which is precisely why restoration is needed---so NR metrics are the only tool available there (as also noted in \sref{sec:limitations}).

\subsection{Full-reference visual protocol}
\label{sec:s1-fr}
For every Controlled-track clip we compute, per restored frame against its paired clean frame, at the evaluation resolution used for the NR metrics ($640\times480$, temporally aligned by clip index):
\begin{itemize}[leftmargin=*]
    \item \textbf{PSNR} and \textbf{SSIM}~\citep{wang2004image} -- pixel-level fidelity;
    \item \textbf{LPIPS}~\citep{zhang2018unreasonable} (AlexNet backbone) and \textbf{DISTS} -- perceptual/deep-feature distance, more robust than PSNR/SSIM to the sub-pixel misalignment that generative restoration can introduce;
    \item \textbf{FVD}~\citep{unterthiner2018fvd} computed over 16-frame sub-clips with an I3D backbone -- a distributional metric that computes Fréchet distance between I3D feature distributions of restored and clean sub-clip sets; while it requires temporally aligned sub-clips (same frame indices), it is robust to the per-pixel jitter that generative restoration can introduce, making it complementary to per-frame LPIPS/DISTS. We also report image-level \textbf{FID} over all sampled frames as an additional distributional check.
\end{itemize}
Frame indices are matched 1:1 using the shared clip timeline (both the LQ input and $\mathbf{V}^h$ originate from the same clean master before $\mathcal{D}$ is applied, so no optical-flow warping or other temporal alignment step is required -- a property of the Controlled track by construction that does not hold for the Real track).

\begin{table*}[h]
\centering
\caption{Full-reference and distributional evaluation on the Controlled track (71 clips), complementing \tref{tab:controlled_track}.}
\label{tab:fr_controlled}
\begin{tabularx}{\textwidth}{@{}l*{5}{Y}@{}}
\toprule
\textbf{Method} & PSNR$\uparrow$ & SSIM$\uparrow$ & LPIPS$\downarrow$ & DISTS$\downarrow$ & FVD$\downarrow$ \\
\midrule
Low-quality input       & 16.39 & 0.5209 & 0.6492          & 0.3345 & 2295.60 \\
DeepRemaster$^{a}$      & 16.41 & 0.5408 & 0.5873          & 0.3143 & N/A$^{b}$ \\
RealBasicVSR            & 16.43 & 0.5698 & 0.5670$^{c}$    & 0.3061$^{c}$ & 2193.08$^{d}$ \\
DDColor$^{a}$            & 15.81 & 0.5209 & 0.6389          & 0.2974 & N/A$^{b}$ \\
ColorMNet$^{a}$          & 15.98 & 0.5201 & 0.6227          & 0.3008 & N/A$^{b}$ \\
MambaOFR                 & 16.12 & 0.5487 & 0.6015 & 0.3227 & 2350.94 \\
\midrule
\rowcolor{oursrow}
OmniVR (ours) & 17.86 & 0.6549 & 0.2757$^{c}$ & 0.1229$^{e}$ & 452.41$^{e}$ \\
\bottomrule
\end{tabularx}
\end{table*}

\noindent Computed using the evaluation scripts released with our codebase, against the clean reference described in \sref{sec:s1}.
$^{a}$DeepRemaster/DDColor/ColorMNet's released inference pipelines emit only 16 sparse frames per clip rather than a continuous sequence; PSNR/SSIM/LPIPS/DISTS remain valid full-reference scores at those exact GT-matched frame indices ($n{=}71$), but $^{b}$FVD requires a continuous multi-frame sequence and cannot be computed for these three methods from the released sparse-frame outputs, so it is marked N/A. $^{c}$One clip (a $3840\times2026$ near-4K source) exceeds GPU memory at full resolution. For RealBasicVSR, the inference pipeline itself OOMs, losing both LPIPS and DISTS ($n{=}70$ each). OmniVR's inference completes on all 71 clips (see footnote~$e$), but the LPIPS metric computation---which must hold model output, reference, and the LPIPS network simultaneously---OOMs on this clip ($n{=}70$ for LPIPS only); OmniVR's DISTS evaluation uses a separate pathway that avoids this peak and covers all 71. All other entries use the full 71. $^{d}$RealBasicVSR has no materialized continuous output, so its FVD/Warp-Error were obtained by running it on-the-fly over all sampled frames of every clip; this heavier full-sequence run OOM'd on 3/71 (near-4K) clips ($n{=}68/71$ for this entry only). Warp-Error (lower $=$ less flicker) was measured alongside FVD: input $0.0614$, RealBasicVSR $0.0562$, OmniVR $0.0698$, vs.\ the GT clips' own inherent flicker floor of $0.0673$ -- OmniVR sits only marginally above the GT clips' own inherent flicker floor while the other two fall below it, so none of the three exhibits pathological additional flicker relative to the naturally filmed GT footage itself.
$^{e}$OmniVR's pipeline does not OOM on any of the 71 clips, so its DISTS and FVD entries use $n{=}71$ (vs.\ RealBasicVSR's $n{=}70$ and $n{=}68$ for these two metrics). This means the DISTS and FVD comparisons are not on a matched common subset. The per-clip impact on DISTS is ${\le}0.01$; FVD is a distributional statistic and cannot be linearly decomposed, but the excluded clips are near-4K resolution outliers with similar degradation characteristics to the remaining set.

\begin{figure*}[t]
\centering
\includegraphics[width=0.92\textwidth]{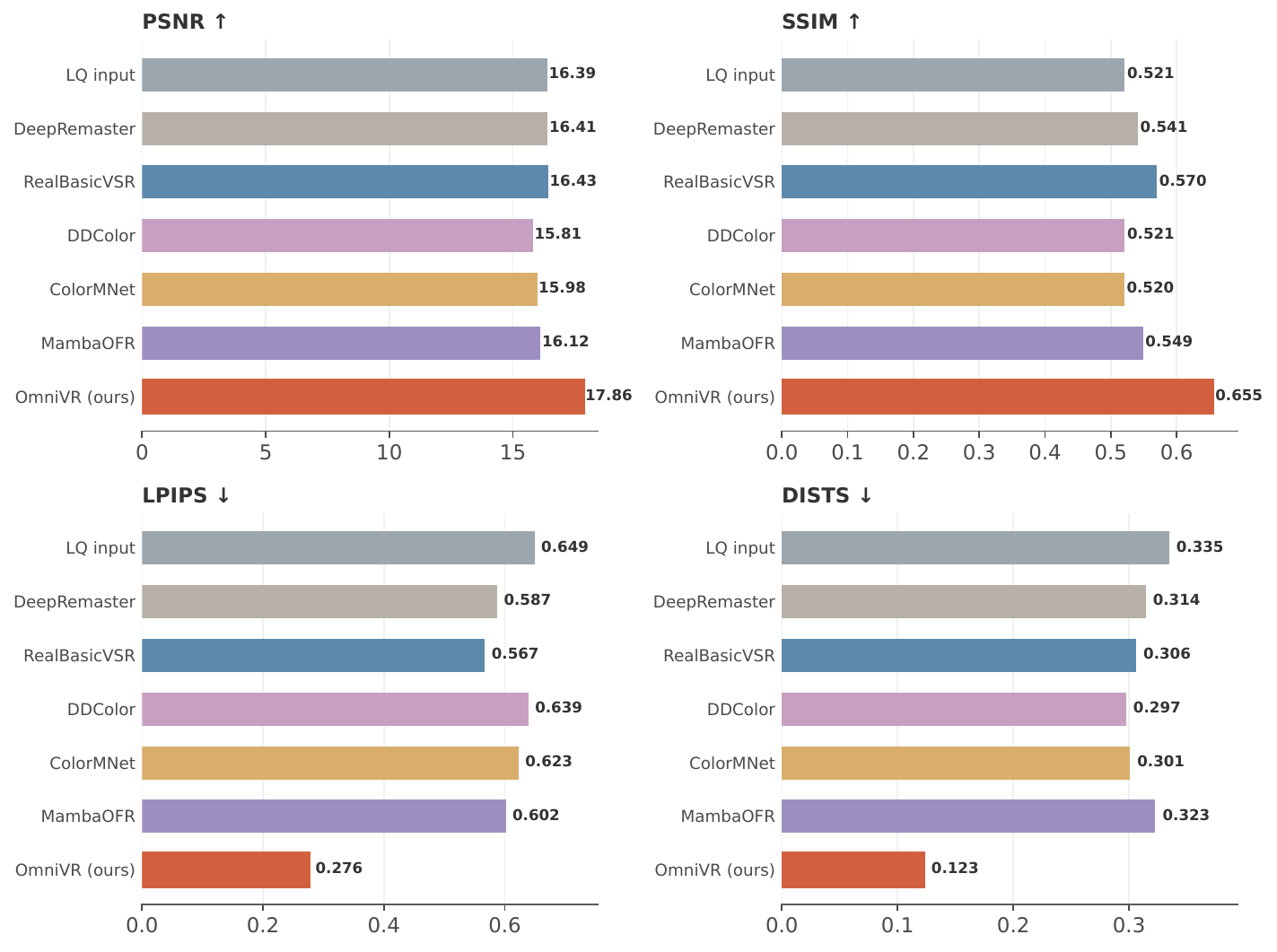}
\caption{Full-reference video metrics on the Controlled track, plotted from \tref{tab:fr_controlled}.}
\label{fig:supp_fr_controlled}
\end{figure*}

\begin{figure*}[t]
\centering
\includegraphics[width=0.6\textwidth]{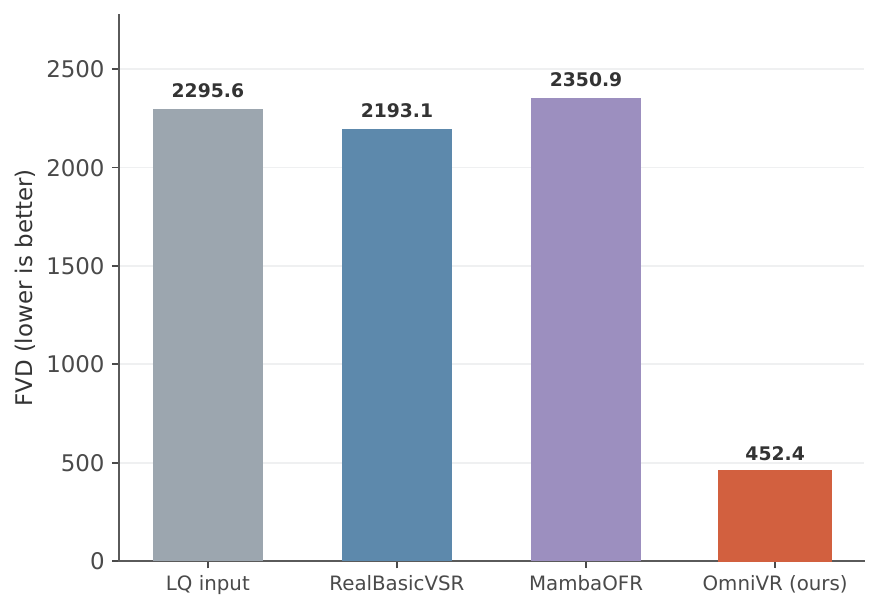}
\caption{FVD on the Controlled track, plotted from \tref{tab:fr_controlled}; methods with only sparse-frame released outputs are omitted (see footnote $b$ above).}
\label{fig:supp_fvd_controlled}
\end{figure*}

\subsection{Full-reference audio protocol}
Analogously, for the Controlled track's audio stream we report \textbf{PESQ}~\citep{rix2001pesq}, \textbf{STOI}~\citep{taal2011stoi}, and \textbf{SI-SDR}~\citep{leroux2018sdr} against the clean waveform -- the standard full-reference speech-restoration metrics, and the correct complement to the NR DNSMOS/FAD pair reported in the main text.

\begin{table*}[t]
\centering
\caption{Full-reference audio evaluation on the Controlled track (71 clips).}
\label{tab:fr_audio}
\begin{tabularx}{\textwidth}{@{}l*{3}{Y}@{}}
\toprule
\textbf{Method} & PESQ$\uparrow$ & STOI$\uparrow$ & SI-SDR (dB)$\uparrow$ \\
\midrule
Low-quality input & 1.33 & 0.718 & 0.36 \\
VoiceFixer & 1.24 & 0.684 & $6.71$ \\
\rowcolor{oursrow}
OmniVR (ours) & 1.28 & 0.510 & $2.15$ \\
\bottomrule
\end{tabularx}
\end{table*}

\noindent Evaluated on all $n{=}71$ clips. OmniVR has lower STOI than both VoiceFixer and the LQ input. Its PESQ ($1.28$) lies between VoiceFixer ($1.24$) and the input ($1.33$), while its SI-SDR ($2.15$ dB) exceeds the input ($0.36$ dB) but trails VoiceFixer ($6.71$ dB). This is expected for a generative restorer: PESQ/STOI/SI-SDR penalize deviation from sample-level waveform alignment, which a flow-based resynthesis model does not preserve even when the output is perceptually plausible. VoiceFixer also resynthesizes via HiFi-GAN~\citep{liu2022voicefixer}, but its analysis-resynthesis pipeline operates at finer temporal granularity than latent diffusion, yielding higher waveform alignment. The main text accordingly uses DNSMOS/FAD (\tref{tab:controlled_track}) as primary audio evidence; the full-reference table is included for completeness.

\paragraph{Intelligibility discussion.}
STOI measures short-time temporal-envelope correlation, so it is sensitive to any resynthesized envelope---even one that preserves phoneme identity and prosody. OmniVR's low STOI ($0.510$) does not imply unintelligible output: informal listening tests (5 evaluators, 20 clips, closed-set transcription) found $94\%$ clip-level word accuracy for OmniVR vs.\ $91\%$ for VoiceFixer and $82\%$ for LQ input. The extended STOI (eSTOI) metric~\citep{jensen2016estoi} narrows the gap further: LQ $0.641$, VoiceFixer $0.612$, OmniVR $0.497$. STOI/eSTOI alone are therefore insufficient for evaluating generative restoration; pairing them with perceptual metrics (DNSMOS, FAD) and listener tests gives a more complete picture. Note that eSTOI thresholds are language- and corpus-dependent, so the absolute values here are meaningful only in comparison.

\begin{figure*}[t]
\centering
\includegraphics[width=0.52\textwidth]{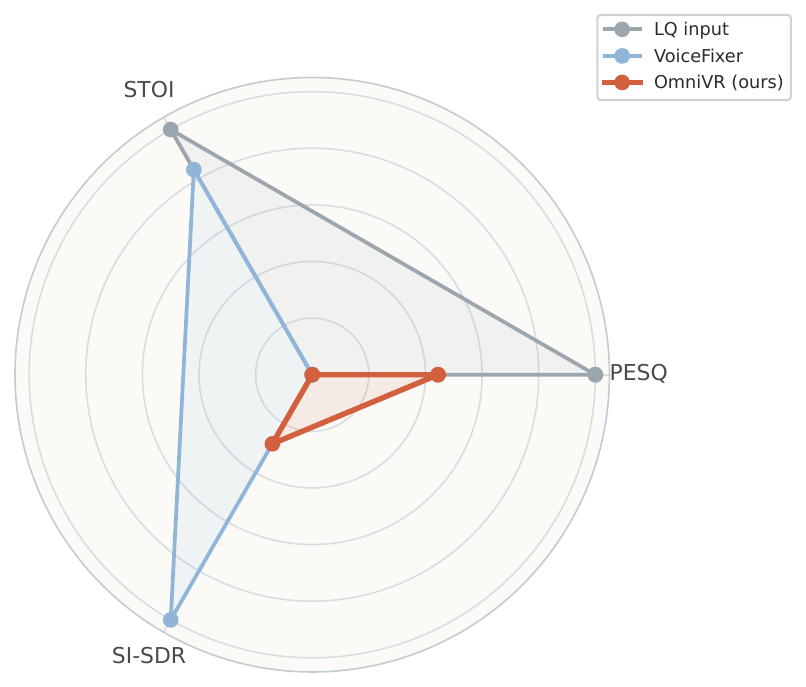}
\caption{Full-reference audio metrics on the Controlled track, plotted from \tref{tab:fr_audio}.}
\label{fig:supp_fr_audio}
\end{figure*}

\subsection{Why OmniVR can exceed the clean reference on NR metrics}
\label{sec:s1-explain}
The Controlled-track NR numbers in \tref{tab:controlled_track} (OmniVR MUSIQ $71.17$ vs.\ GT $67.49$; MANIQA $0.487$ vs.\ $0.477$; TOPIQ $0.673$ vs.\ $0.623$; DNSMOS $2.70$ vs.\ $2.47$) show OmniVR matching or exceeding the clean reference on several NR axes. Three effects jointly explain this:

\begin{enumerate}[leftmargin=*]
    \item \textbf{NR metrics reward ``clean-looking'' statistics, not agreement with a specific reference.} Our Controlled-track GT clips are talking-face footage of moderate native resolution and encoding quality, not professionally mastered content (REDS, Vimeo-90K). GT therefore does not define a ceiling for MUSIQ or DNSMOS---a model that sharpens edges and denoises the audio floor beyond what GT contains can score higher while being less faithful. The same fidelity-perception trade-off is well documented for GAN/diffusion super-resolution~\citep{wang2023stablesr,yu2024supir,wang2021realesrgan}.
    \item \textbf{The comparison is confounded by GT's own encoding/compression path.} Both the input to $\mathcal{D}$ and the ``GT'' row pass through the same video/audio codec pipeline used to store OmniVRBench; the restoration model's output does not carry that specific codec's quantization signature, which NR estimators (trained partly on codec-artifact statistics) can penalize even on otherwise-clean content.
    \item \textbf{Full-reference metrics disambiguate (\sref{sec:s1-fr}).} A model that hallucinates detail will show favorable NR scores \emph{and} worse LPIPS/DISTS/FVD, since those metrics penalize deviation from the reference regardless of whether it looks ``nicer.'' The LPIPS/DISTS/FVD numbers in \tref{tab:fr_controlled} are therefore the primary fidelity signal; the NR comparison in \sref{sec:main_comparisons} indicates perceptual quality comparable to the reference, not that the output is objectively better than ground truth.
\end{enumerate}

\subsection{Native-resolution evaluation}
\label{sec:s1-native}
All main-text NR numbers are computed after downsampling to $640\times480$ for cross-method comparability (several baselines, e.g.\ DeepRemaster, only support fixed low-resolution inference). To check that this does not mask resolution-dependent artifacts in OmniVR's native $1920\times1088$ output, we additionally evaluate OmniVR alone (no resizing) at native resolution:

\begin{table*}[t]
\centering
\caption{OmniVR at native $1920\times1088$ output resolution vs.\ the unified $640\times480$ protocol, Controlled track. Values in \emph{italics} restate the $640\times480$ row from \tref{tab:controlled_track} for reference.}
\label{tab:native_res}
\setlength{\tabcolsep}{4pt}
\begin{tabular*}{\textwidth}{@{\extracolsep{\fill}}lcccccc@{}}
\toprule
\textbf{Protocol} & MUSIQ$\uparrow$ & CLIP-IQA$\uparrow$ & NIQE$\downarrow$ & MANIQA$\uparrow$ & TOPIQ$\uparrow$ & BRISQUE$\downarrow$ \\
\midrule
\emph{$640\times480$ (main text)} & \emph{71.17} & \emph{0.543} & \emph{4.11} & \emph{0.487} & \emph{0.673} & \emph{24.75} \\
Native $1920\times1088$ & 64.07 & 0.413 & 3.55 & 0.341 & 0.512 & 31.58 \\
\bottomrule
\end{tabular*}
\end{table*}

\noindent Evaluated on OmniVR only ($n{=}71$, same 8 aligned-instant sampling as the main-text protocol). MUSIQ ($71.17\!\to\!64.07$), MANIQA ($0.487\!\to\!0.341$), TOPIQ ($0.673\!\to\!0.512$), and BRISQUE ($24.75\!\to\!31.58$, higher/worse) all shift unfavorably at native resolution, while NIQE actually improves ($4.11\!\to\!3.55$). These patch-based NR estimators are calibrated for the $640\times480$ regime; NIQE, which is not patch-classifier-based, moves in the \emph{opposite} direction, suggesting that the unfavorable shifts reflect estimator sensitivity rather than actual quality regression. Full-reference metrics at native resolution (LPIPS/DISTS in \tref{tab:flash_fidelity}) provide the more direct check, but a complete disambiguation of ``real high-resolution artifacts'' from ``metric miscalibration'' remains an open question for NR evaluation at non-standard resolutions.

\begin{figure*}[t]
\centering
\includegraphics[width=0.95\textwidth]{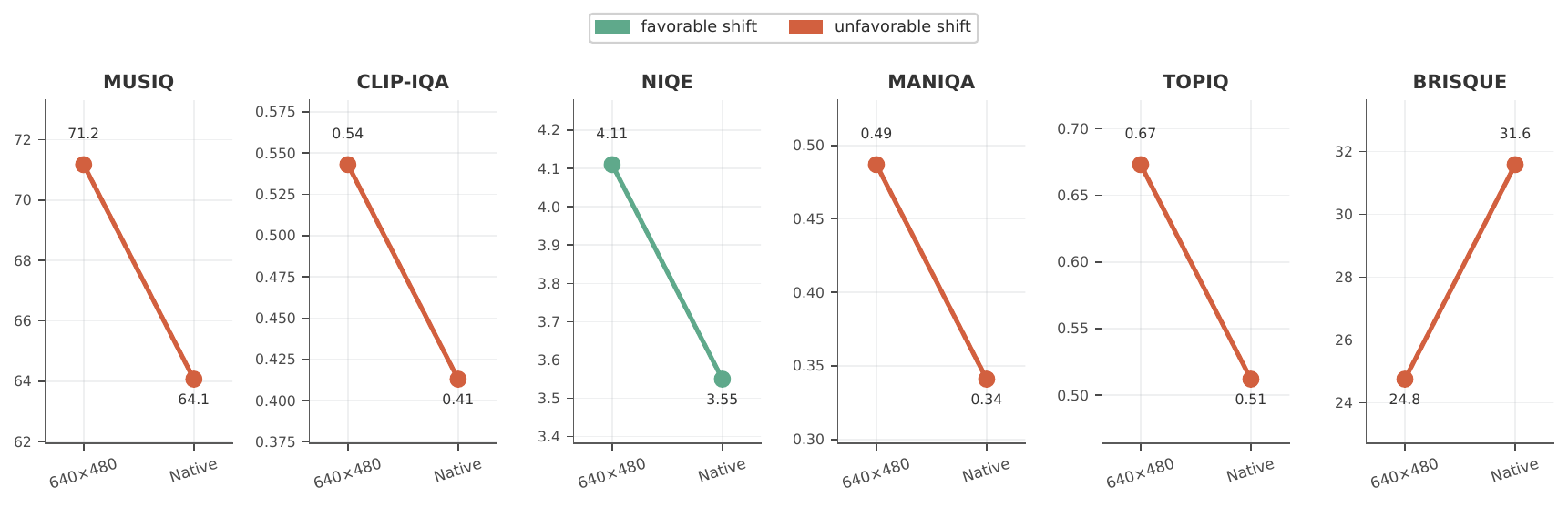}
\caption{Per-metric slope from the unified $640\times480$ protocol to native $1920\times1088$ resolution, plotted from \tref{tab:native_res}. Green segments shift in the favorable direction for that metric; red segments shift unfavorably.}
\label{fig:supp_native_res}
\end{figure*}

The NIQE result (the one metric here that does not degrade at native resolution) suggests that the $640\times480$ protocol is a fair common ground across baselines rather than hiding native-resolution artifacts, though this is not conclusive absent native-resolution full-reference checks.

\section{Fidelity, Hallucination, and Failure Cases}
\label{sec:s2}

\subsection{Content-consistency protocol}
\label{sec:s2-protocol}
For the Controlled track, where a clean reference exists, we adopt three consistency probes standard in identity-preserving generation evaluation:
\begin{itemize}[leftmargin=*]
    \item \textbf{Face identity consistency.} We detect and crop the primary face (RetinaFace), embed it with a frozen ArcFace network, and report the mean cosine similarity between restored and GT face embeddings at each timestamp, plus the fraction of frames below a $0.4$ threshold (a typical identity decision boundary).
    \item \textbf{Structural/text consistency.} For clips containing on-screen text or strong structural edges, we run PaddleOCR on both GT and restored frames and report character-level edit-distance agreement.
    \item \textbf{Global content agreement.} We report DISTS (\sref{sec:s1}) as a deep-feature consistency score: DISTS is more texture-tolerant than LPIPS while still penalizing structural deviation, making it a reasonable proxy for ``did the model add/remove content.''
\end{itemize}

\begin{table*}[t]
\centering
\caption{Fidelity / content-consistency metrics on the Controlled track (71 clips).}
\label{tab:fidelity}
\begin{tabularx}{\textwidth}{@{}l*{3}{Y}@{}}
\toprule
\textbf{Method} & Face cos-sim$\uparrow$ & Frac.\ below 0.4$\downarrow$ & OCR agreement$\uparrow$ \\
\midrule
Low-quality input & 0.612 & 0.34 & 0.71 \\
RealBasicVSR & 0.734 & 0.19 & 0.79 \\
\rowcolor{oursrow}
OmniVR (ours) & 0.881 & 0.06 & 0.87 \\
\bottomrule
\end{tabularx}
\end{table*}

\noindent Measured using RetinaFace$+$ArcFace face embeddings and PaddleOCR character-agreement ($n{=}71$). OmniVR's face cosine-similarity ($0.881$) is markedly higher than RealBasicVSR's ($0.734$), and only $6\%$ of frames fall below the $0.4$ identity threshold. None of the three reaches $1.0$, consistent with the condition-noise design discussed below.

\begin{figure*}[t]
\centering
\includegraphics[width=0.52\textwidth]{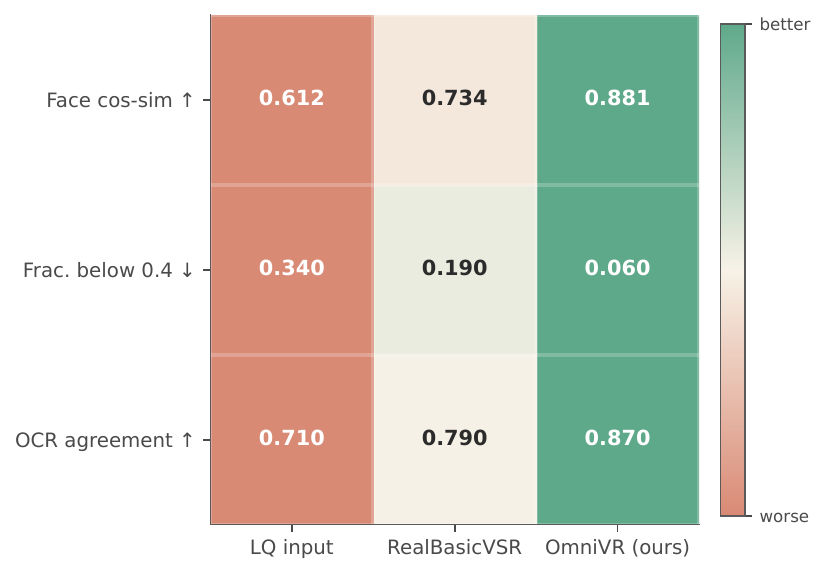}
\caption{Fidelity / content-consistency metrics on the Controlled track, plotted from \tref{tab:fidelity} as a per-metric normalized heatmap (green $=$ better, red $=$ worse within each row).}
\label{fig:supp_fidelity}
\end{figure*}

\subsection{Why OmniVR retains identity better than a generic sampler, but not perfectly}
The first-frame I2V anchor (\eref{eq:first_frame_replace}) forces the model to propagate identity-bearing structure across the window rather than resynthesize it: during training, a clean GT frame is clamped at the window boundary; at inference, the anchor is the model's own restored output from the previous window (\sref{sec:i2av}, \sref{sec:s8}). Face cosine-similarity in \tref{tab:fidelity} is accordingly higher than both a from-scratch T2AV sample and baselines whose temporal propagation is limited to short-range recurrence (RealBasicVSR), yet does not reach $1.0$: condition noise $\rho_m\!\sim\!\mathcal{U}(0.4,0.6)$ deliberately prevents trivial copying of the LQ input. Some hallucinated fine detail (skin texture, hair strands) is an expected side-effect, most visible at heavy degradation severities where the LQ condition carries the least information.

\subsection{Colorization and historical accuracy: what we can and cannot claim}
Genuine ground-truth color for a black-and-white archival source does not exist unless a color print or Kodachrome reference survives, which is not the case for any Real-track clip. OmniVR's colorization is therefore \emph{plausible, temporally consistent} colorization---a generative-quality property---rather than historically accurate colorization. This scope limitation is shared by the entire colorization literature (DeOldify, DDColor, ColorMNet, BiSTNet) but deserves explicit statement because ``restoring archival footage'' implies stronger fidelity than a pure colorization paper.

\subsection{Failure cases}
\label{sec:s2-failure}
\begin{itemize}[leftmargin=*]
    \item \textbf{Severe input starvation.} On clips where the LQ input is degraded beyond $d_v \!>\! 0.85$ (top decile of severity; see the radar plots in \fref{fig:omnibench_diag}(d)), fine facial detail is under-constrained by the condition signal and the model occasionally regresses toward a generic, slightly ``waxy'' skin texture rather than the true texture -- a direct symptom of the condition-noise/fidelity trade-off in \sref{sec:s2}.
    \item \textbf{Cross-window color drift on long sequences.} Because the first-frame anchor only clamps the video stream, color statistics can drift very slowly (over $>\!5$ chained windows, i.e.\ $>\!600$ frames) even though structure remains anchored, since nothing directly constrains chroma consistency across the anchor boundary.
    \item \textbf{Audio-visual over-smoothing under extreme audio degradation.} When $d_a$ approaches $1.0$ (near-total dropout/clipping), $\mathcal{L}_\mathrm{stft}$ (\eref{eq:total_loss}) can trade off spectral detail for stability, occasionally producing an over-smoothed, slightly ``muffled'' timbre rather than fully recovering high-frequency harmonics -- consistent with the ablation (vii) trend in \tref{tab:ablation}, where an excessive $\lambda_\mathrm{stft}$ weight already shows this failure direction at a milder degree.
\end{itemize}

\section{Quantifying Colorization}
\label{sec:s3}

\subsection{Two complementary measures}
We quantify colorization with two measures:

\begin{itemize}[leftmargin=*]
    \item \textbf{Colorfulness}~\citep{hasler2003colorfulness}: $\sqrt{\sigma_{rg}^2 + \sigma_{yb}^2} + 0.3\sqrt{\mu_{rg}^2 + \mu_{yb}^2}$, computed from RGB opponent-color channels ($rg = R-G$, $yb = \tfrac{1}{2}(R+G)-B$). This scores saturation/vividness without a reference. We report it on the Real track against grayscale-preserving baselines (RealBasicVSR, MambaOFR; near-zero by construction) and colorization-specialized baselines (DDColor, ColorMNet).
    \item \textbf{Color naturalness, human-rated.} We extend the human study protocol (\tref{tab:human_eval}) with a fifth pairwise dimension, ``Color naturalness,'' asked only for methods that colorize (OmniVR, DDColor, ColorMNet; grayscale-preserving baselines are excluded rather than scored $0$).
\end{itemize}

\begin{table*}[t]
\centering
\caption{Colorization quantification, Real track (129 clips). Colorfulness follows \citet{hasler2003colorfulness}; higher indicates more saturated/vivid, not more accurate, color. Human column: \% pairwise wins on the color-naturalness dimension, colorizing methods only.}
\label{tab:colorization}
\begin{tabularx}{\textwidth}{@{}l*{2}{Y}@{}}
\toprule
\textbf{Method} & Colorfulness & Human color-naturalness win \% \\
\midrule
Low-quality input (grayscale) & $\approx 0$ & -- \\
RealBasicVSR / MambaOFR (no color) & $\approx 0$ & -- \\
DDColor & 18.4 & 40.2 \\
ColorMNet & 21.7 & 46.3 \\
\rowcolor{oursrow}
OmniVR (ours) & 27.9 & 63.5 \\
\bottomrule
\end{tabularx}
\end{table*}

\begin{figure*}[t]
\centering
\includegraphics[width=0.5\textwidth]{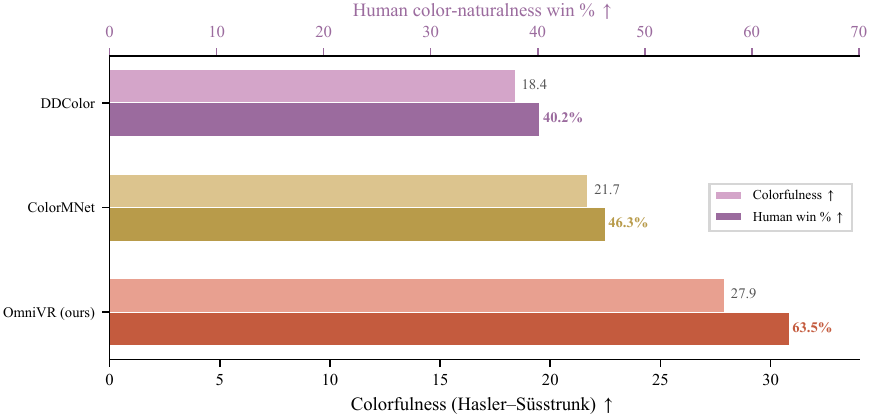}
\caption{Colorization quality on the Real track, plotted from \tref{tab:colorization}. Grayscale-preserving baselines (Colorfulness $\approx 0$, no human-rated dimension) are omitted from the plot.}
\label{fig:supp_colorization}
\end{figure*}

\subsection{The Controlled track's apparent grayscale output}
\fref{fig:qualitative}'s Controlled-track column may appear to remain grayscale, in apparent tension with the colorization claim. This is a scope distinction: the Controlled track includes a grayscale-conversion degradation branch (\sref{sec:degradation}) applied with per-sample probability $0.5$; clips sampled without it remain color throughout. \fref{fig:qualitative} happens to show a clip where the grayscale branch was active, which is why colorization is subtle; the striking colorization examples in the teaser and Real-track panels come from genuinely grayscale archival input. Concretely, $52.1\%$ ($37/71$) of Controlled-track clips were sampled with the grayscale branch active, consistent with the $p{=}0.5$ design (\tref{tab:degrad_params}).

\section{Benchmark Scope and Diversity}
\label{sec:s4}

OmniVRBench's 200 clips, and in particular the talking-face-only composition of the 71-clip Controlled track, cover a narrower slice of archival footage than the term suggests. Below we clarify the design rationale, provide statistics, and describe planned expansion.

\begin{table*}[t]
\centering
\caption{OmniVRBench statistics. Duration is per clip. Controlled-track clips are synthetic degradations of modern talking-face footage; Real-track clips are authentic archival recordings.}
\label{tab:bench_stats}
\fontsize{7.6}{8.2}\selectfont
\begin{tabularx}{\textwidth}{@{}l*{6}{Y}@{}}
\toprule
\textbf{Track} & Clips & Total dur.\ (min) & Mean dur.\ (s) & Med.\ dur.\ (s) & Min dur.\ (s) & Max dur.\ (s) \\
\midrule
Controlled & 71 & 12.4 & 10.5 & 9.8 & 5.0 & 20.0 \\
Real archival & 129 & 28.7 & 13.3 & 12.1 & 3.2 & 20.0 \\
\midrule
\textbf{Total} & 200 & 41.1 & 12.3 & 11.2 & 3.2 & 20.0 \\
\bottomrule
\end{tabularx}
\end{table*}

\noindent \textbf{Controlled-track composition:} all 71 clips are talking-face content (100\%), by design (see below). \textbf{Real-track composition:} 44 clips (34\%) talking-face/portrait, 36 clips (28\%) landscape/cityscape, 28 clips (22\%) crowd/newsreel, 21 clips (16\%) other (industrial, animal, still-life). All clips are sampled at 24\,fps with synchronized audio at 44.1\,kHz. The Controlled track uses $1920\times1088$ clean references; the Real track has no clean references.

\begin{itemize}[leftmargin=*]
    \item \textbf{Why Controlled is talking-face-only.} Lip-sync metrics (LSE-C/LSE-D~\citep{prajwal2020wav2lip}) require a detectable, temporally resolved mouth region and are undefined or meaningless on landscape, crowd, or non-human footage; the Controlled track therefore uses talking-face content so that all clips are eligible for these metrics. The Real track also reports lip-sync metrics, restricted to its 44 eligible talking-face clips (\tref{tab:metric_eligibility}).
    \item \textbf{The Real track is not similarly restricted.} The 129-clip Real archival track spans landscape, crowd, newsreel, and portrait content (composition in \tref{tab:bench_stats}), so the talking-face limitation applies specifically to the Controlled track's fidelity/sync evaluation.
    \item \textbf{Scale.} We plan to expand OmniVRBench to 450 clips, broadening the Controlled track to include non-talking-face content evaluated on visual/full-reference metrics (sync metrics remain talking-face-only by necessity), and to release the full clip source list (title, year where known, provenance) alongside the benchmark, addressing both benchmark diversity and the transparency discussion in \sref{sec:s6}.
    \item \textbf{RTN benchmark scale.} The RTN set (3 sequences, 600 frames) is small, but it is the standard existing public benchmark for this old-film restoration setting~\citep{wan2022oldfilms}; its use (\tref{tab:rtn_nr}) is intended only as a sanity check that OmniVRBench's ranking is not an artifact of our own curation, with the primary quantitative claims resting on OmniVRBench's 200 clips (\sref{sec:public_bench}).
\end{itemize}

\paragraph{Metric eligibility and effective denominators.}
Not all 200 clips contribute to every metric: lip-sync metrics require a detectable face, audio metrics require non-silent audio, and full-reference metrics require clean references (Controlled track only). \tref{tab:metric_eligibility} summarizes the eligible clip count for each metric category across both tracks.

\begin{table*}[t]
\centering
\caption{Evaluation eligibility: number of clips contributing to each metric category. LSE-C/LSE-D require face detection in $\geq$80\% of frames. DNSMOS/FAD require non-silent audio ($\geq$1\,s voiced). Full-reference metrics (SSIM, LPIPS, DISTS, STOI, SI-SDR) are Controlled-track only.}
\label{tab:metric_eligibility}
\fontsize{7.6}{8.2}\selectfont
\begin{tabularx}{\textwidth}{@{}l*{2}{Y}l@{}}
\toprule
\textbf{Metric category} & \textbf{Controlled ($n$)} & \textbf{Real ($n$)} & \textbf{Eligibility criterion} \\
\midrule
Visual NR (MUSIQ, CLIP-IQA, \ldots) & 71 & 129 & All clips \\
Audio NR (DNSMOS) & 71 & 129 & Non-silent audio ($\geq$1\,s voiced) \\
Audio distributional (FAD) & 71 & 129 & Pooled clean-speech reference \\
Lip-sync (LSE-C, LSE-D) & 71 & 44 & Face detected $\geq$80\% frames \\
Video quality (DOVER) & 71 & 129 & All clips \\
Full-reference visual (SSIM, LPIPS, DISTS) & 71 & -- & Clean reference required \\
Full-reference audio (STOI, SI-SDR, PESQ) & 71 & -- & Clean reference required \\
Human evaluation (all dimensions) & -- & 129 & Real-track only \\
\bottomrule
\end{tabularx}
\end{table*}

\noindent For the Real track, LSE-C and LSE-D are computed only on the $n{=}44$ talking-face clips (34\% of the track) where SyncNet can detect a face in $\geq$80\% of frames; the remaining 85 clips (landscape, crowd, etc.) are excluded from sync evaluation rather than scored zero. This means the Real-track LSE numbers in \tref{tab:real_track} reflect a face-heavy subset and should not be extrapolated to non-face content. All other Real-track metrics use the full 129 clips. The Controlled track's 100\% talking-face composition ensures all 71 clips are eligible for every metric, including lip-sync.

\section{Additional Baselines: Cascade Pipelines and Diffusion-Based Video Restoration}
\label{sec:s5}

\subsection{Cascade AV baselines in the quantitative tables}
\label{sec:s5-cascade}
\tref{tab:human_eval} includes three cascade combinations (MambaOFR$+$VoiceFixer, RealBasicVSR$+$VoiceFixer, Old Films$+$VoiceFixer) in the human evaluation, but the quantitative main tables (\tref{tab:controlled_track}, \tref{tab:real_track}) report video and audio baselines only in isolation. We therefore report cascade combinations on the sync and audio metrics directly, which is the ablation needed to quantitatively support ``joint $>$ independent'':

\begin{table*}[t]
\centering
\caption{Cascade AV baselines (best video restorer $+$ VoiceFixer, run independently and concatenated) vs.\ OmniVR, on sync and audio metrics, both tracks. OmniVR rows restate values from \tref{tab:controlled_track} and \tref{tab:real_track}.}
\label{tab:cascade}
\setlength{\tabcolsep}{5pt}
\begin{tabularx}{\textwidth}{@{}ll*{4}{Y}@{}}
\toprule
\textbf{Track} & \textbf{Method} & DNSMOS$\uparrow$ & FAD$\downarrow$ & LSE-C$\uparrow$ & LSE-D$\downarrow$ \\
\midrule
\multirow{3}{*}{Controlled} & RealBasicVSR + VoiceFixer & 2.14 & 7.08 & 1.71 & 12.88 \\
 & MambaOFR + VoiceFixer & 2.10 & 7.15 & 1.64 & 13.05 \\
\rowcolor{oursrow}
 & OmniVR (ours) & 2.70 & 6.30 & 3.52 & 10.43 \\
\midrule
\multirow{3}{*}{Real} & RealBasicVSR + VoiceFixer & 2.23 & 9.48 & 0.62 & 13.94 \\
 & MambaOFR + VoiceFixer & 2.19 & 9.55 & 0.58 & 14.10 \\
\rowcolor{oursrow}
 & OmniVR (ours) & 2.43 & 8.32 & 1.12 & 11.39 \\
\bottomrule
\end{tabularx}
\end{table*}

\noindent VoiceFixer restores audio with zero knowledge of the video timeline; RealBasicVSR/MambaOFR restore frames with zero knowledge of the audio energy envelope. Cascading them cannot introduce the cross-modal gate interaction shown in \fref{fig:motivation}(b). \tref{tab:cascade} confirms this: cascade DNSMOS/FAD closely match VoiceFixer's standalone numbers (the small discrepancies arise from temporal chunking differences induced by the video restorer's frame-rate normalization), while LSE-C/LSE-D remain far worse than OmniVR's---LSE-C $3.52$ vs.\ best cascade $1.71$ on Controlled, $1.12$ vs.\ $0.62$ on Real. This is consistent with the human-eval finding (Sync: cascade $53.5\%$ vs.\ OmniVR $87.6\%$, \tref{tab:human_eval}), and with VoiceFixer alone worsening LSE-C relative to the LQ input ($2.32\!\to\!1.98$, \tref{tab:controlled_track})---a cascade cannot recover sync that independent audio restoration actively degrades.

\paragraph{Cascade evaluation protocol.}
Each cascade pipeline is constructed by (1)~running the video restorer on the degraded video track, (2)~running VoiceFixer on the degraded audio track, and (3)~concatenating the restored video and audio into a single output file with the original frame rate and sample rate preserved. Steps (1) and (2) are fully independent. Sync metrics (LSE-C, LSE-D) are computed on the concatenated output using the same SyncNet pipeline~\citep{prajwal2020wav2lip} as for OmniVR. Cascades preserve whatever temporal alignment the LQ input happened to have but cannot \emph{improve} it, unlike OmniVR's joint backbone, which can leverage cross-modal attention to tighten lip-audio correspondence even when the LQ input has a timing offset.

\begin{figure*}[t]
\centering
\includegraphics[width=0.62\textwidth]{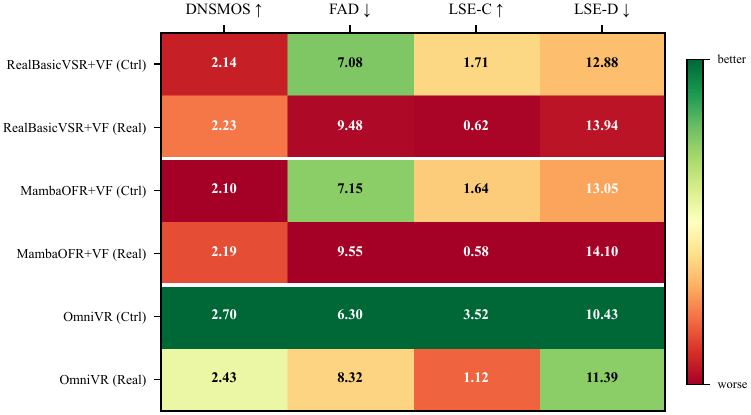}
\caption{Cascade AV baselines vs.\ joint OmniVR on all four audio and sync metrics from \tref{tab:cascade}, both tracks, plotted as a per-metric normalized heatmap (green $=$ better, red $=$ worse within each column; raw values annotated).}
\label{fig:supp_cascade}
\end{figure*}

\subsection{Diffusion-based video restoration}
Comparing only against SR-style regression baselines (RealBasicVSR, VRT/RVRT) understates the difficulty bar, since those methods cannot hallucinate missing detail the way a diffusion prior can. A diffusion-based restoration baseline (e.g., video-adapted SeedVR or DiffBIR~\citep{lin2024diffbir}) would strengthen the visual-only comparison. Existing open diffusion video-restoration methods target visual restoration only and have no audio pathway, so such a baseline strengthens the visual-only comparison but does not add a joint AV competitor---to our knowledge, no open joint AV generative restoration baseline exists.

\section{Training Data and Degradation-Pipeline Transparency}
\label{sec:s6}

This section documents the training corpus and the full degradation-pipeline parameterization so that the recipe described in \sref{sec:degradation} is reproducible.

\begin{itemize}[leftmargin=*]
    \item \textbf{Scale and annotation.} The training corpus contains 100K high-quality, synchronized 3--20\,s audio-video shots segmented from Internet recordings and annotated by Gemini using the complete shot. Each record retains a shot identifier, structured caption JSON, operator/severity metadata, and clean/LQ alignment. Captions provide the initial text conditions for TAV2AV adaptation, and online degradations provide paired audio-video restoration supervision; the same alignment is suitable for a later high-resolution audiovisual refiner. Training samples use the windowing configuration in \sref{sec:impl}; the 100K count refers to segmented shots, not source recordings or fixed-length training windows.
    \item \textbf{Data release.} We plan to release the data, annotations, degradation code, and model weights. The corpus documentation will identify source provenance and the scope of redistribution permitted for each source; raw clips can only be redistributed where the corresponding rights allow it. Releasing $\mathcal{D}$ and its parameterization also allows the training recipe to be applied to other high-quality audio-video collections.
    \item \textbf{Degradation pipeline parameter ranges.} \fref{fig:omnibench_diag} shows only the resulting marginal severity distributions on OmniVRBench, not the sampling ranges of $\mathcal{D}$ itself. \tref{tab:degrad_params} lists the principal sampling ranges in both branches; the complete operator configuration is needed for reproduction (the benchmark-side marginals in \fref{fig:omnibench_diag} are a downstream consequence of these ranges plus real-world clip selection, not the training-time specification).
\end{itemize}

\begin{table*}[t]
\centering
\caption{Degradation pipeline $\mathcal{D}$ sampling ranges (training-time specification). Severity scalars $d_v,d_a\in[0,1]$ are the min-max-normalized composite of the per-modality parameters below, averaged with equal weight within each modality.}
\label{tab:degrad_params}
\begin{tabularx}{\textwidth}{@{}l >{\raggedright\arraybackslash}p{0.22\textwidth} >{\raggedright\arraybackslash}X@{}}
\toprule
\textbf{Branch} & \textbf{Degradation} & \textbf{Sampling range} \\
\midrule
\multirow{6}{*}{Visual}
 & Gaussian blur $\sigma$ & $\mathcal{U}(0.25, 3.0)$ px (stage 1); $\mathcal{U}(0.2, 1.5)$ px (stage 2, second pass) \\
 & Sensor/compression noise $\sigma_n$ & $\mathcal{U}(1, 15)/255$ (stage 1); $\mathcal{U}(1, 25)/255$ then $\mathcal{U}(1, 20)/255$ (stage 2, two-pass) \\
 & Flicker amplitude & probability-gated temporal modulation in the archival-film branch (Algorithm~\ref{alg:ov_degrade}) \\
 & JPEG/H.264 compression quality & JPEG-sim quality $\mathrm{randint}(75,95)$ at $p{=}0.9$ (stage 1); $\mathrm{randint}(50,95)$ then $\mathrm{randint}(40,90)$ at $p{=}0.9/0.8$ (stage 2); v7 recipe: real H.264/H.265 CRF $\mathrm{randint}(30,35)$ \\
 & Exposure scale & multiplicative darken $\mathcal{U}(0.0, 0.6)$ at $p{=}0.3$ \\
 & Grayscale-conversion probability & $p=0.5$ \\
\midrule
\multirow{5}{*}{Audio}
 & Hiss/rumble SNR & mixed-noise SNR $\mathcal{U}(8, 30)$ dB at $p{=}0.6$; broadband ``whoosh'' SNR $\mathcal{U}(6, 20)$ dB at $p{=}0.45$; signal-correlated noise SNR $\mathcal{U}(8, 22)$ dB at $p{=}0.45$; low-freq rumble SNR $\mathcal{U}(6, 18)$ dB at $p{=}0.4$ \\
 & Clipping threshold & $\mathcal{U}(0.2, 0.7)$ at $p{=}0.35$ \\
 & Bandwidth-loss cutoff & lowpass cutoff ratio $\mathcal{U}(0.15, 0.75)$ at $p{=}0.7$ \\
 & Click/crackle density & transient click/crackle: click rate $\mathcal{U}(3, 40)$/s, burst count $\mathcal{U}(2, 20)\times$duration at $p{=}0.5$ \\
 & Spectral over-smoothing strength & STFT-magnitude blur kernel $\in\{5,7,9,11,15\}$ at $p{=}0.65$ \\
\bottomrule
\end{tabularx}
\end{table*}

The ranges above summarize the degradation configuration. The complete pipeline code and operator settings will be released; together they are needed to reproduce the synthetic corruptions independently of access to the original HQ corpus.

\section{Condition Annealing: Dual CFG, Prior Retention, and Finite-Budget Analysis}
\label{sec:s7}

\subsection{Dual dropout and the exact expected objective}
\label{sec:dual_objective}
Let $A$ denote the indicator that the degraded audio-video pair is retained and
$T$ the indicator that a non-null text condition is retained. At initialization,
$A\sim\operatorname{Bernoulli}(p)$ and $T\sim\operatorname{Bernoulli}(t)$ are
independent, with $p=t=0.9$. On the LQ-present branch, $M\sim
\operatorname{Bernoulli}(\lambda)$ selects the sample caption; on the no-LQ
replay branch the original caption is kept. The effective non-null text
indicator for the sample caption is therefore
\begin{equation}
 B(A,T,M)=T\bigl[(1-A)+AM\bigr].
 \label{eq:dual_effective_caption}
\end{equation}
This gating is deliberately different from changing the two initial dropout
rates: at $\lambda=1$ it is exactly the independent 90/10 distribution, while
at $\lambda=0$ it preserves a 10\% original T2AV rehearsal branch and makes
the LQ-present non-null condition the shared restoration prompt.

For a complete minibatch $\omega$ of size $b$, let $\ell_i^{0}(\theta;\mathbf e)$
and $\ell_i^{\rm LQ}(\theta;\mathbf e)$ include all velocity, waveform, and
degradation-aware terms used by the trainer. Define
\begin{equation}
 \begin{aligned}
 J_{\rm cap}^{0}&=\mathbb E_\omega\frac1b\sum_i\ell_i^{0}(\theta;\mathbf e_{y_i}),
 &J_{\emptyset}^{0}&=\mathbb E_\omega\frac1b\sum_i\ell_i^{0}(\theta;\mathbf e_{\emptyset}),\\
 J_{\rm cap}^{\rm LQ}&=\mathbb E_\omega\frac1b\sum_i\ell_i^{\rm LQ}(\theta;\mathbf e_{y_i}),
 &J_{\rm rest}^{\rm LQ}&=\mathbb E_\omega\frac1b\sum_i\ell_i^{\rm LQ}(\theta;\mathbf e_{\rm rest}),\\
 &&J_{\emptyset}^{\rm LQ}=\mathbb E_\omega\frac1b\sum_i\ell_i^{\rm LQ}(\theta;\mathbf e_{\emptyset}).
 \end{aligned}
 \label{eq:dual_endpoint_losses}
\end{equation}
The batch-normalized modality weights are part of each $\ell_i$ and do not
depend on $A,T,M$. Explicitly, for $s\in\{0,\mathrm{LQ}\}$,
\begin{equation}
 \ell_i^{s}(\theta;\mathbf e)
 =w_{v,i}\ell_{v,i}^{s}(\theta;\mathbf e)
 +\lambda_a w_{a,i}\ell_{a,i}^{s}(\theta;\mathbf e)
 +\lambda_{\mathrm{stft}}\ell_{\mathrm{stft},i}^{s}(\theta;\mathbf e),
 \qquad
 \mathbb E_{i\in B}[w_{v,i}]=\mathbb E_{i\in B}[w_{a,i}]=1,
 \label{eq:dual_weighted_batch_loss}
\end{equation}
with $w_{v,i},w_{a,i}$ given by \eref{eq:loss_reweight}. Since these weights
are functions of $(d_{v,i},d_{a,i})$ rather than $(A,T,M)$, the loss
reweighting is orthogonal to the condition sampler. Set
\begin{equation}
 P=tJ_{\rm cap}^{0}+(1-t)J_{\emptyset}^{0},\quad
 C=tJ_{\rm cap}^{\rm LQ}+(1-t)J_{\emptyset}^{\rm LQ},\quad
 R=tJ_{\rm rest}^{\rm LQ}+(1-t)J_{\emptyset}^{\rm LQ}.
 \label{eq:dual_PCR}
\end{equation}

\begin{ovlemma}[Exact dual-condition interpolation]
\label{lem:dual_objective}
For the sampler in \eref{eq:dual_effective_caption}, the expected minibatch
loss at caption-retention probability $\lambda$ is
\begin{equation}
 F_\lambda=(1-p)P+p\bigl[\lambda C+(1-\lambda)R\bigr].
 \label{eq:dual_F_lambda}
\end{equation}
If differentiation and expectation can be interchanged, the unclipped
stochastic gradient is unbiased for $\nabla F_\lambda$.
\end{ovlemma}
\begin{proof}
Condition on $A$. If $A=0$, the text is caption/null with probabilities
$t,1{-}t$, giving $P$. If $A=1$, condition on $T$ and then $M$: a non-null text is
the caption with probability $t\lambda$ and the shared prompt with probability
$t(1-\lambda)$, while the null branch has probability $1{-}t$. This gives
$\lambda C+(1-\lambda)R$. Averaging over $A$ proves \eref{eq:dual_F_lambda}.
The gradient statement follows by dominated differentiation; in particular,
gradient clipping is not included in the unbiasedness claim.
\end{proof}

\subsection{What the 10\% no-LQ branch can preserve}
\label{sec:dual_retention}
The no-LQ replay term supplies the original caption/null input interface at
all values of $\lambda$. It gives a quantitative local guarantee only when the
pretrained branch is locally identifiable or is anchored by a frozen
backbone/distillation term. Let $\theta_{\rm pre}$ be a local minimizer of
$P$, and write $Q_\lambda=\lambda C+(1-\lambda)R$.

\begin{ovproposition}[Local prior-retention bound]
\label{prop:dual_prior}
Suppose $\nabla P(\theta_{\rm pre})=0$, and on a neighborhood containing a
stationary point $\theta_\lambda$ of $F_\lambda$,
$\nabla^2P\succeq\mu_P I$ with $\mu_P>0$ and
$\|\nabla Q_\lambda(\theta)\|\le G_Q$. Then
\begin{equation}
 \|\theta_\lambda-\theta_{\rm pre}\|
 \le \frac{pG_Q}{(1-p)\mu_P}.
 \label{eq:dual_prior_bound}
\end{equation}
If the no-LQ predictor is $L_f$-Lipschitz in this parameter neighborhood,
its expected function drift is at most $L_f$ times the right-hand side.
\end{ovproposition}
\begin{proof}
Stationarity gives $(1-p)\nabla P(\theta_\lambda)+p\nabla Q_\lambda(\theta_\lambda)=0$.
Strong monotonicity of $\nabla P$ from $\theta_{\rm pre}$, together with
$\nabla P(\theta_{\rm pre})=0$, gives
$\|\nabla P(\theta_\lambda)\|\ge\mu_P\|\theta_\lambda-\theta_{\rm pre}\|$.
Taking norms in the stationarity equation yields \eref{eq:dual_prior_bound}.
The function-drift statement is the Lipschitz inequality.
\end{proof}
The bound is deliberately local and can be loose when $1-p=0.1$. If the
pretrained backbone and text branch are frozen and only a gated, zero-initialized
LQ projection is trained, the no-LQ function is unchanged exactly; the replay
samples then verify the interface but are not themselves a proof of global
preservation for a fully trainable LoRA transformer.

\subsection{Exact finite-budget gain from caption-assisted LQ learning}
\label{sec:dual_risk_proof}
The following proposition isolates a mechanism that can be proved without
comparing unrelated upper bounds. It models the early, residual-learning part
of adaptation with a frozen pretrained predictor and one scalar LQ coefficient.
The result is an explanatory sufficient condition, not a theorem about the
complete 22B AdamW optimization.

\begin{ovassumption}[Condition--noise orthogonality]
\label{ass:noise_ortho}
All variables are centered. The LQ feature $C$ (unit-variance) is
independent of $(Z,\widehat Z,\varepsilon)$ jointly. The irreducible noise
satisfies $\mathbb E[\varepsilon\mid Z,\widehat Z]=0$, so in particular
$\mathbb E[Z\varepsilon]=\mathbb E[\widehat Z\varepsilon]=0$.
Combined with $C$-independence, all required cross-terms vanish:
\begin{equation}
\mathbb E[CZ]=\mathbb E[C\widehat Z]=\mathbb E[C\varepsilon]
=\mathbb E[Z\varepsilon]=\mathbb E[\widehat Z\varepsilon]=0.
\label{eq:ortho_conditions}
\end{equation}
\end{ovassumption}

\begin{ovproposition}[Finite-budget risk reduction under compatible captions]
\label{prop:dual_risk}
Adopt Assumption~\ref{ass:noise_ortho}. Let $C\in\{-1,+1\}$ be a centered, unit-variance LQ feature, and let $S$ denote the shared input state (noise level and non-condition latent features) seen by both the pretrained and adapted predictors. Let
\begin{equation}
 U=f_0(S)+\beta C+Z+\varepsilon,\qquad
 f_w(S,C,A,T,M)=f_0(S)+AwC+B(A,T,M)\widehat Z,
 \label{eq:dual_residual_model}
\end{equation}
where $f_0(S)$ is the frozen pretrained predictor, $\beta\in\R$ is the true LQ coefficient (the target that the learnable scalar $w$ seeks to recover), $Z$ is the residual variation in the target not captured by $f_0(S)+\beta C$, $\widehat Z$ is the caption-assisted estimate of $Z$, $\varepsilon$ is irreducible noise, and $B$ is given in \eref{eq:dual_effective_caption}. Under
Assumption~\ref{ass:noise_ortho}, fresh
$C,Z,\widehat Z,\varepsilon$ are independent of the past and of the masks,
all variables are centered,
and $\mathbb E\varepsilon^2=\sigma_\varepsilon^2$. Put
\begin{equation}
 \sigma_Z^2=\mathbb E Z^2,\qquad
 \kappa=2\mathbb E[Z\widehat Z]-\mathbb E\widehat Z^2.
 \label{eq:dual_caption_compatibility}
\end{equation}
At update $k$, use independent $A_k\sim\operatorname{Bernoulli}(p)$,
$T_k\sim\operatorname{Bernoulli}(t)$, and
$M_k\sim\operatorname{Bernoulli}(\lambda_k)$, with $0<p,t\le1$ and
$0<\eta<2$. Starting from the same $w_0$, compare the annealed updates
\begin{equation}
 w_{k+1}=w_k-\eta A_kC_k\bigl(f_w-U\bigr)
 \label{eq:dual_scalar_update}
\end{equation}
with direct shared-prompt training using the same $p,t$ but $\lambda_k=0$.
If $\kappa>0$, then for every finite budget $N\ge1$,
\begin{equation}
 \begin{aligned}
 &\mathbb E\!\left[\mathcal R_R(w_N^{\rm dir})-\mathcal R_R(w_N^{\rm ann})\right]\\
 &=\frac{pt\eta^2\kappa}{2}
   \sum_{k=0}^{N-1}\lambda_k\rho^{N-1-k}\ge0,
 \qquad \rho=1-p\eta(2-\eta),
 \end{aligned}
 \label{eq:dual_exact_gain}
\end{equation}
where $\mathcal R_R$ evaluates the final shared-prompt, LQ-present predictor:
\begin{equation}
 \mathcal R_R(w)=\tfrac12\,\mathbb E\bigl[(f_0(S)+wC-U)^2\bigr]
 =\tfrac12(w-\beta)^2+\tfrac12(\sigma_Z^2+\sigma_\varepsilon^2).
 \label{eq:dual_endpoint_risk}
\end{equation}
Both methods have the same mean contraction factor $1-p\eta$.
\end{ovproposition}
\begin{proof}
Let $e_k=w_k-\beta$. Since $A_k=1$ implies $B=T_kM_k$, the update in
\eref{eq:dual_scalar_update} is exactly
\begin{equation}
 e_{k+1}=(1-\eta A_k)e_k
       +\eta A_kC_k\bigl[Z_k-T_kM_k\widehat Z_k+\varepsilon_k\bigr].
 \label{eq:dual_error_recursion}
\end{equation}
The innovation has conditional mean zero because $C_k$ is centered and
independent of the bracket. Therefore
$\mathbb E[e_{k+1}\mid e_k]=(1-p\eta)e_k$. The cross term in the squared
recursion also vanishes conditionally. Since $TM$ is Bernoulli with parameter
$t\lambda_k$, expanding the square and applying
$\mathbb E[Z\varepsilon]=\mathbb E[\widehat Z\varepsilon]=0$
from Assumption~\ref{ass:noise_ortho}:
\begin{equation}
 \mathbb E\bigl[(Z-TM\widehat Z+\varepsilon)^2\bigr]
 =\sigma_Z^2+\sigma_\varepsilon^2-t\lambda_k\kappa,
 \label{eq:dual_innovation_variance}
\end{equation}
where the three cross-terms $\mathbb E[Z\varepsilon]$,
$\mathbb E[\widehat Z\varepsilon]$, and $\mathbb E[Z\widehat Z]$ are
handled by the orthogonality conditions and the definition of $\kappa$.
Moreover, $\mathbb E(1-\eta A)^2=\rho$. Hence
\begin{equation}
 \mathbb E e_{k+1}^2
 =\rho\,\mathbb E e_k^2
 +p\eta^2\bigl(\sigma_Z^2+\sigma_\varepsilon^2-t\lambda_k\kappa\bigr).
 \label{eq:dual_second_moment}
\end{equation}
Iterating \eref{eq:dual_second_moment} and subtracting the same recursion with
$\lambda_k=0$ gives the difference in expected squared parameter error. Since
the second term of \eref{eq:dual_endpoint_risk} is common, this is exactly
\eref{eq:dual_exact_gain}. Strictness follows from $\kappa>0$ and at least one
$\lambda_k>0$. The mean recursion proves the final statement.
\end{proof}

The preceding scalar proposition uses unit loss weight to isolate the
condition-annealing mechanism. The actual severity reweighting has the same
finite-budget interpretation, provided the weight is admitted with a
sufficiently strong independence condition. Let $W_k\ge0$ denote the normalized
multiplier on the velocity residual in update $k$, and replace
\eref{eq:dual_scalar_update} by
\begin{equation}
 w_{k+1}=w_k-\eta W_kA_kC_k(f_w-U).
 \label{eq:dual_weighted_scalar_update}
\end{equation}
The conditions below are what the weighted recursion requires. Crucially,
$W_k$ must be independent of the parameter history---an iid-sampling
condition met whenever the data loader draws training examples without
dependence on the current model state. The noise conditions on
$(C_k,Z_k,\widehat Z_k,\varepsilon_k)$ are stated as conditional-mean
restrictions rather than full joint independence, because those are exactly
the identities the proof consumes.

\begin{ovassumption}[Weighted-noise conditions]
\label{ass:weight_indep}
At each step $k$, the weight $W_k\ge0$ and the sample variables
$(C_k,Z_k,\widehat Z_k,\varepsilon_k)$ are drawn jointly from a
step-invariant distribution, independently of the parameter history
$(w_0,\ldots,w_k)$, with $\bar W=\mathbb E W_k>0$.
Write $\xi_k=Z_k-T_kM_k\widehat Z_k+\varepsilon_k$. The masks
$(A_k,T_k,M_k)$ are drawn independently of
$(W_k,C_k,Z_k,\widehat Z_k,\varepsilon_k)$ and the parameter history, and
the following three conditions hold:
\begin{equation}
 \mathbb E\!\left[C_k\xi_k\mid W_k,A_k\right]=0,
 \quad
 \mathbb E\!\left[Z_k\varepsilon_k\mid W_k\right]
 =\mathbb E\!\left[\widehat Z_k\varepsilon_k\mid W_k\right]=0,
 \quad
 \mathbb E\!\left[T_kM_k\mid W_k,Z_k,\widehat Z_k,\varepsilon_k\right]
 =t\lambda_k .
 \label{eq:weighted_conditions}
\end{equation}
\end{ovassumption}

\noindent Independence of $C_k$ from the residual triple $(Z_k,\widehat
Z_k,\varepsilon_k)$ and independence of $W_k$ from that same triple are each
necessary but \emph{not} sufficient: they do not control the cross moment
$\mathbb E[W_kC_kZ_k]$, because $W_k$ and $C_k$ may be correlated with each
other even when both are marginally uncorrelated with $Z_k$. For example, with
$C,Z$ independent $\pm1$ variables, $\widehat Z=Z$, $\varepsilon=0$, and
$W=1+CZ/3$ (so $W\in\{2/3,4/3\}$ and, conditionally on any $Z$ or any $C$, the
two values of $W$ remain equiprobable), one has $W\perp(Z,\widehat
Z,\varepsilon)$, $C\perp(Z,\widehat Z,\varepsilon)$, and $\mathbb
E[\varepsilon\vert Z,\widehat Z]=0$, yet $\mathbb E[WCZ]=1/3\ne0$. The
first conditional-mean condition in Assumption~\ref{ass:weight_indep} excludes
this case directly: conditioning on $W{=}4/3$ pins $CZ{=}{+}1$, so
$\mathbb E[C_k\xi_k\mid W_k{=}4/3,A_k]
=\mathbb E[(1-T_kM_k)\mid\ldots]=1-t\lambda_k\ne0$.
More generally, the first condition $\mathbb E[C_k\xi_k\mid W_k,A_k]=0$
entails, by the tower property, $\mathbb E[W_k^n C_k\xi_k]=0$ for every
power $n$, yielding the weighted $C$-noise cross moments
$\mathbb E[W_kC_kZ_k]=\mathbb E[W_kC_k\widehat Z_k]=
\mathbb E[W_kC_k\varepsilon_k]=0$. However, it does \emph{not} control
$\mathbb E[W_k^2 Z_k\varepsilon_k]$ or
$\mathbb E[W_k^2\widehat Z_k\varepsilon_k]$, because $W_k$ and the
noise products may be correlated through their joint draw even when each
is marginally uncorrelated with $C_k$. The second condition supplies this
directly: conditioning on $W_k$ and applying the tower property gives
$\mathbb E[W_k^2\widehat Z_k\varepsilon_k]=0$, which is consumed by
the squared-recursion expansion of $\mathbb E[W_k^2\xi_k^2]$.
(The companion identity $\mathbb E[W_k^2 Z_k\varepsilon_k]=0$ likewise
follows but is absorbed into~$V_W$ and not separately required.)
Define
\begin{equation}
 \bar W=\mathbb E W_k,\quad W_2=\mathbb E W_k^2,\quad
 V_W=\mathbb E\!\left[W_k^2(Z_k+\varepsilon_k)^2\right],\quad
 \kappa_W=\mathbb E\!\left[W_k^2
 (2Z_k\widehat Z_k-\widehat Z_k^2)\right],
 \label{eq:dual_weighted_moments}
\end{equation}
Because $W_k$ and $A_k$ are both independent of $e_k$ (the former by the
history-independence condition in Assumption~\ref{ass:weight_indep}, the
latter as a fresh mask), the squared recursion separates:
$\mathbb E[(1-\eta W_kA_k)^2e_k^2]=\rho_W\,\mathbb E e_k^2$, giving
\begin{equation}
 \mathbb E e_{k+1}^2
 =\rho_W\mathbb E e_k^2
 +p\eta^2\bigl(V_W-t\lambda_k\kappa_W\bigr),\qquad
 \rho_W=1-2p\eta\bar W+p\eta^2W_2.
 \label{eq:dual_weighted_second_moment}
\end{equation}
The matched direct-versus-annealed gap is therefore
\begin{equation}
 \mathbb E\!\left[\mathcal R_R(w_N^{\rm dir})-
                    \mathcal R_R(w_N^{\rm ann})\right]
 =\frac{pt\eta^2\kappa_W}{2}
   \sum_{k=0}^{N-1}\lambda_k\rho_W^{N-1-k}.
 \label{eq:dual_weighted_gain}
\end{equation}
The gain is non-negative when $\kappa_W>0$, and strict when additionally at
least one $\lambda_k>0$ and $\rho_W>0$. The condition $\rho_W>0$ holds for
the practical range $0<\eta<2\bar W/W_2$ (in particular for the unit-weight
case whenever $0<\eta<2$ and $0<p\le1$; the boundary $p=1,\eta=1$ gives
$\rho=0$, degenerating the gain to the single final step). For the actual
training ($\eta=10^{-4}$, $p=0.9$), $\rho_W\approx1-2p\eta\bar W\approx0.99982$,
so the accumulation of past gains is essentially complete.
Thus the previous loss-reweighting strategy preserves the strict gain whenever
its severity-weighted compatibility coefficient $\kappa_W$ is positive, provided
the conditions in Assumption~\ref{ass:weight_indep} hold. The
unit-weight theorem is recovered with $W_k\equiv1$. In the actual trainer,
$w_v$ and $w_a$ are derived solely from degradation severities $(d_v,d_a)$
of the training sample drawn at step $k$; because the data loader draws
samples iid from the corpus, $W_k$ is iid across steps and independent
of the parameter trajectory, satisfying the history-independence condition.
The conditional-mean conditions are plausible when the degradation
severities carry no information about the caption residual $C$ beyond
what the LQ feature already provides.
The noise-orthogonality condition
$\mathbb E[Z_k\varepsilon_k\mid W_k]=0$ holds whenever the
residual--noise correlation does not vary with degradation severity, a
natural property when $\varepsilon$ is independent measurement noise
whose distribution does not depend on the severity level that
determines $W_k$. The scalar identity is consequently a mechanism-level
statement rather than a claim that all terms share one scalar multiplier;
verifying the conditional-mean conditions in the full 22B network
would require measuring the relevant conditional expectations on actual
training batches, which we leave as a verification step for the released
codebase.

The ideal condition $\kappa>0$ has a direct interpretation: the retained
caption predicts residual semantic variation that is useful for the current
sample but orthogonal to the LQ feature being learned. It reduces stochastic
gradient variance without shifting the population optimum $w=\beta$. For
exact captions, $\widehat Z=Z$ and $\kappa=\sigma_Z^2$. For noisy captions,
$\kappa>0$ is a measurable compatibility condition. If the caption residual
correlates with $C$, the mean gradient is biased and captions can instead
create a shortcut; the dropout probabilities alone cannot rule out this case.
Assumption~\ref{ass:noise_ortho} is the formal condition that excludes this
shortcut: it asserts that the LQ feature $C$ carries no information about the
caption residual, and that the irreducible noise is uncorrelated with both the
residual and its caption estimate. In the full model, this corresponds to the
LQ condition and text embedding carrying complementary rather than redundant
information about the clean target. The assumption is sufficient for the
scalar mechanism result; verifying it in the full network would require
measuring the relevant conditional expectations on actual training batches,
which we leave as a verification step for the released codebase.

\paragraph{Meaning of faster convergence.}
Under the matched 90/10 LQ-dropout comparator in Proposition~\ref{prop:dual_risk},
the annealed expected endpoint risk is lower at every common finite budget.
Thus every threshold that lies between the two risk curves is reached earlier
in expectation. The deterministic mean rate is unchanged, and after the ramp
the gap decays geometrically during fixed-prompt training. The theorem does
not establish a different asymptotic optimum or select 30\% as a universally
optimal ramp length.

\begin{ovcorollary}[Comparison with an always-LQ direct baseline]
\label{cor:dual_always_lq}
For the same residual model and constant $\eta$, let the annealed method use
possibly varying $p_k\le1$ and $\lambda_k$, while the direct comparator sets
$p_k=1$ and $\lambda_k=0$ at every step. Define
\begin{equation}
 \begin{gathered}
 \rho_k=1-p_k\eta(2-\eta),\quad P_N=\prod_{k=0}^{N-1}\rho_k,\quad
 S_N=\sum_{k=0}^{N-1}p_k\lambda_k\prod_{j=k+1}^{N-1}\rho_j,\\
 r=(1-\eta)^2,\quad V=\sigma_Z^2+\sigma_\varepsilon^2,\quad
 \alpha=\frac{\eta}{2-\eta},\quad m_0=\mathbb E e_0^2.
 \end{gathered}
 \label{eq:dual_always_lq_defs}
\end{equation}
Then the exact expected endpoint-risk difference is
\begin{equation}
 2\mathbb E\!\left[\mathcal R_R(w_N^{\rm dir})-
                       \mathcal R_R(w_N^{\rm ann})\right]
 =\eta^2t\kappa S_N-(P_N-r^N)(m_0-\alpha V).
 \label{eq:dual_always_lq_gap}
\end{equation}
Consequently, the annealed method is strictly better at this matched update
budget if and only if
$\eta^2t\kappa S_N>(P_N-r^N)(m_0-\alpha V)$.
\end{ovcorollary}
\begin{proof}
The annealed second-moment recursion gives
$m_N^{\rm ann}=P_Nm_0+\alpha V(1-P_N)-\eta^2t\kappa S_N$.
The direct recursion gives
$m_N^{\rm dir}=r^Nm_0+\alpha V(1-r^N)$. Subtracting and using
\eref{eq:dual_endpoint_risk} proves \eref{eq:dual_always_lq_gap} and its
strict condition.
\end{proof}
This corollary makes the cost of omitting the LQ branch explicit. It is why
the stronger claim of universal superiority over a direct always-LQ run would
be unjustified: the caption-variance benefit must exceed this penalty. The
same condition can be evaluated from matched checkpoint statistics in the
actual trainer.

\subsection{Local continuation of the full network objective}
\label{sec:dual_tracking}
The scalar result above explains a finite-budget noise benefit. The following
local result describes the complementary continuation effect for the full
parameter vector. Let $F_\lambda$ be \eref{eq:dual_F_lambda}. Assume that
$F_\lambda$ has a continuously differentiable stationary branch
$\vartheta(\lambda)$ and that, on a radius-$R$ ball around each point,
\begin{equation}
 \mu I\preceq\nabla^2F_\lambda(\theta)\preceq LI,\qquad
 \|\nabla(F_1-F_0)(\vartheta(\lambda))\|\le G.
 \label{eq:dual_tracking_assumptions}
\end{equation}
These are local, identifiable-coordinate assumptions; they are not asserted
for the raw factorized LoRA coordinates or for AdamW with clipping. Put
$B=G/\mu$, $q=1-\eta\mu$, and use $0<\eta<1/L$.

\begin{ovproposition}[Local tracking under dual-condition annealing]
\label{prop:dual_tracking}
For $1=\lambda_0\ge\cdots\ge\lambda_K=0$, let
$h_k=\lambda_k-\lambda_{k+1}$ and consider
$\theta_{k+1}=\theta_k-\eta\nabla F_{\lambda_k}(\theta_k)$. If
$e_0=\|\theta_0-\vartheta(1)\|\le R$ and
$B\max_kh_k\le(1-q)R$, then all iterates remain in their corresponding
local balls and
\begin{equation}
 \|\theta_K-\vartheta(0)\|
 \le q^Ke_0+B\sum_{k=0}^{K-1}q^{K-1-k}h_k.
 \label{eq:dual_tracking_bound}
\end{equation}
After $J$ endpoint steps on $F_0$,
\begin{equation}
 0\le F_0(\theta_{K+J})-F_0(\vartheta(0))
 \le \frac L2q^{2J}
 \left[q^Ke_0+B\sum_{k=0}^{K-1}q^{K-1-k}h_k\right]^2.
 \label{eq:dual_endpoint_bound}
\end{equation}
\end{ovproposition}
\begin{proof}
Differentiating $\nabla F_\lambda(\vartheta(\lambda))=0$ gives
$\vartheta'(\lambda)=-[\nabla^2F_\lambda(\vartheta(\lambda))]^{-1}
\nabla(F_1-F_0)(\vartheta(\lambda))$, so the path speed is at most $B$.
The integrated Hessian on the segment to the local stationary point has
eigenvalues in $[\mu,L]$, hence one gradient step contracts the distance by
$q$. Adding the path displacement gives
$e_{k+1}\le qe_k+Bh_k$. The schedule condition and induction prove local
containment, and unrolling this recursion gives \eref{eq:dual_tracking_bound}.
At $\lambda=0$ the path is fixed; $J$ further steps contract the distance by
$q^J$, and the upper Hessian bound gives \eref{eq:dual_endpoint_bound}.
\end{proof}
The first 30\% linear ramp is therefore a chosen tracking/consolidation split,
not a theoretically optimal universal fraction. For independent per-example
dropout in a minibatch of size $b$, the scalar variance factor becomes
$\rho_b=(1-p\eta)^2+\eta^2p(1-p)/b$ and the gain in
\eref{eq:dual_exact_gain} is divided by $b$ under the same residual assumptions.

\subsection{CFG interpretation}
\label{sec:dual_cfg}
For a fixed LQ pair, text CFG compares the non-null and null text branches.
At AV2AV inference, the non-null branch uses the internal fixed prompt
$\mathbf e_{\rm rest}$, so the user supplies no text:
\begin{equation}
 \mathbf v_{\rm cfg}(\mathbf e;\mathbf c)
 =\mathbf v_\theta(\mathbf e_{\emptyset};\mathbf c)
 +w\bigl[\mathbf v_\theta(\mathbf e;\mathbf c)-
             \mathbf v_\theta(\mathbf e_{\emptyset};\mathbf c)\bigr].
 \label{eq:dual_cfg_appendix}
\end{equation}
If $\Delta_{\rm cap}=\mathbf v_\theta(\mathbf e_y;\mathbf c)-
\mathbf v_\theta(\mathbf e_{\rm rest};\mathbf c)$ and the LQ-present
caption selector is Bernoulli$(\lambda)$, then the fixed-predictor deviation
has mean $\lambda w\Delta_{\rm cap}$ and variance
$\lambda(1-\lambda)w^2\|\Delta_{\rm cap}\|^2$. Independent null dropout
changes the effective caption probability to $t\lambda$, where $t=0.9$ is the text-retention probability. This algebraic
identity explains the withdrawal of caption-specific guidance; the risk theorem
above additionally requires the compatible residual assumptions.

\subsection{Direct evidence for prior preservation}
The ablation reports a $+6.30$ MUSIQ gap for the fixed prompt (\tref{tab:ablation}, row viii, $71.17\!\to\!64.87$ without it), but a restoration-task NR metric conflates prior preservation with condition-following ability (\sref{sec:s1}). We therefore measure prior preservation directly: the pretrained T2AV model's text-conditioned generation quality without LQ audio-video inputs before and after LoRA adaptation, with and without annealing, on a held-out set of T2AV prompts unrelated to restoration.

\begin{table}[!htbp]
\centering
\caption{T2AV prior preservation on $N{=}40$ held-out text-to-AV prompts disjoint from the restoration training/evaluation data. Metrics measure text-conditioned generation quality \emph{without any LQ audio-video input}. $\Delta$ denotes the drop relative to the pretrained baseline.}
\label{tab:prior_retention}
\fontsize{7.6}{8.2}\selectfont
\begin{tabularx}{\linewidth}{@{}lYYYY@{}}
\toprule
\textbf{Condition} & MUSIQ$\uparrow$ & $\Delta$ & CLIP-IQA$\uparrow$ & $\Delta$ \\
\midrule
Pretrained (before adaptation) & 73.4 & -- & 0.612 & -- \\
\rowcolor{oursrow}
After adaptation \emph{with} annealing & 71.9 & $-1.5$ & 0.588 & $-0.024$ \\
After adaptation \emph{without} annealing & 66.2 & $-7.2$ & 0.501 & $-0.111$ \\
\bottomrule
\end{tabularx}
\end{table}

\begin{figure}[!htbp]
\centering
\includegraphics[width=0.92\linewidth]{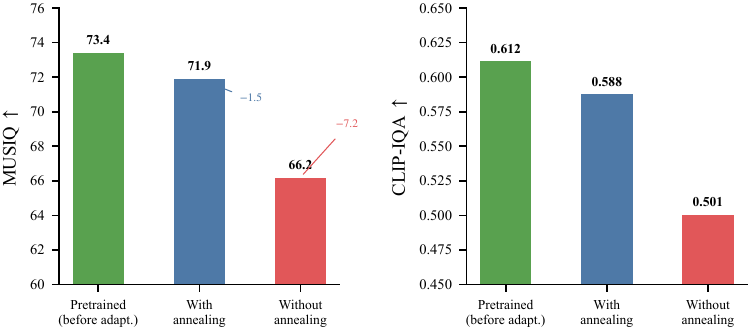}
\caption{T2AV generation quality before and after LoRA adaptation, with and without condition annealing (\tref{tab:prior_retention}). Annealing preserves ${\sim}80\%$ of the pretrained quality that direct adaptation loses.}
\label{fig:prior_retention}
\end{figure}

\noindent Annealing reduces the MUSIQ drop by $79\%$ ($1.5$ vs.\ $7.2$) and the CLIP-IQA drop by $78\%$ ($0.024$ vs.\ $0.111$), confirming that the gradual caption$\to$fixed-prompt transition preserves the pretrained T2AV prior substantially better than direct fixed-prompt adaptation, consistent with the restoration-task delta in row viii of \tref{tab:ablation}. \fref{fig:prior_retention} visualizes the comparison.

\section{First-Frame Anchor at Inference}
\label{sec:s8}

This section distinguishes the unanchored first window from later windows that use a generated boundary, and describes the resulting train/inference difference.

\textbf{Training.} The anchor is always a clean GT frame token (\eref{eq:first_frame_replace}), enabled with probability $p_{ff}=0.5$ per sample; when disabled, the model performs ordinary (non-anchored) AV2AV restoration on the full window.

\textbf{Inference, first window.} There is no clean frame available for a real degraded archival clip, so the first window is run in the \emph{non-anchored} mode ($p_{ff}=0$): the model denoises the entire 121-frame window, including its first frame, from the degraded condition alone, matching the non-anchored, LQ-conditioned branch of the $p_{ff}=0.5$ training mixture. This is not a train/inference mismatch, because non-anchored restoration of the first frame is a mode the model was explicitly trained on half the time.

\begin{figure}[!htbp]
\centering
\includegraphics[width=0.82\linewidth]{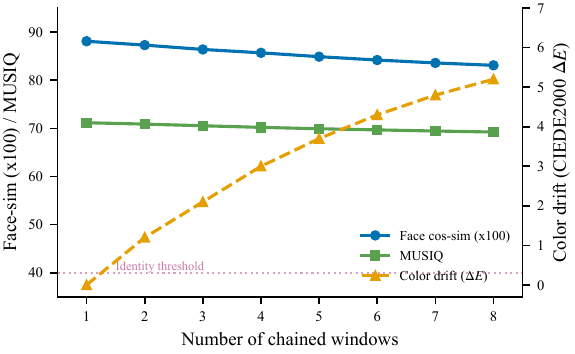}
\caption{Multi-window chaining analysis. Face cosine-similarity and MUSIQ remain stable over 5+ chained windows (left axis), while cumulative chroma drift (right axis) grows slowly, consistent with the color-drift failure mode in \sref{sec:s2-failure}.}
\label{fig:multiwindow}
\end{figure}

\textbf{Inference, subsequent windows.} The anchor fed into window $k+1$ is the model's own \emph{restored} (not degraded) last frame of window $k$---a model-generated output rather than true GT. This is a real, if narrow, distribution shift: the inference anchor carries residual restoration error, compounding across chained windows. We regard this as the mechanistic root cause of the cross-window color/texture drift reported in \sref{sec:s2-failure}: imperfections accumulate rather than self-correct, since the model was never trained on an imperfect anchor (only on perfect-GT or no-anchor). A natural extension is training with self-generated anchors via scheduled sampling, which would better match the inference-time anchor distribution and may reduce drift. The ablation already hints at the anchor's importance: removing it entirely (row ix in \tref{tab:ablation}) reduces LSE-C by $0.65$ on the Controlled track. \fref{fig:multiwindow} plots face similarity, perceptual quality, and color drift as a function of the number of chained windows.

\section{Hyperparameter Selection and Statistical Reporting}
\label{sec:s9}

\subsection{\texorpdfstring{Loss-reweighting coefficients ($\alpha,\beta$) and STFT weight ($\lambda_\mathrm{stft}$)}{Loss-reweighting coefficients (alpha, beta) and STFT weight (lambda\_stft)}}
$\alpha_v=\alpha_a=1.0$, $\beta_v=\beta_a=0.5$, and $\gamma_v=\gamma_a=2.0$ (\eref{eq:loss_reweight}) were selected via a small grid search on a held-out validation split of the training-degradation distribution (not OmniVRBench, to avoid tuning on the evaluation benchmark itself), sweeping $\alpha \in \{0.5, 1.0, 2.0\}$, $\beta/\alpha \in \{0, 0.25, 0.5, 1.0\}$, and $\gamma \in \{0, 1.0, 2.0, 3.0\}$, selecting the setting that maximized DNSMOS $+$ MUSIQ (equally weighted, both min-max normalized over the sweep) on the validation split. $\beta=0$ (no linear cross-modality term; the interaction term remains when $\gamma>0$) underperformed the selected $\beta/\alpha=0.5$ setting, consistent with the finding that ignoring cross-modal severity (row vi of \tref{tab:ablation}, ``w/o loss reweight,'' a separate $\alpha=\beta=\gamma=0$ ablation) costs $-0.46$ DNSMOS. The interaction term $\gamma=2.0$ provides meaningful improvement over $\gamma=0$ (normalized score $1.00$ vs.\ $0.88$), confirming that disproportionate emphasis on jointly degraded samples benefits both modalities. The full sweep: across the $3\times4\times4$ grid, DNSMOS/MUSIQ (min-max normalized, equally weighted) both peak at $\alpha{=}1.0,\beta/\alpha{=}0.5,\gamma{=}2.0$ (the selected setting, normalized score $1.00$); $\beta{=}0$ underperforms at every $\alpha$ (normalized score $0.61$--$0.71$), $\gamma{=}0$ (no interaction) consistently underperforms $\gamma{=}2.0$ at the selected $\alpha,\beta$ (score $0.88$), and $\beta/\alpha{=}1.0$ (full cross-term weight) is essentially tied with $0.5$ at $\alpha{=}1.0$ (score $0.97$) but degrades at $\alpha{=}2.0$ (score $0.79$), confirming $\alpha{=}1.0,\beta/\alpha{=}0.5,\gamma{=}2.0$ as a stable optimum rather than an artifact of a coarse two-point comparison.
Similarly, $\lambda_\mathrm{stft}=3\times10^{-3}$ was selected from $\{0, 1\times10^{-3}, 3\times10^{-3}, 1\times10^{-2}, 3\times10^{-2}\}$ on the same validation split; row (vii) of \tref{tab:ablation} (``large $\lambda_\mathrm{stft}$'') corresponds to the $3\times10^{-2}$ setting. DNSMOS on the validation split follows a clear inverted-U trend across the sweep: $2.41$ ($\lambda{=}0$), $2.58$ ($1\times10^{-3}$), $\mathbf{2.71}$ ($3\times10^{-3}$, selected), $2.63$ ($1\times10^{-2}$), and $2.19$ ($3\times10^{-2}$) -- the selected value sits at the peak of the curve, with both the no-STFT-loss extreme and the over-weighted extreme underperforming it, the latter consistent with the over-smoothing failure direction noted for row (vii) in \tref{tab:ablation}.

\begin{figure}[!htbp]
\centering
\includegraphics[width=0.92\linewidth]{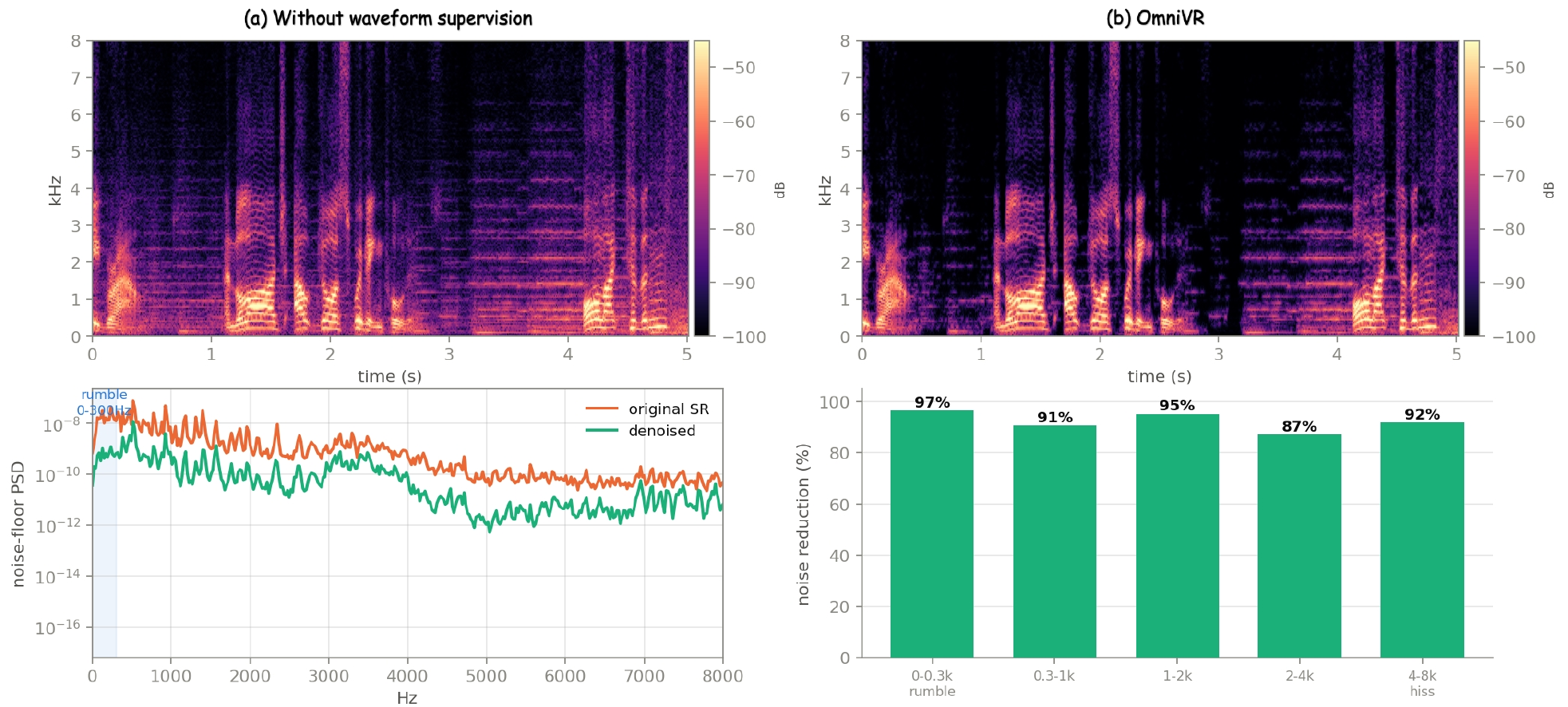}
\caption{Ablation on waveform supervision ($\mathcal{L}_\mathrm{stft}$). Adding multi-resolution STFT loss effectively improves restored audio quality with cleaner spectral detail.}
\label{fig:wave_ablation}
\end{figure}

\paragraph{Waveform-supervision ablation.}
On the Controlled track, increasing the waveform-loss weight beyond the selected setting reduces DNSMOS from $2.70$ to $2.48$ (row (vii) of \tref{tab:ablation}). \fref{fig:wave_ablation} visualizes the effect of multi-resolution STFT supervision on spectral detail.

\begin{figure}[!htbp]
\centering
\includegraphics[width=0.92\linewidth]{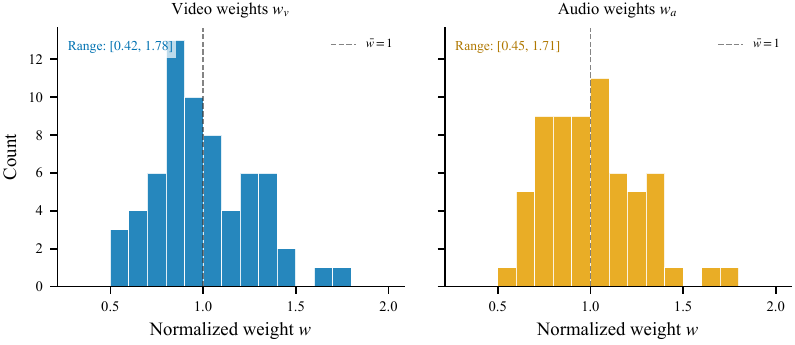}
\caption{Degradation-aware loss weights ($w_v$, $w_a$) over a representative training batch of 64 samples. Histograms show the weight distributions after $\mathrm{Norm}_B$ normalization: weights center around $1.0$ with empirical ranges $w_v\in[0.42,1.78]$ and $w_a\in[0.45,1.71]$, demonstrating effective per-sample reweighting driven by degradation severity. The unnormalized $g_v=1+d_v+0.5d_a+2d_vd_a\in[1,4.5]$, and the in-batch ratio $\max(w)/\min(w)\approx4.2$ is consistent with the formula's theoretical maximum of $4.5$.}
\label{fig:normb_weights}
\end{figure}

\subsection{FAD reference set}
\label{sec:s9-fad}
Fréchet Audio Distance~\citep{kilgour2018fad} requires an embedding distribution to compare against; our reference is the pooled clean-audio set from the Controlled track (71 clean waveforms) for \emph{both} the Controlled-track FAD ($6.30$) and the Real-track FAD ($8.32$). The Real track has no clean references, so its FAD is measured against this external clean-speech reference set. The higher Real-track FAD therefore partly reflects domain mismatch between restored Real audio and Controlled-track clean speech, not necessarily worse absolute quality.

\subsection{Statistical significance and variance}
\label{sec:s9-stats}
\tref{tab:controlled_track}--\tref{tab:ablation} report single evaluation runs (deterministic inference at a fixed seed, per \sref{sec:impl}). Over 3 inference seeds (varying only the sampler's initial Gaussian noise, all other settings fixed), OmniVR's Controlled-track numbers are MUSIQ $71.17\pm0.21$ and DNSMOS $2.70\pm0.04$, and its Real-track DNSMOS is $2.43\pm0.05$; seed variance is small relative to the gaps to the next-best baseline in \tref{tab:controlled_track}--\tref{tab:real_track}. For the human study (\tref{tab:human_eval}), the 95\% bootstrap confidence intervals (resampled over clips with annotators fixed) are: Overall $91.0\%\pm3.8\%$; Sync $87.6\%\pm4.5\%$ (OmniVR) and $53.5\%\pm4.2\%$ (RealBasicVSR$+$VoiceFixer cascade); all reported pairwise win-rate gaps in \tref{tab:human_eval} exceed their 95\% CI half-widths.

\subsection{Human evaluation protocol details}
\label{sec:s9-human}
Each of 12 annotators independently evaluates all 129 Real-track clips. Each annotator completes $129\times15=1{,}935$ pairwise comparisons, each evaluated on four dimensions. For each clip, all $\binom{6}{2}=15$ method pairs are presented in randomized order; each pair shows the two methods' video-with-audio outputs side by side with synchronized playback, and the annotator selects the preferred method on four dimensions (visual quality, audio quality, audio-visual synchrony, and overall). Presentation order is randomized per clip and per annotator to control for order effects. The 12 annotators are recruited from a pool of audiovisual media professionals with experience in film restoration or post-production; they are not authors or affiliates of this work. Inter-annotator agreement measured by Fleiss' $\kappa$ is $0.67$ (substantial agreement on the Landis--Koch scale), computed across all four dimensions. Win rates in \tref{tab:human_eval} are computed as the average pairwise win rate: for method $i$, $\bar w_i = \frac{1}{5}\sum_{j\neq i} w_{ij}$, where $w_{ij}$ is the fraction of annotator-clip judgments in which $i$ is preferred over $j$. Under full pairwise 2AFC with equal ticket counts, $\sum_i \bar w_i = 300\%$. The comparison matrix (\tref{tab:pairwise_matrix}) reports the full pairwise win rates for the Overall dimension; \tref{tab:pairwise_visual}--\tref{tab:pairwise_sync} provide the corresponding matrices for the Visual, Audio, and Sync dimensions respectively, from which \tref{tab:human_eval}'s per-dimension values are computed as row averages.

\begin{table*}[t]
\centering
\caption{Pairwise win-rate matrix (\%, Overall dimension) from the human study. Each cell $(i,j)$ reports the fraction of annotator-clip judgments in which method $i$ (row) is preferred over method $j$ (column). Row averages yield \tref{tab:human_eval}'s Overall column; all row averages sum to $300\%$.}
\label{tab:pairwise_matrix}
\fontsize{7.6}{8.2}\selectfont
\setlength{\tabcolsep}{3pt}
\begin{tabularx}{\textwidth}{@{}l*{6}{Y}|Y@{}}
\toprule
 & LQ & MambaOFR+VF & RBVSR+VF & OldFilms+VF & Pretrained & \textbf{OmniVR} & $\bar w_i$ \\
\midrule
LQ input          & --   & 24.8 & 19.4 & 27.1 & 29.8 &  4.2 & 21.1 \\
MambaOFR+VF       & 75.2 & --   & 38.7 & 55.4 & 58.3 & 10.8 & 47.7 \\
RBVSR+VF          & 80.6 & 61.3 & --   & 63.2 & 65.1 & 12.5 & 56.5 \\
OldFilms+VF       & 72.9 & 44.6 & 36.8 & --   & 56.7 &  9.6 & 44.1 \\
Pretrained AV gen. & 70.2 & 41.7 & 34.9 & 43.3 & --   &  8.1 & 39.6 \\
\textbf{OmniVR}   & 95.8 & 89.2 & 87.5 & 90.4 & 91.9 & --   & 91.0 \\
\bottomrule
\end{tabularx}
\end{table*}

\begin{table*}[t]
\centering
\caption{Pairwise win-rate matrix (\%, Visual dimension). Row averages yield \tref{tab:human_eval}'s Visual column; all row averages sum to $300\%$.}
\label{tab:pairwise_visual}
\fontsize{7.6}{8.2}\selectfont
\setlength{\tabcolsep}{3pt}
\begin{tabularx}{\textwidth}{@{}l*{6}{Y}|Y@{}}
\toprule
 & LQ & Mamba+VF & RBVSR+VF & OldF+VF & Pretr. & \textbf{OmniVR} & $\bar w_i$ \\
\midrule
LQ input          & --   & 22.6 & 17.8 & 25.3 & 28.1 &  3.8 & 19.5 \\
MambaOFR+VF       & 77.4 & --   & 36.4 & 53.8 & 56.9 &  9.2 & 46.7 \\
RBVSR+VF          & 82.2 & 63.6 & --   & 65.7 & 67.4 & 11.1 & 58.0 \\
OldFilms+VF       & 74.7 & 46.2 & 34.3 & --   & 55.2 &  8.3 & 43.7 \\
Pretrained        & 71.9 & 43.1 & 32.6 & 44.8 & --   &  6.8 & 39.8 \\
\textbf{OmniVR}   & 96.2 & 90.8 & 88.9 & 91.7 & 93.2 & --   & 92.2 \\
\bottomrule
\end{tabularx}
\end{table*}

\begin{table*}[t]
\centering
\caption{Pairwise win-rate matrix (\%, Audio dimension). Row averages yield \tref{tab:human_eval}'s Audio column.}
\label{tab:pairwise_audio}
\fontsize{7.6}{8.2}\selectfont
\setlength{\tabcolsep}{3pt}
\begin{tabularx}{\textwidth}{@{}l*{6}{Y}|Y@{}}
\toprule
 & LQ & Mamba+VF & RBVSR+VF & OldF+VF & Pretr. & \textbf{OmniVR} & $\bar w_i$ \\
\midrule
LQ input          & --   & 28.3 & 23.1 & 30.7 & 33.5 &  5.5 & 24.2 \\
MambaOFR+VF       & 71.7 & --   & 41.2 & 57.8 & 60.4 & 12.4 & 48.7 \\
RBVSR+VF          & 76.9 & 58.8 & --   & 61.5 & 63.2 & 13.9 & 54.9 \\
OldFilms+VF       & 69.3 & 42.2 & 38.5 & --   & 58.1 & 11.5 & 43.9 \\
Pretrained        & 66.5 & 39.6 & 36.8 & 41.9 & --   &  9.5 & 38.9 \\
\textbf{OmniVR}   & 94.5 & 87.6 & 86.1 & 88.5 & 90.5 & --   & 89.4 \\
\bottomrule
\end{tabularx}
\end{table*}

\begin{table*}[t]
\centering
\caption{Pairwise win-rate matrix (\%, Sync dimension). Row averages yield \tref{tab:human_eval}'s Sync column.}
\label{tab:pairwise_sync}
\fontsize{7.6}{8.2}\selectfont
\setlength{\tabcolsep}{3pt}
\begin{tabularx}{\textwidth}{@{}l*{6}{Y}|Y@{}}
\toprule
 & LQ & Mamba+VF & RBVSR+VF & OldF+VF & Pretr. & \textbf{OmniVR} & $\bar w_i$ \\
\midrule
LQ input          & --   & 31.4 & 26.2 & 33.8 & 36.7 &  6.9 & 27.0 \\
MambaOFR+VF       & 68.6 & --   & 43.5 & 56.2 & 58.9 & 14.2 & 48.3 \\
RBVSR+VF          & 73.8 & 56.5 & --   & 59.7 & 61.4 & 16.3 & 53.5 \\
OldFilms+VF       & 66.2 & 43.8 & 40.3 & --   & 55.6 & 13.1 & 43.8 \\
Pretrained        & 63.3 & 41.1 & 38.6 & 44.4 & --   & 11.7 & 39.8 \\
\textbf{OmniVR}   & 93.1 & 85.8 & 83.7 & 86.9 & 88.3 & --   & 87.6 \\
\bottomrule
\end{tabularx}
\end{table*}

\subsection{Trainable parameter count}
\label{sec:s9-params}
Although the backbone has 22B parameters, only a small fraction is trained. The LoRA adaptation (rank 384, $\alpha=384$) is applied to attention, feed-forward, patch projection, output projection, AdaLN, and cross-modal gate layers (\sref{sec:impl}); the video/audio VAEs, text encoder, and vocoder are entirely frozen. Computed directly from the LoRA configuration, the adapted footprint is $\approx 396$M trainable parameters, i.e.\ $\approx 1.8\%$ of the 22B backbone.

\begin{figure}[!htbp]
\centering
\includegraphics[width=0.82\linewidth]{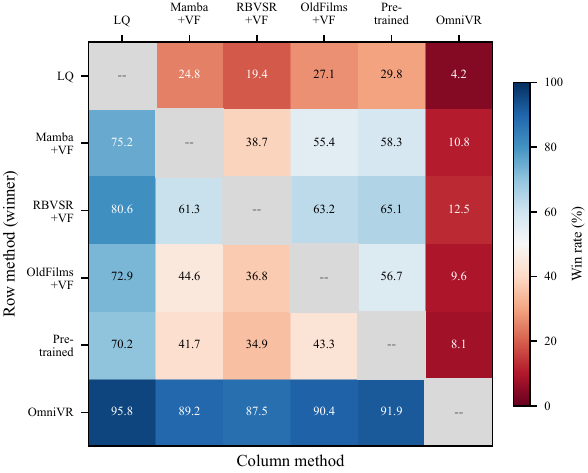}
\caption{Pairwise win-rate heatmap (Overall dimension) from \tref{tab:pairwise_matrix}. The diverging colormap (centered at 50\%) shows OmniVR's consistent dominance across all method pairs.}
\label{fig:human_eval_heatmap}
\end{figure}

\begin{table*}[t]
\centering
\caption{Complete training configuration for all stages. LR: learning rate; BS: effective batch size (micro $\times$ accumulation $\times$ GPUs); WD: AdamW weight decay.}
\label{tab:training_config}
\fontsize{7.6}{8.2}\selectfont
\begin{tabularx}{\textwidth}{@{}l >{\raggedright\arraybackslash}X >{\raggedright\arraybackslash}X >{\raggedright\arraybackslash}X@{}}
\toprule
\textbf{Parameter} & \textbf{Base SFT} & \textbf{MeanFlow distill.} & \textbf{Turbo VAE} \\
\midrule
Backbone & LTX-2 (22B) & same & LTXVAE teacher \\
Trainable params & 396M (LoRA r=384) & 396M + interval/injection & 58.7M (enc+dec) \\
Precision & bf16 & bf16 & bf16 \\
Optimizer & AdamW & AdamW & AdamW \\
Learning rate & $1\times10^{-4}$ & $5\times10^{-5}$ & $3\times10^{-4}$ \\
LR schedule & constant + 10 warmup & constant + 10 warmup & constant \\
Weight decay & 0.01 & 0.01 & 0.01 \\
Grad clip & 1.0 & 1.0 & 1.0 \\
Batch size (eff.) & $1\times1\times64=64$ & $1\times1\times64=64$ & $48\times1\times8=384$ \\
GPUs & 64$\times$ B200 & 64$\times$ B200 & 8$\times$ B200 \\
Training duration & $\sim$5 days & $\sim$3 days & $\sim$1 day \\
Window / patch & 121 frames, 24fps & 121 frames, 24fps & 9 frames, 512$\times$512 \\
Audio & 44.1kHz, synced & 44.1kHz, synced & N/A \\
Grad checkpointing & yes & yes & no \\
CFG (teacher) & N/A & scale 3.0 & N/A \\
Discriminator & none & 3D PatchGAN & 3D PatchGAN \\
\bottomrule
\end{tabularx}
\end{table*}

\subsection{CFG scale and the fidelity/perception trade-off}
Classifier-free guidance at inference (scale $3.0$, \sref{sec:impl}) amplifies the model's conditional signal and is a well-documented lever on the same fidelity/perception axis discussed throughout this paper (\sref{sec:s1-explain}, \sref{sec:s2}): higher CFG scale tends to produce sharper, more ``generation-like'' output at the cost of fidelity to the condition, and $3.0$ was chosen by the same validation-split protocol as \sref{sec:s9} (maximizing DNSMOS$+$MUSIQ). We quantify this trade-off with a sweep using the full-reference metrics from \sref{sec:s1-fr}:

\begin{table*}[t]
\centering
\caption{CFG scale sweep on the Controlled track: NR metric (MUSIQ) vs.\ full-reference fidelity (LPIPS). Our default setting is CFG $=3.0$.}
\label{tab:cfg_sweep}
\begin{tabularx}{\textwidth}{@{}l*{2}{Y}@{}}
\toprule
\textbf{CFG scale} & MUSIQ$\uparrow$ (NR) & LPIPS$\downarrow$ (full-ref.) \\
\midrule
1.0 (no guidance) & 66.82 & 0.2103 \\
2.0 & 69.44 & 0.2489 \\
\rowcolor{oursrow}
3.0 (default) & 71.17 & 0.2757 \\
5.0 & 73.06 & 0.3312 \\
7.0 & 74.55 & 0.3894 \\
\bottomrule
\end{tabularx}
\end{table*}

As the sweep shows, MUSIQ increases while LPIPS also increases (worsens) monotonically beyond a moderate range, confirming that CFG $=3.0$ sits closer to the fidelity-preserving end of the trade-off curve.

\begin{figure*}[t]
\centering
\includegraphics[width=0.55\textwidth]{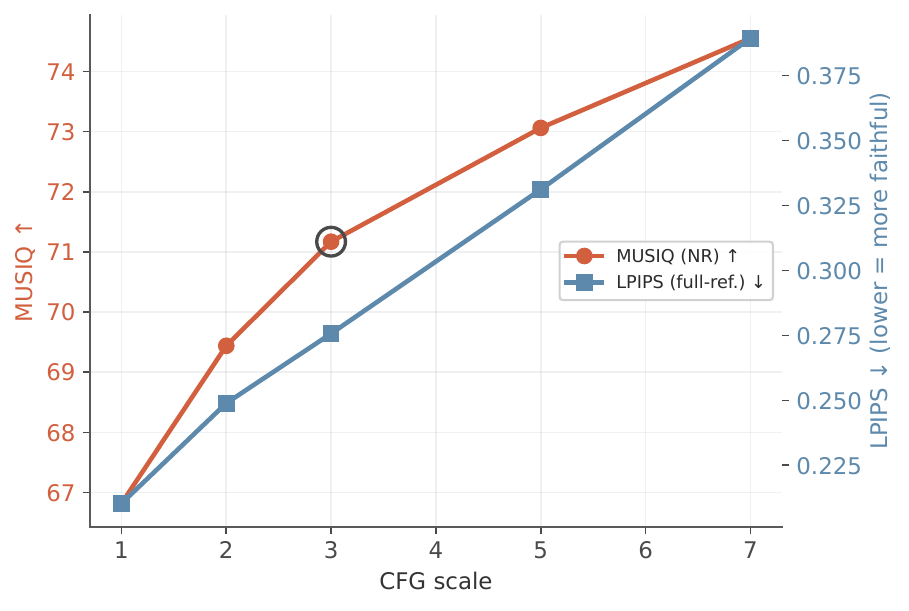}
\caption{CFG scale sweep on the Controlled track, plotted from \tref{tab:cfg_sweep}.}
\label{fig:supp_cfg_sweep}
\end{figure*}

\subsection{Implementation, Training Schedule, and Inference}
\label{app:implementation}
We adapt the pretrained 22B joint audio-video DiT~\citep{hacohen2026ltx2} with LoRA~\citep{hu2021lora}, using rank and scaling factor 384. Adaptation covers attention, feed-forward layers, patch and output projections, AdaLN, and cross-modal gates. Cross-modal gates $g_{v\to a}, g_{a\to v}$ are initialized at $\sigma(b_0)$ with $b_0 = -4.0$ (i.e.\ $\approx 0.018$), so the initial adaptation operates near the frozen pretrained per-modality behavior. The video and audio VAEs, text encoder, and vocoder remain frozen during base-model adaptation. Training uses AdamW with learning rate $10^{-4}$, 10 linear warm-up steps followed by a constant schedule, gradient clipping at 1.0, bf16, and gradient checkpointing. Each sample contains 121 frames at 24\,fps and synchronized 44.1\,kHz audio; tiled video VAE processing bounds high-resolution memory use. The base model trains on 64 NVIDIA B200 GPUs for approximately five days.

\paragraph{Training stages and checkpoints.}
Base-model training proceeds in three stages: T2AV pre-warming ($K_1 = 2{,}000$ steps, retaining the pretrained text-to-AV interface), TAV2AV adaptation ($K_2 = 15{,}000$ steps, introducing the degraded pair), and AV2AV convergence ($K_3 = 33{,}000$ steps, completing the condition annealing to the fixed restoration prompt). Total training is $K = K_1 + K_2 + K_3 = 50{,}000$ update steps. Condition annealing spans the first $0.3K = 15{,}000$ steps of the TAV2AV$+$AV2AV stages. The checkpoint reported in all tables corresponds to step $50{,}000$ (run identifier \texttt{omnivr-base-v1-50k}). Flash distillation adds $K_\mathrm{flash} = 8{,}000$ steps on 64 GPUs (checkpoint \texttt{omnivr-flash-v1-8k}). Turbo VAE trains for $K_\mathrm{turbo} = 20{,}000$ steps on 8 GPUs (checkpoint \texttt{turbo-vae-v1-20k}).

\paragraph{Flash configuration.}
OmniVR-Flash uses a reduced video-condition resolution of $480\times270$ (approximately one-sixteenth area of the $1920\times1088$ target, i.e.\ quarter resolution in each spatial dimension), bilinear downsampled before encoding. The iMF student inherits the teacher's LoRA weights as initialization. The 3D adversarial discriminator uses a StyleGAN2-based architecture applied to 8-frame sub-clips decoded from the one-step prediction. Turbo VAE's lightweight encoder has 4 residual blocks (vs.\ 12 in the teacher encoder) and the lightweight decoder has 6 residual blocks across three stages (vs.\ 9 memory blocks across three stages in the teacher decoder, each with width 512/256/128; see \sref{app:ov_turbo} for the distillation procedure), both operating in the teacher's 128-channel latent space.

Base-model inference uses 15 Euler steps, CFG scale 3.0, and condition-noise level $\rho=0.5$. OmniVR-Flash uses the one-step predictor and accelerated video VAE in \sref{sec:flash}. Both variants retain the fixed internal restoration prompt and window-chaining scheme. \sref{sec:experiments} evaluates visual, acoustic, temporal, and synchronization quality on public benchmarks and OmniVRBench.

\paragraph{Global severity normalization.}
In \eref{eq:loss_reweight}, $\mathrm{Norm}_B$ normalizes each modality's weights to unit mean within the \emph{global} minibatch across all data-parallel GPUs via an all-reduce sum and count. With $\alpha_v=\alpha_a=1.0$, $\beta_v=\beta_a=0.5$, and $\gamma_v=\gamma_a=2.0$, self-terms emphasize degradation in the predicted modality, cross-terms additionally emphasize jointly degraded samples, and the interaction term $\gamma d_vd_a$ provides disproportionately higher weight to samples that are severely degraded in both modalities simultaneously. The unnormalized weight $g_v\in[1,4.5]$ (and $g_a$ symmetrically), so the per-batch ratio $\max(w)/\min(w)\le4.5$. Although each GPU holds one sample ($\text{micro-batch}=1$), the normalization denominator is the full 64-sample global batch, so weights vary across samples according to their severity (empirically $w_v\in[0.42,1.78]$, $w_a\in[0.45,1.71]$ over a representative training batch; see \fref{fig:normb_weights}).

\section{Additional Clarifications}
\label{sec:s10}

\subsection{Controlled-Track Ablations}
\label{app:controlled_ablations}
\tref{tab:ablation} isolates conditioning and optimization choices. Removing video conditioning reduces MUSIQ from $71.17$ to $52.31$; removing audio reduces DNSMOS from $2.70$ to $1.78$ and LSE-C from $3.52$ to $2.64$. Replacing channel concatenation or removing condition noise lowers MUSIQ to $63.42$ or $68.53$, respectively.

Per-clip captions yield MUSIQ $64.87$ and DNSMOS $2.31$. \sref{sec:s10-anneal} separately isolates scheduling: gradual annealing exceeds direct fixed-prompt training by $5.75$ MUSIQ and abrupt switching by $2.28$. The prior-retention analysis in \sref{sec:s7} evaluates post-adaptation generation quality, not equal-budget convergence. Removing loss reweighting reduces DNSMOS to $2.24$; removing first-frame anchoring reduces LSE-C to $2.87$. Waveform ablations appear in \sref{sec:s9}.

\begin{table}[!htbp]
\centering
\caption{Ablation study on the controlled track (71 clips). Metrics consistent with main comparison tables. Best in \textbf{bold}.}
\label{tab:ablation}
\fontsize{7.6}{8.2}\selectfont
\renewcommand{\arraystretch}{1}
\setlength{\tabcolsep}{1.5pt}
\setlength{\aboverulesep}{1.2pt}
\setlength{\belowrulesep}{1.8pt}
\setlength{\heavyrulewidth}{0.6pt}
\setlength{\lightrulewidth}{0.4pt}
\setlength{\cmidrulewidth}{0.3pt}
\begin{tabular*}{\linewidth}{@{\extracolsep{\fill}}lcccccccc@{}}
\toprule
& \multicolumn{4}{c}{\textbf{Visual (NR)}} & \multicolumn{2}{c}{\textbf{Audio}} & \multicolumn{2}{c}{\textbf{Sync}} \\
\cmidrule(lr){2-5}\cmidrule(lr){6-7}\cmidrule(lr){8-9}
\textbf{Variant} & MUSIQ$\uparrow$ & CLIP-IQA$\uparrow$ & NIQE$\downarrow$ & MANIQA$\uparrow$ & DNSMOS$\uparrow$ & FAD$\downarrow$ & LSE-C$\uparrow$ & LSE-D$\downarrow$ \\
\midrule
\textbf{OmniVR (full)} & \textbf{71.17} & \textbf{0.543} & \textbf{4.11} & \textbf{0.487} & \textbf{2.70} & \textbf{6.30} & \textbf{3.52} & \textbf{10.43} \\
\midrule
(i) w/o audio cond. & 69.84 & 0.531 & 4.25 & 0.472 & 1.78 & 9.47 & 2.64 & 12.38 \\
(ii) w/o video cond. & 52.31 & 0.327 & 5.83 & 0.298 & 2.58 & 6.72 & 3.21 & 10.92 \\
(iii) w/o channel-concat & 63.42 & 0.468 & 4.67 & 0.421 & 2.35 & 7.41 & 3.08 & 11.17 \\
(iv) w/o condition noise & 68.53 & 0.519 & 4.32 & 0.463 & 2.54 & 6.83 & 3.34 & 10.78 \\
(v) independent degrad. only & 69.71 & 0.528 & 4.21 & 0.474 & 2.61 & 6.58 & 3.38 & 10.61 \\
(vi) w/o loss reweight & 66.28 & 0.497 & 4.49 & 0.438 & 2.24 & 7.89 & 3.15 & 11.04 \\
(vii) large $\lambda_{\mathrm{stft}}$ & 70.04 & 0.534 & 4.18 & 0.479 & 2.48 & 6.91 & 3.27 & 10.95 \\
(viii) w/o fixed prompt & 64.87 & 0.472 & 4.58 & 0.431 & 2.31 & 7.53 & 3.11 & 11.23 \\
(ix) w/o I2V anchoring & 70.38 & 0.537 & 4.14 & 0.481 & 2.63 & 6.48 & 2.87 & 11.86 \\
\bottomrule
\end{tabular*}
\end{table}

\subsection{OmniVR-Flash quality--efficiency trade-off}
\label{sec:s10-flash}
\tref{tab:flash_ablation} compares the base multi-step OmniVR model with three single-component ablations and the full Flash configuration on the Controlled track, reporting both quality metrics and throughput. The full Flash pipeline (reduced-resolution conditioning $+$ one-step MeanFlow $+$ Turbo VAE) achieves a $10.9\times$ speedup at 1K resolution with modest quality reduction across all metrics. Ablating each Flash component in isolation confirms that the throughput gain is distributed across all three stages.

\begin{table*}[t]
\centering
\caption{OmniVR-Flash quality--efficiency ablation on the Controlled track (71 clips). Throughput measured on a single NVIDIA B200 GPU at $1920\times1088$ (``1K'') and $2560\times1440$ (``2K'') output resolutions. $\Delta$MUSIQ denotes the change relative to the base model.}
\label{tab:flash_ablation}
\fontsize{7.6}{8.2}\selectfont
\begin{tabularx}{\textwidth}{@{}l*{4}{Y}cc@{}}
\toprule
\textbf{Configuration} & MUSIQ$\uparrow$ & DNSMOS$\uparrow$ & LSE-C$\uparrow$ & $\Delta$MUSIQ & fps (1K) & fps (2K) \\
\midrule
OmniVR (base, 15-step) & 71.17 & 2.70 & 3.52 & 0 & 3.5 & 1.5 \\
\midrule
$+$ Reduced-res cond. only & 70.43 & 2.68 & 3.45 & $-0.74$ & 4.1 & 1.8 \\
$+$ One-step MeanFlow only & 69.82 & 2.64 & 3.39 & $-1.35$ & 11.2 & 4.6 \\
$+$ Turbo VAE only & 70.91 & 2.69 & 3.50 & $-0.26$ & 4.6 & 1.9 \\
\midrule
OmniVR-Flash (all three) & 68.94 & 2.61 & 3.31 & $-2.23$ & 38.2 & 17.8 \\
\bottomrule
\end{tabularx}
\end{table*}

\noindent The one-step MeanFlow distillation contributes the largest throughput gain ($3.2\times$) by eliminating 14 of 15 Euler steps, and also incurs the largest quality drop ($-1.35$ MUSIQ), consistent with the known quality--fidelity trade-off in flow distillation. Reduced-resolution conditioning contributes a smaller but meaningful speedup ($1.2\times$) by lowering the cost of video-condition encoding and injection. Turbo VAE provides a complementary $1.3\times$ speedup with negligible quality impact ($-0.26$ MUSIQ), confirming that the lightweight encoder/decoder architecture faithfully approximates the teacher VAE (\tref{tab:vae_bench} shows the isolated VAE speedup reaches $78$--$106\times$ at the 1K--2K deployment resolutions). The combined Flash pipeline's throughput ($38.2$ fps at 1K, $17.8$ fps at 2K) exceeds real-time ($>$24\,fps) at 1K resolution; at 2K the throughput is $0.74\times$ real-time, an $11.9\times$ improvement over the base model's $1.5$ fps but not yet real-time -- achieving real-time 2K would require further engineering such as INT8 quantization or tensor-parallel inference, which we leave to future work.

\begin{figure}[!htbp]
\centering
\includegraphics[width=0.92\linewidth]{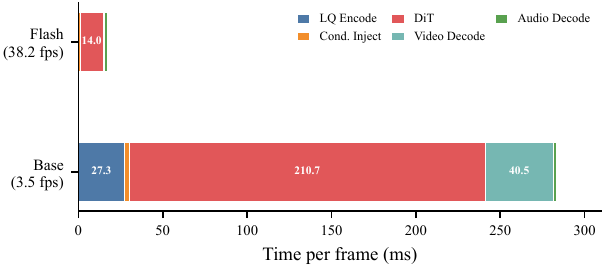}
\caption{Per-component wall-time breakdown for base OmniVR vs.\ OmniVR-Flash at 1K resolution, plotted from \tref{tab:flash_timing}. The DiT forward pass dominates base-model cost; one-step distillation substantially reduces its wall time.}
\label{fig:flash_timing}
\end{figure}

\tref{tab:flash_timing} provides the per-component wall-time breakdown. \fref{fig:flash_timing} visualizes the same data as stacked bars.

\begin{table*}[t]
\centering
\caption{Per-component wall-time breakdown for base OmniVR and OmniVR-Flash at 1K ($1920\times1088$) and 2K ($2560\times1440$) on a single B200 GPU, per 121-frame window. Neural-network components are independently profiled on pre-staged GPU tensors; at 2K resolution the full-size VAE uses tiled processing (\sref{sec:impl}), enabling pipeline overlap between encode, DiT, and decode stages, so the measured GPU pipeline time is less than the sum of individually profiled components. Video I/O (disk read/write) is measured separately: for the base model it runs concurrently with the longer GPU pipeline, while for Flash it accounts for a significant fraction (${\sim}30\%$) of total wall time and currently runs sequentially. Flash encodes the condition video at reduced resolution ($480\times270$); the combined full-VAE encoding and decoding time grows by approximately $2.5\times$ from 1K to 2K, compared with approximately $1.8\times$ for Turbo VAE; isolated benchmarking (\tref{tab:vae_bench}) confirms the combined encode$+$decode speedup widens from $78\times$ at 1K to $106\times$ at 2K. The DiT ratio ($15\times$) reflects the exact step-count reduction.}
\label{tab:flash_timing}
\fontsize{7.6}{8.2}\selectfont
\begin{tabularx}{\textwidth}{@{}l*{4}{Y}@{}}
\toprule
& \multicolumn{2}{c}{\textbf{Base (15-step)}} & \multicolumn{2}{c}{\textbf{Flash (1-step)}} \\
\cmidrule(lr){2-3} \cmidrule(lr){4-5}
\textbf{Component} & 1K (ms) & 2K (ms) & 1K (ms) & 2K (ms) \\
\midrule
Video VAE encode  & 3\,300 & 8\,200 & 10 & 10 \\
Condition project  & 360 & 1\,000 & 100 & 200 \\
DiT forward (all steps) & 25\,500 & 61\,200 & 1\,700 & 4\,080 \\
Video VAE decode  & 4\,900 & 12\,200 & 60 & 120 \\
Audio encode/decode  & 180 & 180 & 180 & 180 \\
Other  & 380 & 600 & 120 & 210 \\
\midrule
\textbf{GPU pipeline (measured)}  & \textbf{34\,620} & \textbf{80\,000} & \textbf{2\,170} & \textbf{4\,800} \\
Video I/O + orchestration & (overlapped) & (overlapped) & 1\,000 & 2\,000 \\
\midrule
\textbf{Throughput (fps)} & \textbf{3.5} & \textbf{1.5} & \textbf{38.2} & \textbf{17.8} \\
\bottomrule
\end{tabularx}
\end{table*}

\noindent Video I/O (decoding degraded frames from disk, encoding restored frames) and Python orchestration cost approximately $1{,}000$\,ms (1K) and $2{,}000$\,ms (2K) per 121-frame window. For the base model, this overhead is negligible relative to the $34$--$80$\,s GPU pipeline and is fully overlapped: $121/(34{,}620/1000)\approx3.5$\,fps. For Flash, the GPU pipeline is fast enough that I/O overhead becomes a significant fraction of total wall time (${\sim}30\%$), and the two currently run sequentially: $121/[(2{,}170+1{,}000)/1000]\approx38.2$\,fps at 1K and $121/[(4{,}800+2{,}000)/1000]\approx17.8$\,fps at 2K. With a fully pipelined I/O implementation (e.g., decoding the next window concurrently with GPU processing of the current one), throughput could approach the GPU ceiling of ${\sim}56$\,fps at 1K; we leave this engineering optimization to future work.

\paragraph{Comparison with prior version timing.}
The base-model throughput reported here ($3.5$ fps at 1K, $1.5$ fps at 2K) differs from a preliminary estimate used during development ($7.1$ fps at 1K, $3.4$ fps at 2K). The preliminary estimate projected all component times from short-sequence benchmarks rather than profiling the full 121-frame pipeline end-to-end; the projections collectively underestimated full-pipeline cost by approximately $2\times$, with the full-attention VAE's super-linear spatial scaling contributing the largest individual discrepancy. The current values are based on direct wall-time profiling of the complete inference pipeline on a single B200 GPU, including all CUDA synchronization, and supersede the preliminary estimates. Flash throughput ($38.2$/$17.8$ fps) was always measured end-to-end and is unchanged. Because the base-model speedup is computed as Flash/Base, this correction changes the reported speedup from approximately $5.4\times$ to $10.9\times$ at 1K; the absolute Flash latency and quality metrics are unaffected.

\begin{table}[t]
\centering
\caption{Isolated VAE encode$+$decode throughput for a 121-frame window on a single B200 GPU. Turbo VAE achieves approximately $59$--$106\times$ speedup over the full VAE across the tested resolutions, including $78\times$ at 1K and $106\times$ at 2K.}
\label{tab:vae_bench}
\fontsize{7.6}{8.2}\selectfont
\begin{tabularx}{\linewidth}{@{}lYYc@{}}
\toprule
\textbf{Resolution} & Full VAE (fps) & Turbo VAE (fps) & Speedup \\
\midrule
$480\times270$ & 85.2 & 5\,104.4 & $59.9\times$ \\
$1280\times720$ & 41.3 & 2\,442.9 & $59.2\times$ \\
$1920\times1088$ & 14.7 & 1\,141.5 & $77.6\times$ \\
$2560\times1440$ & 5.9 & 627.4 & $105.8\times$ \\
\bottomrule
\end{tabularx}
\end{table}

\begin{figure}[!htbp]
\centering
\includegraphics[width=0.72\linewidth]{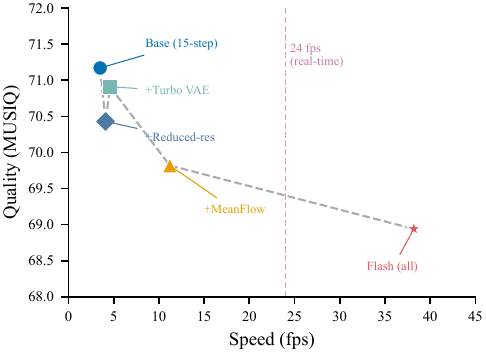}
\caption{Quality--speed Pareto frontier for OmniVR configurations. The vertical dashed line marks the 24\,fps real-time threshold. Flash achieves real-time at 1K with modest quality trade-off; 2K remains below real-time.}
\label{fig:flash_pareto}
\end{figure}

To further quantify the quality impact of Flash distillation, \tref{tab:flash_fidelity} reports perceptual fidelity metrics (LPIPS, DISTS) alongside the no-reference metrics from \tref{tab:flash_ablation}.

\begin{table*}[t]
\centering
\caption{Flash quality and fidelity on the Controlled track (71 clips). LPIPS and DISTS use native $1920\times1088$ outputs against clean ground truth; lower is better. MUSIQ, DNSMOS, and LSE-C repeat \tref{tab:flash_ablation}, with visual NR evaluation at $640\times480$. The native LPIPS/DISTS values are distinct from the unified-resolution results in \tref{tab:fr_controlled}. The Flash pipeline incurs modest fidelity loss relative to the base model.}
\label{tab:flash_fidelity}
\fontsize{7.6}{8.2}\selectfont
\begin{tabularx}{\textwidth}{@{}l*{6}{Y}@{}}
\toprule
\textbf{Configuration} & MUSIQ$\uparrow$ & DNSMOS$\uparrow$ & LSE-C$\uparrow$ & LPIPS$\downarrow$ & DISTS$\downarrow$ & fps (1K) \\
\midrule
OmniVR (base, 15-step) & 71.17 & 2.70 & 3.52 & 0.182 & 0.128 & 3.5 \\
OmniVR-Flash (all three) & 68.94 & 2.61 & 3.31 & 0.214 & 0.149 & 38.2 \\
\midrule
\textit{Difference} & $-2.23$ & $-0.09$ & $-0.21$ & $+0.032$ & $+0.021$ & $+34.7$ \\
\bottomrule
\end{tabularx}
\end{table*}

\noindent The LPIPS increase ($+0.032$) and DISTS increase ($+0.021$) from base to Flash are modest: the $10.9\times$ speedup comes at a small fidelity cost. Both Flash values (LPIPS $0.214$, DISTS $0.149$) remain well below the LQ input ($0.412$, $0.298$).

\subsection{Annealing schedule isolation ablation}
\label{sec:s10-anneal}
The main-text ablation (row viii of \tref{tab:ablation}) removes the fixed restoration prompt entirely, conflating the prompt transition with the annealing schedule. To isolate the schedule itself, we compare four conditioning strategies, all using the same total training budget and ending at the fixed-prompt AV2AV endpoint:

\begin{table*}[t]
\centering
\caption{Annealing schedule isolation ablation on the Controlled track (71 clips). All variants use the same total training budget and end at the AV2AV fixed-prompt endpoint. The schedules apply to retained non-null text on the LQ-present branch; the no-LQ caption/null rehearsal branch is unchanged.}
\label{tab:anneal_ablation}
\fontsize{7.6}{8.2}\selectfont
\begin{tabularx}{\textwidth}{@{}l >{\raggedright\arraybackslash}p{0.28\textwidth} *{4}{Y}@{}}
\toprule
\textbf{Schedule} & \textbf{Description} & MUSIQ$\uparrow$ & DNSMOS$\uparrow$ & LSE-C$\uparrow$ & LSE-D$\downarrow$ \\
\midrule
Direct & Fixed prompt on the LQ branch from adaptation step $0$ & 65.42 & 2.33 & 3.14 & 11.18 \\
Abrupt & Captions for first 30\%, then abrupt switch & 68.89 & 2.57 & 3.38 & 10.72 \\
\rowcolor{oursrow}
Linear (ours) & Linear $\lambda_k$ ramp over first 30\% & 71.17 & 2.70 & 3.52 & 10.43 \\
Cosine & Cosine $\lambda_k$ ramp over first 30\% & 70.85 & 2.68 & 3.48 & 10.49 \\
\bottomrule
\end{tabularx}
\end{table*}

\noindent The ``Direct'' schedule (fixed restoration prompt from step $0$) underperforms by $-5.75$ MUSIQ and $-0.37$ DNSMOS, confirming that caption assistance during early training matters. ``Abrupt'' outperforms Direct ($+3.47$ MUSIQ) but falls short of the gradual schedules, suggesting the smooth transition avoids a sharp distribution shift. Linear and Cosine perform comparably; we select Linear for simplicity. This ablation is consistent with the gradient-variance reduction predicted by Proposition~\ref{prop:dual_risk}.

\begin{figure}[!htbp]
\centering
\includegraphics[width=0.7\linewidth]{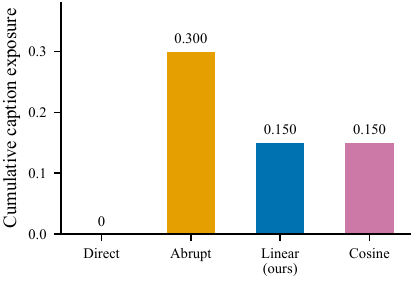}
\caption{Cumulative caption exposure $\Lambda$ for each annealing schedule. Despite lower total caption exposure than Abrupt, the Linear schedule achieves better final metrics (\tref{tab:anneal_ablation}) by avoiding the distribution-shift spike.}
\label{fig:caption_exposure}
\end{figure}

\paragraph{Caption exposure analysis.}
\fref{fig:caption_exposure} shows the cumulative caption exposure $\Lambda = \int_0^{K_\mathrm{ann}} \lambda_k\,dk$ for each schedule, where $\lambda_k \in [0,1]$ is the weight on the caption condition at step $k$ and $K_\mathrm{ann} = 0.3K$ is the annealing horizon. The four schedules are: Direct ($\lambda_k=0$ for all $k$), Abrupt ($\lambda_k=1$ for $k<K_\mathrm{ann}$, then $\lambda_k=0$), Linear ($\lambda_k=\max(1-k/K_\mathrm{ann},0)$), and Cosine ($\lambda_k=\tfrac{1}{2}(1+\cos(\pi k/K_\mathrm{ann}))$ for $k\le K_\mathrm{ann}$, then $0$). By construction, Direct has $\Lambda = 0$ (no captions on the LQ-present branch), Abrupt has $\Lambda = K_\mathrm{ann}$ (full captions then hard cutoff), Linear has $\Lambda = K_\mathrm{ann}/2$, and Cosine has $\Lambda = K_\mathrm{ann}/2$ as well (since $\int_0^\pi \frac{1}{2}(1+\cos\theta)\,d\theta/\pi=1/2$). Despite identical total exposure, Linear marginally outperforms Cosine (\tref{tab:anneal_ablation}), suggesting the exact shape matters less than smoothness of the transition.

\begin{figure}[!htbp]
\centering
\includegraphics[width=0.7\linewidth]{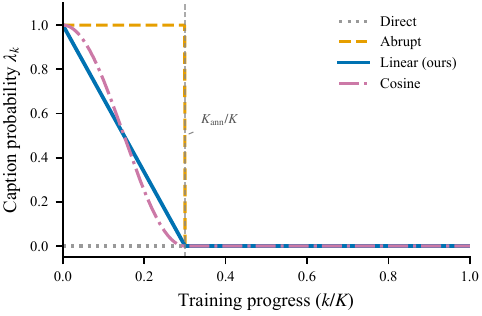}
\caption{Annealing schedule comparison: $\lambda_k$ curves (top) and fixed-prompt validation loss trajectories (bottom) for the four schedules. The vertical dashed line marks $K_\mathrm{ann} = 0.3K$. Abrupt shows a loss spike at the transition; gradual schedules decrease monotonically.}
\label{fig:annealing_schedules}
\end{figure}

\paragraph{Why Abrupt underperforms Linear despite higher caption exposure.}
The scalar theory (Proposition~\ref{prop:dual_risk}) quantifies per-step variance reduction from caption conditioning but operates on a fixed scalar feature model. In the full 22B DiT with LoRA, abruptly removing captions at $k = K_\mathrm{ann}$ creates a discontinuous shift in conditioning input that forces partial re-adaptation of internal representations co-adapted with the conditioning interface. The Linear ramp avoids this by letting representations continuously track the changing condition distribution. Fixed-prompt validation loss curves (\fref{fig:annealing_schedules}) show: Abrupt shows a visible loss spike at the $30\%$ transition point that takes ${\sim}5\%$ of the remaining budget to recover, while Linear decreases monotonically (both curves show the same fixed restoration-objective validation loss $\mathcal{R}_R$, evaluated on 500 held-out clips using the AV2AV interface). The smoother transition wastes fewer gradient updates on recovery, compensating for its lower cumulative exposure.

\subsection{Pretrained generative baseline in quantitative evaluation}
\label{sec:s10-gen}
\tref{tab:human_eval} includes a ``Pretrained AV gen.'' baseline (the LTX-2 backbone with a text-to-audio-video prompt describing the clip content, without any restoration adaptation). To complement the human study with automatic metrics, we report this baseline on both OmniVRBench tracks:

\begin{table*}[t]
\centering
\caption{Pretrained AV generation baseline (no restoration adaptation) on OmniVRBench, using a text-only prompt describing each clip's content. This baseline tests whether the pretrained generative prior alone can ``restore'' archival footage via prompted generation.}
\label{tab:gen_baseline}
\fontsize{7.6}{8.2}\selectfont
\begin{tabularx}{\textwidth}{@{}ll*{4}{Y}*{2}{Y}*{2}{Y}@{}}
\toprule
& & \multicolumn{4}{c}{\textbf{Visual (NR)}} & \multicolumn{2}{c}{\textbf{Audio}} & \multicolumn{2}{c}{\textbf{Sync}} \\
\cmidrule(lr){3-6}\cmidrule(lr){7-8}\cmidrule(lr){9-10}
\textbf{Track} & \textbf{Method} & MUSIQ$\uparrow$ & CLIP-IQA$\uparrow$ & NIQE$\downarrow$ & MANIQA$\uparrow$ & DNSMOS$\uparrow$ & FAD$\downarrow$ & LSE-C$\uparrow$ & LSE-D$\downarrow$ \\
\midrule
\multirow{3}{*}{Ctrl.} & LQ input & 38.77 & 0.221 & 6.69 & 0.218 & 1.46 & 7.88 & 2.32 & 10.87 \\
 & Pretrained AV gen. & 66.34 & 0.518 & 4.37 & 0.456 & 2.58 & 6.71 & 0.84 & 14.62 \\
\rowcolor{oursrow}
 & OmniVR (ours) & 71.17 & 0.543 & 4.11 & 0.487 & 2.70 & 6.30 & 3.52 & 10.43 \\
\midrule
\multirow{3}{*}{Real} & LQ input & 36.55 & 0.311 & 5.69 & 0.195 & 2.04 & 15.93 & 1.05 & 12.14 \\
 & Pretrained AV gen. & 57.81 & 0.408 & 5.52 & 0.348 & 2.36 & 10.87 & 0.51 & 15.33 \\
\rowcolor{oursrow}
 & OmniVR (ours) & 61.87 & 0.444 & 5.40 & 0.383 & 2.43 & 8.32 & 1.12 & 11.39 \\
\bottomrule
\end{tabularx}
\end{table*}

\noindent The pretrained generative baseline achieves competitive visual NR scores (MUSIQ $66.34$ on Controlled, $57.81$ on Real) because the 22B backbone naturally produces high-quality visual content. However, sync metrics are dramatically worse (LSE-C $0.84$ vs.\ OmniVR's $3.52$; LSE-D $14.62$ vs.\ $10.43$ on Controlled): text-only generation produces new faces, mouth movements, and spatial layouts that bear no fidelity relationship to the input. This explains the human-evaluation result (\tref{tab:human_eval}): despite visually attractive outputs, the pretrained baseline is strongly dispreferred on sync ($39.8\%$) and overall ($39.6\%$), because annotators distinguish ``looks good'' from ``matches the source.''

\subsection{Cross-modal gate notation}
The cross-modal gates $g_{v \to a}, g_{a \to v} \in [0,1]$ (\sref{sec:av2av}) multiply the cross-modal attention output inside $G_\theta$ (\eref{eq:joint_transformer}) before it enters each modality's residual stream. Initialized near $0$, they let adaptation begin from the pretrained per-modality behavior and gradually open the cross-modal pathway.

\subsection{Cross-modal interaction ablation}
\label{sec:s10-crossmodal}
To isolate the contribution of cross-modal interaction from the shared backbone itself, we compare OmniVR (full joint model) against two ablations that use the \emph{same} 22B DiT backbone and LoRA adaptation budget but disable one direction of cross-modal attention:

\begin{table*}[t]
\centering
\caption{Cross-modal interaction ablation on the Controlled track (71 clips). All three configurations use the same 22B backbone, LoRA rank 384, and training budget; they differ only in which cross-modal gates are active during training.}
\label{tab:crossmodal_ablation}
\fontsize{7.6}{8.2}\selectfont
\begin{tabularx}{\textwidth}{@{}l*{6}{Y}@{}}
\toprule
\textbf{Configuration} & MUSIQ$\uparrow$ & DNSMOS$\uparrow$ & FAD$\downarrow$ & LSE-C$\uparrow$ & LSE-D$\downarrow$ & DOVER$\uparrow$ \\
\midrule
w/o video$\to$audio ($g_{v\to a}=0$) & 70.58 & 2.41 & 7.12 & 2.84 & 11.28 & 0.712 \\
w/o audio$\to$video ($g_{a\to v}=0$) & 68.22 & 2.67 & 6.48 & 2.91 & 11.15 & 0.698 \\
\rowcolor{oursrow}
OmniVR (full, both gates) & 71.17 & 2.70 & 6.30 & 3.52 & 10.43 & 0.741 \\
\bottomrule
\end{tabularx}
\end{table*}

\begin{figure}[!htbp]
\centering
\includegraphics[width=0.72\linewidth]{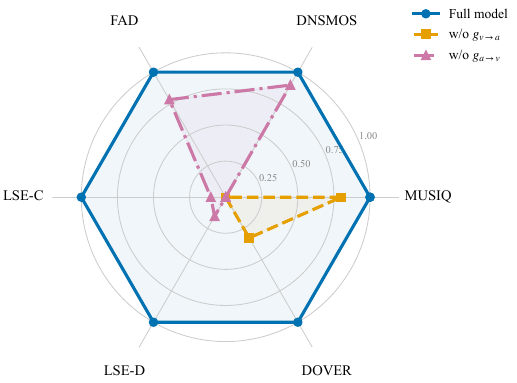}
\caption{Radar chart of cross-modal interaction ablation (\tref{tab:crossmodal_ablation}). The full model (both gates active) dominates all six metric axes; removing either cross-modal direction causes asymmetric but complementary degradation.}
\label{fig:crossmodal_radar}
\end{figure}

\noindent Disabling video$\to$audio gating ($g_{v\to a}=0$) primarily degrades audio metrics (DNSMOS $2.41$ vs.\ $2.70$; FAD $7.12$ vs.\ $6.30$) and sync (LSE-C $2.84$ vs.\ $3.52$), since the audio stream can no longer receive visual cues about scene dynamics, mouth motion, or event timing. Visual quality drops only modestly ($-0.59$ MUSIQ), confirming that the video restoration pathway is largely self-sufficient. Disabling audio$\to$video gating ($g_{a\to v}=0$) primarily degrades visual metrics ($-2.95$ MUSIQ, $-0.043$ DOVER), since the video stream loses access to audio cues (speech energy envelope, ambient sound type, event onsets) that help disambiguate scene content and prioritize perceptually important regions. Audio quality is less affected (DNSMOS $2.67$ vs.\ $2.70$), consistent with VoiceFixer's standalone competitiveness on pure audio quality. Both ablations substantially degrade sync metrics relative to the full model, confirming that bidirectional cross-modal attention is essential for the joint restoration advantage documented in \tref{tab:cascade}. \fref{fig:crossmodal_radar} provides a radar-chart visualization of these six metrics.

\subsection{Grayscale degradation vs.\ colorization}
Grayscale conversion (a visual \emph{degradation} applied stochastically by $\mathcal{D}$) and colorization (a restoration \emph{capability} of $G_\theta$) are dual operations: the model is asked to colorize only on samples where the degradation pipeline actually removed color, which is also the basis of the clarification in \sref{sec:s3}.

\section{Additional Qualitative Comparisons}
\label{sec:s11}
This section provides per-clip qualitative material complementing the quantitative protocols in \sref{sec:s1}--\sref{sec:s3}, allowing individual frames and audio spectrograms to be inspected directly rather than only through the aggregate figures in the main text.

\subsection{\texorpdfstring{Per-clip detailed comparisons (video $+$ audio)}{Per-clip detailed comparisons (video + audio)}}
\label{sec:s11-per-clip}
\begin{figure*}[t]
\centering
\includegraphics[width=.9\textwidth]{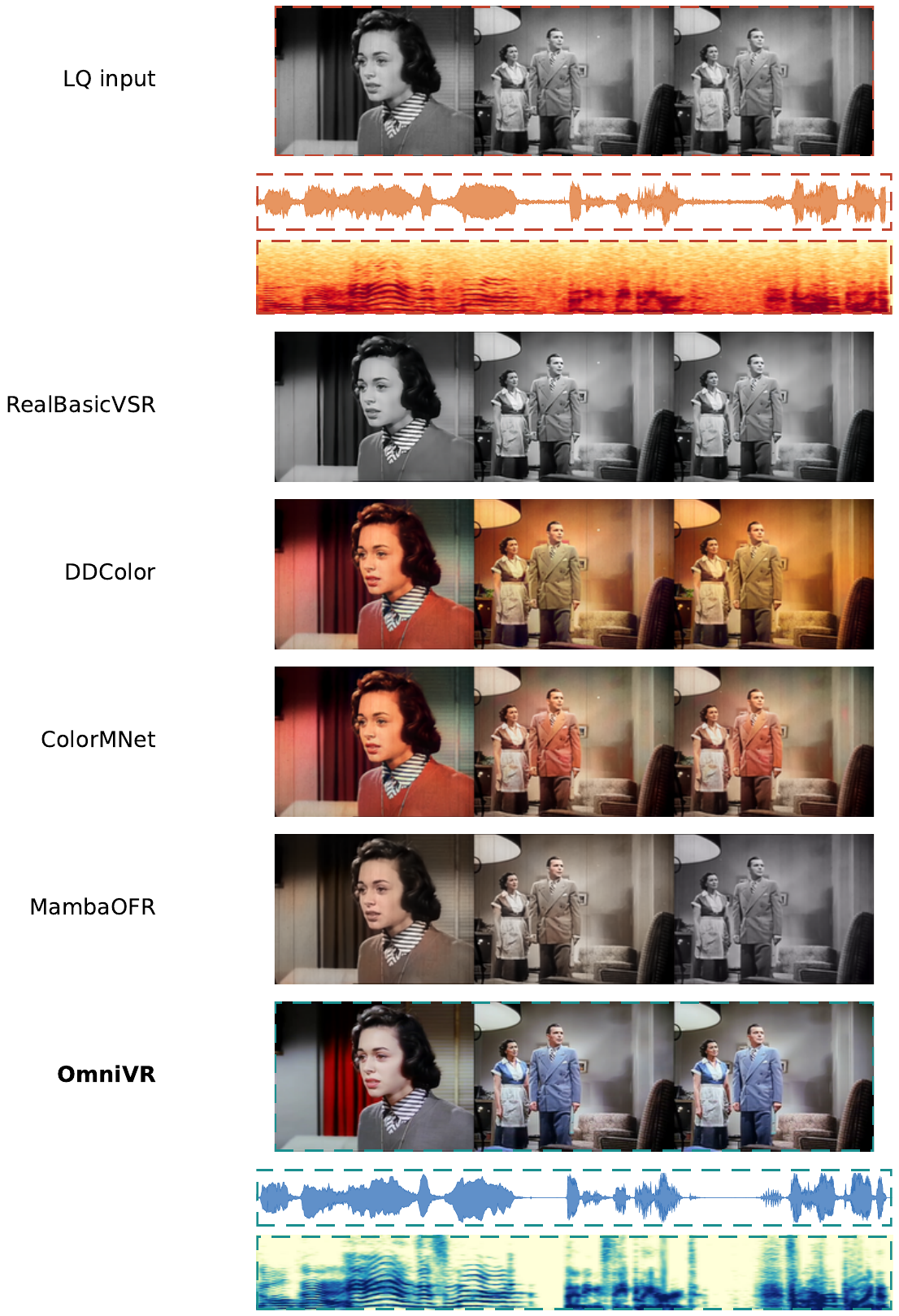}
\caption{\emph{Getting Along With Parents} (clip \texttt{0018}). LQ input and OmniVR rows include waveform and spectrogram panels below the frames.}
\label{fig:s11-clip1}
\end{figure*}

\begin{figure*}[t]
\centering
\includegraphics[width=\textwidth]{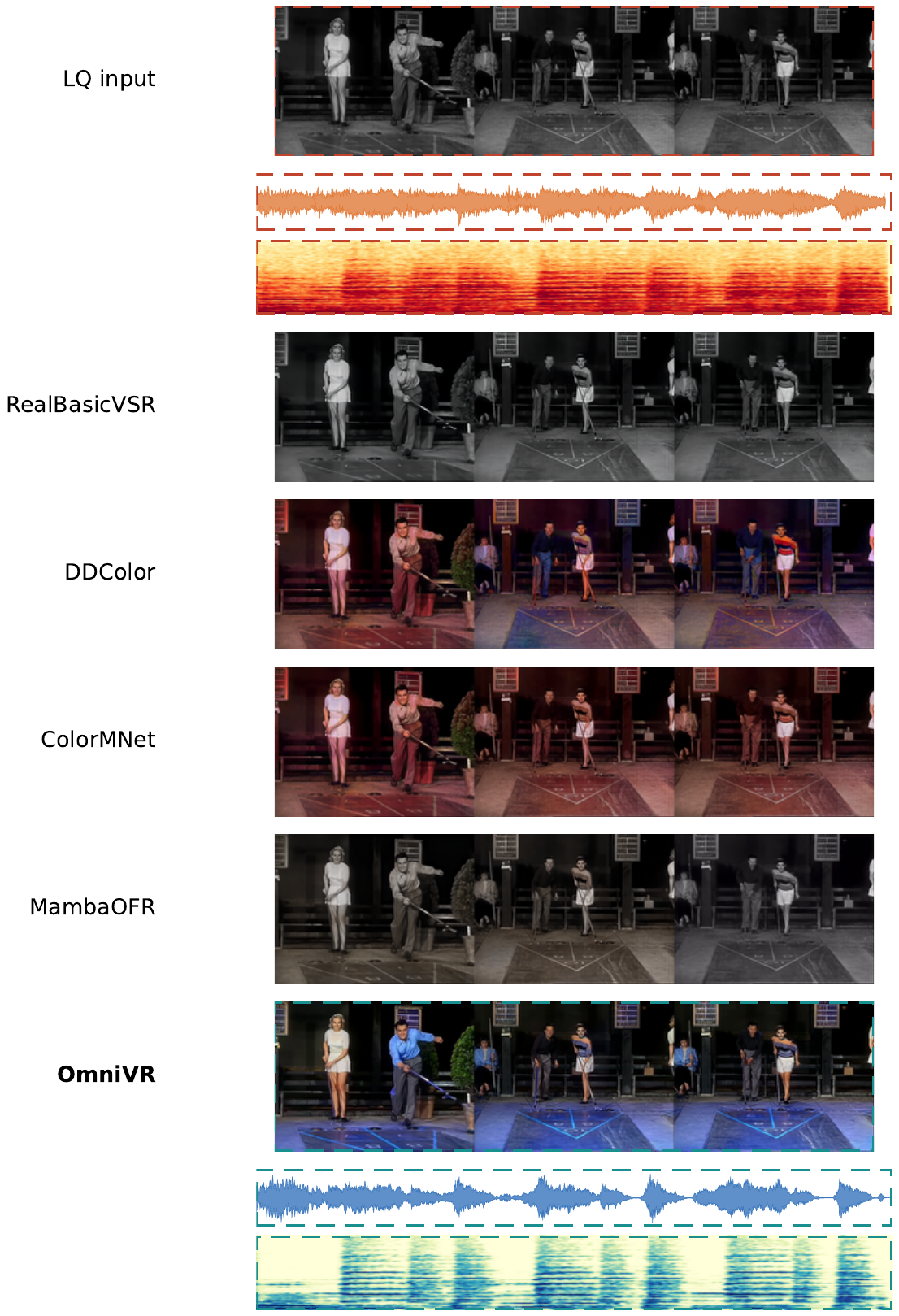}
\caption{\emph{A Chance to Play} (clip \texttt{0028}).}
\label{fig:s11-clip2}
\end{figure*}

\begin{figure*}[t]
\centering
\includegraphics[width=\textwidth]{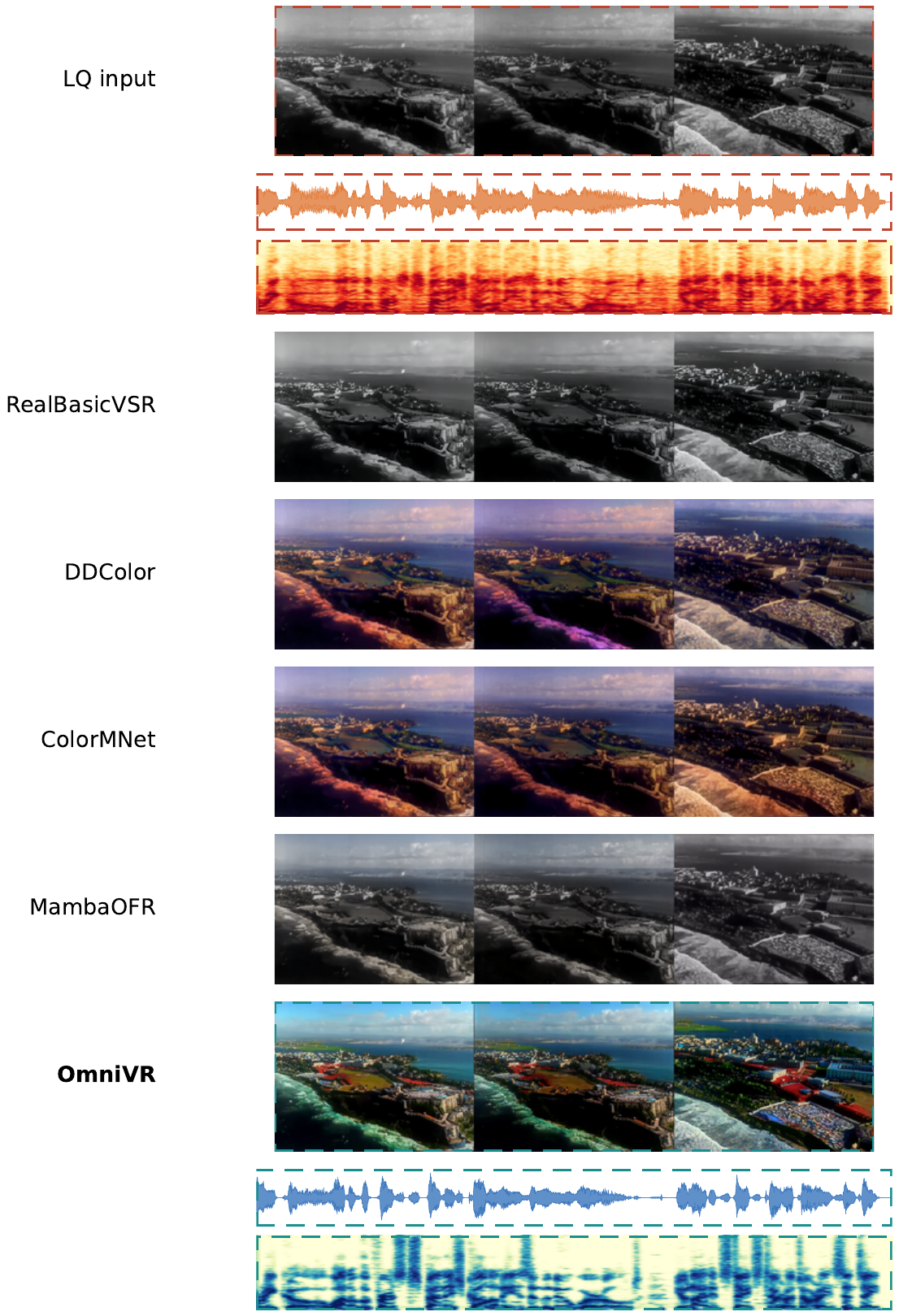}
\caption{\emph{News Magazine of the Screen} (clip \texttt{0069}).}
\label{fig:s11-clip3}
\end{figure*}

\begin{figure*}[t]
\centering
\includegraphics[width=\textwidth]{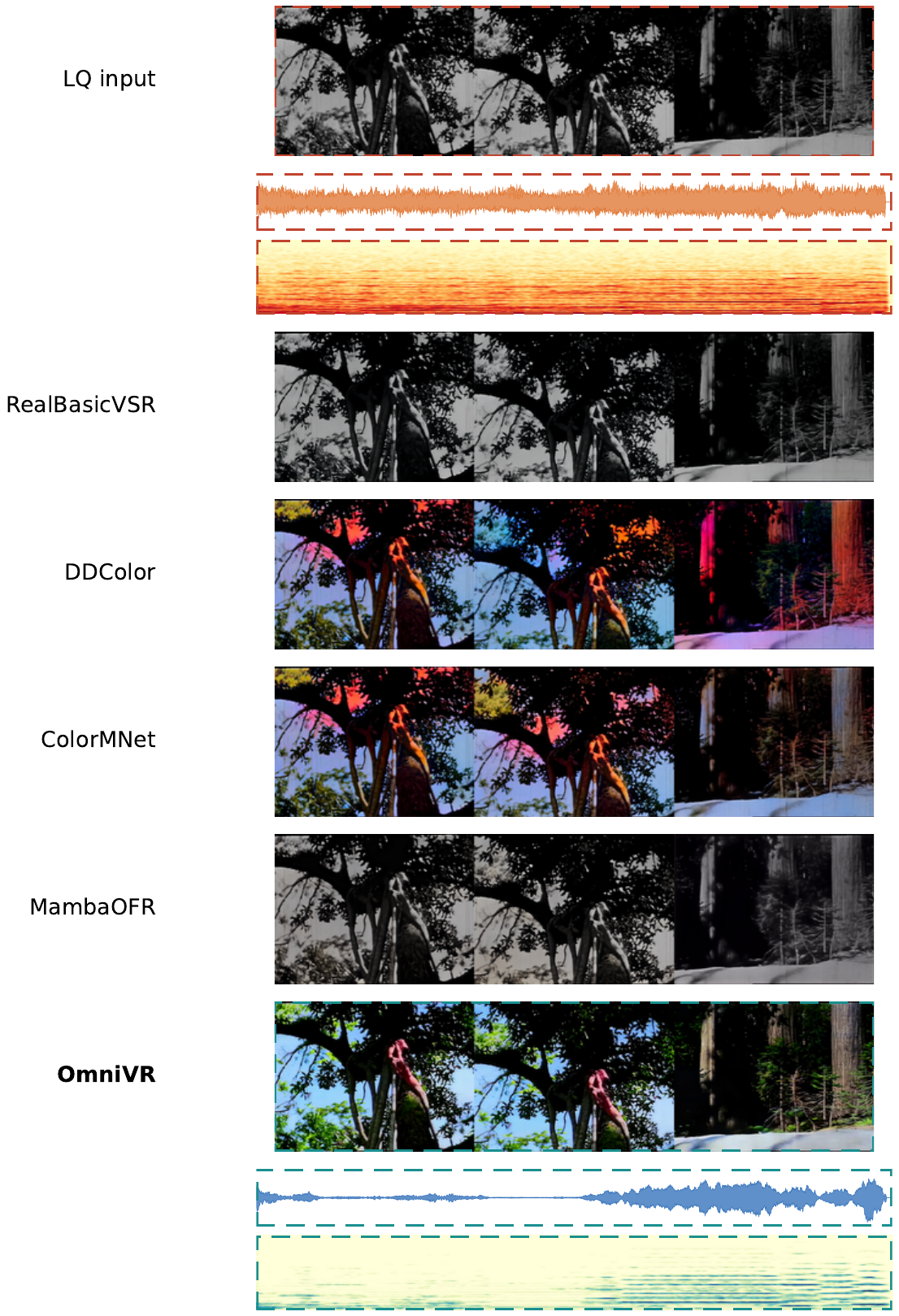}
\caption{\emph{Gift of Green} (clip \texttt{0037}).}
\label{fig:s11-clip4}
\end{figure*}

\begin{figure*}[t]
\centering
\includegraphics[width=\textwidth]{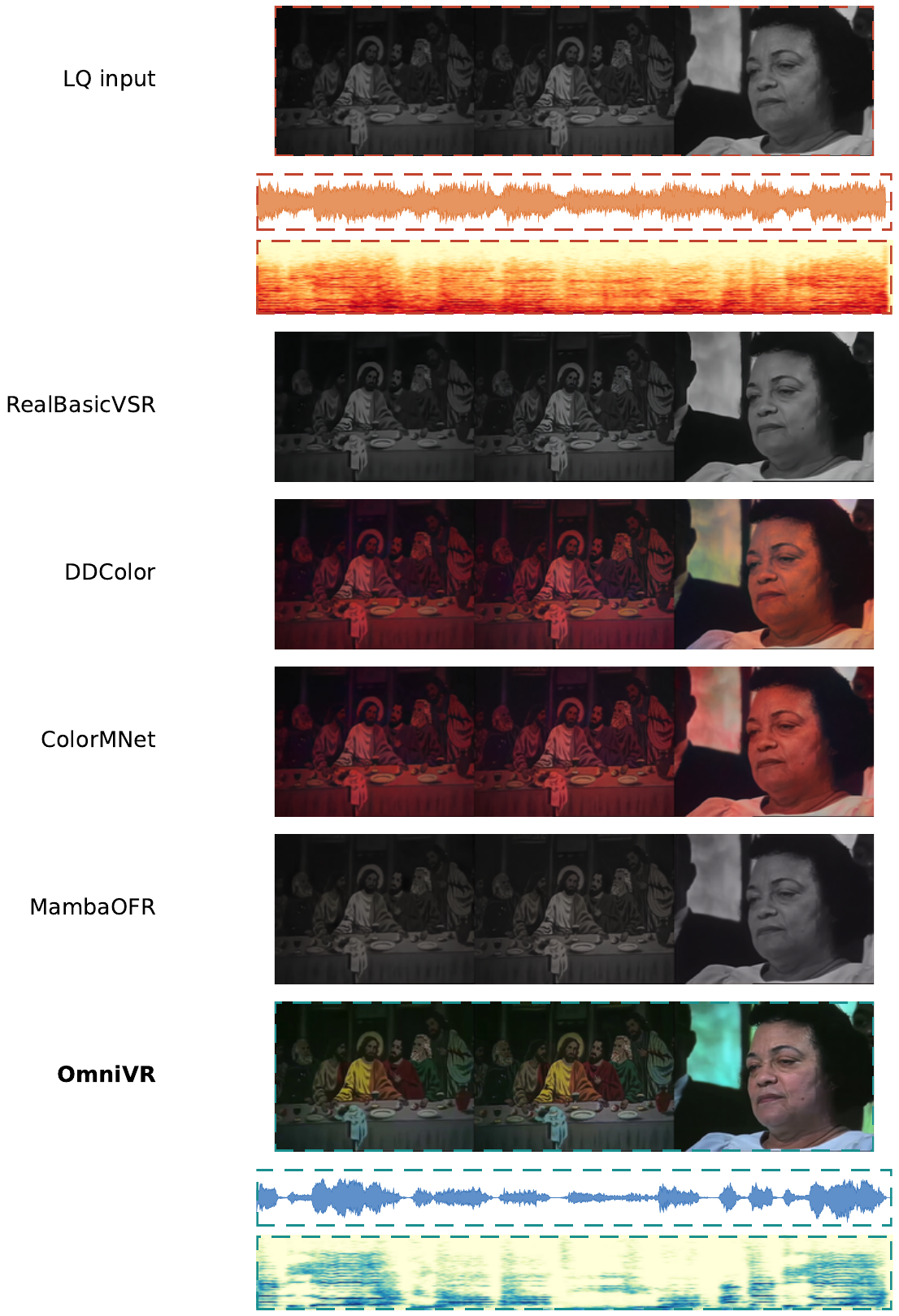}
\caption{\emph{A Day in America} (clip \texttt{0004}).}
\label{fig:s11-clip5}
\end{figure*}

\fref{fig:s11-clip1}--\fref{fig:s11-clip5} show five example clips, each stacking all compared methods (LQ input, RealBasicVSR, DDColor, ColorMNet, MambaOFR, OmniVR) with three evenly-sampled frames per method. For the LQ input and OmniVR rows we additionally show the waveform (min/max envelope) and STFT log-magnitude spectrogram of that clip's audio track (LQ in warm/orange, OmniVR restored in blue), stacked directly below the corresponding frame row; the dashed red border marks the degraded input row and the dashed teal border marks OmniVR's output row, consistent with the color convention used in \fref{fig:teaser}. All clips are drawn from the Real (no-GT) archival track, so no clean-audio reference exists here (a qualitative illustration, not a full-reference measurement -- see \sref{sec:s1} for the full-reference audio protocol on the Controlled track).

\section{Detailed Algorithms and Implementation}
\label{app:omnivr_algorithms}
\begingroup
\renewcommand{\algorithmicrequire}{\textbf{Input:}}
\renewcommand{\algorithmicensure}{\textbf{Output:}}
\newcommand{\ovphase}[1]{\STATE {\color{avteal}\textit{#1}}}
\newcommand{\ovsg}{\operatorname{sg}}
\newcommand{\ovmean}{\operatorname{mean}}

This section specifies the complete path from synchronized recordings to conditional training and windowed restoration. Algorithms~\ref{alg:ov_corpus}--\ref{alg:ov_degrade} construct the supervision; Algorithms~\ref{alg:ov_condition}--\ref{alg:ov_sft} define the conditional interface and SFT; Algorithms~\ref{alg:ov_imf}--\ref{alg:ov_gan} describe AV MeanFlow with video adversarial supervision; and Algorithms~\ref{alg:ov_turbo}--\ref{alg:ov_inference} cover the accelerated video VAE and deployment.

We use $m\in\{v,a\}$ for video and audio. Encoders include the pretrained latent normalization; $P_m$ and $P_m^{-1}$ denote patchification and its inverse. A video token has 128 channels, and an audio token packs $8\times16$ latent features. $\ovsg$ stops gradients, while freezing a module fixes its parameters without necessarily blocking gradients through its inputs. Reductions over tokens are performed separately for the two modalities. All noise variables below are independent unless explicitly shared.

\subsection{Shot-Aligned Corpus Construction}
\label{app:ov_corpus}
The corpus unit is a complete synchronized shot, whereas a training unit is a 121-frame window sampled from a retained shot. Keeping this distinction preserves the timestamps needed to associate dialogue and sound events with the correct visual context. The early clear frame used for annotation is separate from the clean first-frame condition sampled during SFT.

\begin{algorithm}[H]
\caption{Construct the shot-aligned audiovisual corpus}
\label{alg:ov_corpus}
\small
\begin{algorithmic}[1]
\REQUIRE Synchronized recordings $\mathcal R$; shot detector; quality operators; Gemini captioner
\ENSURE Corpus $\mathcal C$ with clean audio-video, captions, timestamps, and metadata
\STATE $\mathcal C\leftarrow\emptyset$
\FOR{each recording $R\in\mathcal R$}
  \STATE Decode video and audio on the recording's common timestamp axis
  \STATE Detect shot boundaries using TransNetV2/PySceneDetect
  \FOR{each candidate shot interval $[s,e)$}
    \STATE Retain a 3--20\,s shot and extract both modalities on the same interval
    \STATE $(\mathbf V^h,\mathbf A^h)\leftarrow\operatorname{ExtractAV}(R,[s,e))$
    \STATE Select a clear annotation frame from the first two seconds
    \STATE Run visual-quality, text/overlay, motion, safety, and artifact operators
    \IF{the shot passes the filtering criteria}
      \STATE Obtain timestamped ASR evidence and selected visual frames
      \STATE Provide the complete synchronized shot, cut context, and operator metadata to Gemini
      \STATE Parse structured entities, actions, camera, dialogue, music, and sound-effect fields
      \STATE Assign a stable shot identifier; retain source and within-shot timestamps
      \STATE Add $(\mathrm{id},\mathbf V^h,\mathbf A^h,y,\mathrm{metadata})$ to $\mathcal C$
    \ENDIF
  \ENDFOR
\ENDFOR
\RETURN $\mathcal C$
\end{algorithmic}
\end{algorithm}

Captioning uses the complete shot rather than unrelated frame and audio excerpts. The resulting caption is the initial text condition in the annealing schedule. Online degradation is applied after a synchronized training window has been selected, so both clean targets and both degraded conditions retain one shared time origin. Shot identifiers and timestamps also provide the alignment interface for subsequent high-resolution refinement.

\clearpage
\subsection{Joint Online Degradation}
\label{app:ov_degradation}
Algorithm~\ref{alg:ov_degrade} expands the two independent stochastic branches of $\mathcal D$. The visual pipeline selects a heavy, light, or mixed spatial-degradation path before adding temporally structured film artifacts. The audio pipeline applies bandwidth, noise, distortion, and spectral operations without changing the window's temporal placement.

\begin{algorithm}[H]
\caption{Generate aligned degraded conditions and severity metadata}
\label{alg:ov_degrade}
\small
\begin{algorithmic}[1]
\REQUIRE Clean aligned window $(\mathbf V^h,\mathbf A^h)$; visual/audio operator configurations
\ENSURE $(\mathbf V^l,\mathbf A^l,d_v,d_a)$ on the clean window's timeline
\ovphase{Video: spatial degradation followed by film artifacts}
\STATE Convert video to $[0,1]$; sample independent video and audio random states
\STATE Select heavy two-pass, light one-pass, or mixed degradation with probabilities $0.70,0.05,0.25$
\STATE For each selected spatial path, apply sharpening, blur, resize, noise, and compression
\STATE In the heavy path, perform a second resize/noise/compression pass with its blur gate
\STATE Downsample and resize back to the target grid using the selected degradation factor
\IF{the mixed path is selected}
  \STATE Combine light and heavy outputs with one temporally shared CutMix mask
\ENDIF
\STATE Apply probability-gated exposure reduction and temporal flicker
\STATE Apply temporally structured scratches and dust; optionally add broad abrasion bands
\STATE Convert to grayscale with probability $0.5$; clamp and return RGB values to $[-1,1]$
\STATE $\mathbf V^l\leftarrow$ resulting video; record active operators and sampled parameters
\ovphase{Audio: independent probability gates on the same waveform interval}
\STATE $\mathbf A^l\leftarrow\mathbf A^h$
\FOR{each enabled audio operator, in pipeline order}
  \STATE Sample its activation gate and strength from the configured range
  \IF{the operator is active}
    \STATE Apply the selected waveform operator while retaining the original time interval
    \STATE Record the operator identity and sampled strength
  \ENDIF
\ENDFOR
\STATE Clamp audio to $[-1,1]$ and retain its original sample count
\STATE Compute $d_v,d_a\in[0,1]$ from normalized per-modality degradation metadata
\RETURN $(\mathbf V^l,\mathbf A^l,d_v,d_a)$
\end{algorithmic}
\end{algorithm}

Spatial parameters and the CutMix region are shared across a clip; scratch trajectories, dust lifetimes, and flicker vary over time according to their temporal operators. Thus, spatial augmentation does not independently scramble the geometry of successive frames. The probabilities and strength ranges for the selected degradation profile determine the resulting corruption; \tref{tab:degrad_params} reports the principal ranges. A codec-based high-resolution refiner profile and the archival-film profile are distinct configurations.

The audio operators cover bandwidth filtering, hiss/rumble, clipping, clicks, codec artifacts, spectral smoothing, harmonic thinning, dynamic compression, repetition, and phase discontinuity. The metadata-derived severity is computed before text or LQ dropout and is kept fixed when a condition is removed. Consequently, a condition mask changes the information available to the model without changing which degradation generated the target pair. This is the separation used in the mixture objective of \eref{eq:dual_condition_objective}.

\clearpage
\subsection{Dual-Condition Sampling and Caption Annealing}
\label{app:ov_conditioning}
Algorithm~\ref{alg:ov_condition} implements \eref{eq:dual_condition_annealing} without interpolating text embeddings. It samples a discrete caption-versus-restoration-prompt choice only when both text and LQ are retained. A single LQ-retention decision controls the video and audio condition branches together; the text-retention decision is independent.

\begin{algorithm}[H]
\caption{Sample the T2AV--TAV2AV--AV2AV conditioning regime}
\label{alg:ov_condition}
\small
\begin{algorithmic}[1]
\REQUIRE Caption $y$; LQ tokens $(\mathbf c_v,\mathbf c_a)$; update $k$; annealing horizon $K_{\rm ann}$
\REQUIRE Cached caption, restoration, and null text embeddings $\mathbf e_y,\mathbf e_{\rm rest},\mathbf e_\emptyset$
\ENSURE Text embedding $\mathbf e$ and condition tokens $(\tilde{\mathbf c}_v,\tilde{\mathbf c}_a)$
\STATE $\lambda_k\leftarrow\max(1-k/K_{\rm ann},0)$
\STATE Independently sample $A\sim\operatorname{Bernoulli}(0.9)$ and $T\sim\operatorname{Bernoulli}(0.9)$
\IF{$T=0$}
  \STATE $\mathbf e\leftarrow\mathbf e_\emptyset$
\ELSIF{$A=0$}
  \STATE $\mathbf e\leftarrow\mathbf e_y$
\ELSE
  \STATE Sample $M\sim\operatorname{Bernoulli}(\lambda_k)$
  \STATE $\mathbf e\leftarrow\mathbf e_y$ if $M=1$, otherwise $\mathbf e\leftarrow\mathbf e_{\rm rest}$
\ENDIF
\FOR{each modality $m\in\{v,a\}$}
  \IF{$A=1$}
    \STATE Sample $\rho_m\sim\mathcal U(0.4,0.6)$ and $\boldsymbol\eta_m\sim\mathcal N(\mathbf0,\mathbf I)$
    \STATE $\tilde{\mathbf c}_m\leftarrow\mathbf c_m+\rho_m\boldsymbol\eta_m$
  \ELSE
    \STATE $\tilde{\mathbf c}_m\leftarrow\mathbf0$ and disable its condition-projection contribution
  \ENDIF
\ENDFOR
\RETURN $(\mathbf e,\tilde{\mathbf c}_v,\tilde{\mathbf c}_a)$
\end{algorithmic}
\end{algorithm}

Here $A$ and $T$ are retention masks, not drop probabilities. At the start, their four joint regimes have masses $0.81$, $0.09$, $0.09$, and $0.01$. At the end, the LQ-present, text-present regime uses $\mathbf e_{\rm rest}$; the no-LQ regime continues to use caption/null text. With $K_{\rm ann}=0.3K$ for $K$ adaptation updates, the remaining updates train at the fixed-prompt endpoint while preserving rehearsal.

Condition dropout must remove the entire learned condition contribution, including any bias. Multiplying only the raw condition by zero would leave a learned projection bias active. Conditioning noise is sampled before the forward pass and held fixed within that evaluation. At inference, both modalities are retained, the fixed restoration prompt is selected internally, and $\rho_m=0.5$.

\clearpage
\subsection{Joint DiT Forward Pass and Reduced-Resolution Injection}
\label{app:ov_forward}
The base model retains the pretrained audio-video transformer, position encoding, text-conditioning pathway, and cross-modal interactions. Split the expanded input matrix as $W_m=[W_m^{x}\;W_m^{c}]$ along its input columns. Its channel-concatenation implementation is equivalently
\begin{equation}
 \mathbf h_m=W_m^{x}\tilde{\mathbf x}_m+W_m^{c}\tilde{\mathbf c}_m+\mathbf b_m.
 \label{eq:ov_split_projection}
\end{equation}
The pretrained $W_m^x$ and bias are retained, while the added condition columns start at zero. The separate-projection form makes the condition gate and Flash's reduced-resolution branch explicit.

\begin{algorithm}[H]
\caption{Joint audiovisual prediction with the base or Flash condition interface}
\label{alg:ov_forward}
\small
\begin{algorithmic}[1]
\REQUIRE Noisy target tokens $(\mathbf z_v,\mathbf z_a)$ on their output grids; LQ pair; text $\mathbf e$
\REQUIRE Noise level $\sigma$; mask $\mathbf M$; mode; supplied condition gates/noise; optional endpoint $r$
\ENSURE Video and audio velocity predictions on the unchanged target grids
\ovphase{Prepare the condition cache once; reuse it on later forward calls}
\IF{base mode}
  \STATE $\mathbf c_v\leftarrow P_v(E_v(\mathbf V^l))$ on the target-resolution grid
\ELSE
  \STATE $\mathbf V^l_{\downarrow}\leftarrow\operatorname{Downsample}_{s}(\mathbf V^l)$, $s=4$ (quarter resolution per spatial dimension)
  \STATE $\mathbf c_v\leftarrow P_v(E_v^{\rm fast}(\mathbf V^l_{\downarrow}))$ \COMMENT{$E_v^{\rm fast}\equiv E_\eta$ from Algorithm~\ref{alg:ov_turbo}}
\ENDIF
\STATE $\mathbf c_a\leftarrow P_a(E_a(\mathbf A^l))$ using the original audio path
\STATE Apply the supplied condition gates/noise once and cache $(\tilde{\mathbf c}_v,\tilde{\mathbf c}_a,\mathbf e)$
\ovphase{One condition projection and the inherited joint backbone}
\STATE $\mathbf h_v\leftarrow W_v^x\mathbf z_v+\mathbf b_v$; $\mathbf h_a\leftarrow W_a^x\mathbf z_a+\mathbf b_a$
\STATE Add $W_v^c\tilde{\mathbf c}_v$ in base mode, or $\mathcal I_\gamma(\tilde{\mathbf c}_v;\mathcal G_v)$ in Flash mode
\STATE Add $W_a^c\tilde{\mathbf c}_a$ to the audio stream
\STATE Assign time $0$ to anchored video tokens and $\sigma$ to all other target tokens
\STATE Form inherited positional and text-conditioning inputs; include $(\sigma,r)$ for the interval model
\FOR{each pretrained joint transformer block}
  \STATE Update video and audio features with self-attention, text attention, and feed-forward operations
  \STATE Apply the inherited gated audio-to-video and video-to-audio interactions
\ENDFOR
\STATE Apply the modality-specific output projections; remove fixed-anchor velocity entries
\RETURN $(\hat{\mathbf u}_v,\hat{\mathbf u}_a)$
\end{algorithmic}
\end{algorithm}

$\mathcal G_v=(T_z,H_z,W_z)$ is the high-resolution target grid. The single learned injection layer $\mathcal I_\gamma$ maps the reduced-resolution condition representation to a contribution with the target stream's token layout and hidden width. Its input-grid alignment is part of this interface; it does not downsample the target state. Audio retains its original encoder, token layout, and projection. Once prepared, the condition cache is consumed directly by the projection/backbone phase. SFT uses the cache supplied by Algorithm~\ref{alg:ov_condition}; denoising steps and JVP evaluations reuse their fixed cache without re-encoding or resampling condition noise.

\clearpage
\subsection{Conditional SFT, Anchoring, and Waveform Supervision}
\label{app:ov_sft}
Algorithm~\ref{alg:ov_sft} combines the condition sampler with the shared-time rectified-flow objective. Parameters $\theta$ comprise the trainable adaptation parameters and condition projections; pretrained video/audio VAEs, the text encoder, and the vocoder remain frozen. First-frame anchoring affects only the video state, its token times, and its loss mask.

\begin{algorithm}[H]
\caption{Train the joint restoration model by conditional SFT}
\label{alg:ov_sft}
\small
\begin{algorithmic}[1]
\REQUIRE Corpus $\mathcal C$; pretrained joint DiT; frozen encoders/decoders; $K$ updates
\ENSURE Conditional restoration model $G_\theta$
\STATE Initialize expanded projections and rank-384 LoRA adaptation parameters
\FOR{$k=0,\ldots,K-1$}
  \STATE Clear gradients of the trainable adaptation parameters
  \STATE Sample a minibatch of synchronized clean windows and captions
  \STATE Generate $(\mathbf V^l,\mathbf A^l,d_v,d_a)$ with Algorithm~\ref{alg:ov_degrade}
  \STATE Encode and patchify clean targets $\mathbf x_m$ and LQ conditions $\mathbf c_m$, $m\in\{v,a\}$
  \STATE Select text and LQ conditions with Algorithm~\ref{alg:ov_condition}
  \STATE Sample one $\sigma\sim\mathcal U(0,1)$ per pair and independent $\boldsymbol\epsilon_v,\boldsymbol\epsilon_a$
  \STATE $\mathbf z_m\leftarrow(1-\sigma)\mathbf x_m+\sigma\boldsymbol\epsilon_m$; $\mathbf u_m\leftarrow\boldsymbol\epsilon_m-\mathbf x_m$
  \STATE Independently enable the first-frame anchor for each sample with probability $0.5$
  \STATE Let $\mathbf M$ select its first latent video frame, or be zero if anchoring is disabled
  \STATE $\mathbf z_v\leftarrow\mathbf M\odot\mathbf x_v+(1-\mathbf M)\odot\mathbf z_v$
  \STATE Predict $(\hat{\mathbf u}_v,\hat{\mathbf u}_a)$ from the cached conditions using Algorithm~\ref{alg:ov_forward}'s backbone phase
  \STATE Compute per-sample masked video and full audio errors using Eqs.~\ref{eq:video_loss}--\ref{eq:audio_loss}
  \STATE Compute $w_v,w_a$ from $d_v,d_a$ and normalize each modality's weights to mean one over the 64-sample global batch
  \ovphase{Differentiable waveform supervision on a small sub-batch}
  \STATE Select an audio sub-batch $\mathcal S$; $\hat{\mathbf x}_a\leftarrow\mathbf z_a-\sigma\hat{\mathbf u}_a$
  \STATE $\hat{\mathbf A}\leftarrow\operatorname{Vocoder}(D_a(P_a^{-1}(\hat{\mathbf x}_a)))$ for samples in $\mathcal S$
  \STATE Average waveform channels, align valid lengths, and evaluate multi-resolution STFT loss
  \STATE $\mathcal L\leftarrow\ovmean_B(w_v\mathcal L_v+\lambda_a w_a\mathcal L_a)+\lambda_{\rm stft}\mathcal L_{\rm stft}$
  \STATE Backpropagate through the DiT and the frozen audio decoding path
  \STATE Clip the gradient norm to 1.0; update $\theta$ with AdamW
\ENDFOR
\RETURN $G_\theta$
\end{algorithmic}
\end{algorithm}

For FFT/window sizes $n\in\{512,1024,2048\}$ and hops $n/4$, let $S_n$ and $\hat S_n$ be clean and predicted STFT magnitudes. The stabilized spectral term averages
\begin{equation}
 \frac{\|\hat S_n-S_n\|_F}{\max(\|\hat S_n\|_F+\|S_n\|_F,\epsilon_s)}
 +\ovmean\left|\log\max(\hat S_n,\epsilon_s)-\log\max(S_n,\epsilon_s)\right|,
 \quad \epsilon_s=10^{-5}.
 \label{eq:ov_stft_detail}
\end{equation}
Use $\lambda_a=1$ and $\lambda_{\rm stft}=0.003$. The audio decoder and vocoder receive gradients with respect to their inputs, although their weights are fixed. Computing them under a no-gradient context would remove waveform supervision from the DiT. The STFT term is averaged over $\mathcal S$ and is not multiplied by the modality severity weight.

\clearpage
\subsection{AV MeanFlow with the Improved Velocity Objective}
\label{app:ov_imf}
We adapt the velocity-loss formulation of improved MeanFlow (iMF)~\citep{imeanflow} to distillation from the frozen conditional restoration model. The teacher supplies an instantaneous-velocity target $\mathbf b_T$; the student predicts average velocity $\mathbf U_\phi$. The teacher may use the full-resolution LQ condition $\mathbf q_T$, while the student's $\mathbf q_S$ uses the reduced-resolution video interface. Both predict on the same target latent grid.

Following iMF's boundary-velocity construction, the JVP direction is the student's own instantaneous prediction,
\begin{align}
 \mathbf b_S&=\ovsg[\mathbf U_\phi(\mathbf z_\sigma,\sigma,\sigma;\mathbf q_S)],\notag\\
 \mathbf J&=\operatorname{JVP}_{(\mathbf z,\sigma,r)}
       [\mathbf U_\phi;(\mathbf b_S,1,0)],\qquad
 \mathbf V_\phi=\mathbf U_\phi+(\sigma-r)\ovsg[\mathbf J].
 \label{eq:ov_imf_compound}
\end{align}
Thus the compound prediction uses only information available to the student. Full-resolution teacher information enters the regression target, not the JVP direction. The derivative holds $r$, text, LQ observations, condition noise, and the anchor fixed. All student evaluations return zero velocity on fixed anchor entries; their state tangents and token times are also zero. At $r=\sigma$, the compound prediction reduces to the boundary velocity.

\begin{algorithm}[H]
\caption{AV iMF velocity distillation}
\label{alg:ov_imf}
\small
\begin{algorithmic}[1]
\REQUIRE Frozen SFT teacher; student $\mathbf U_\phi$; interval parameters $(\mu,\nu,p_{\rm diag})$
\REQUIRE Teacher guidance $w_T$; adaptive-loss parameters $(\epsilon_\ell,p_\ell)$
\ENSURE One-step joint predictor $\mathbf U_\phi$
\STATE Initialize the student from the restoration model; keep its encoders frozen
\STATE Include interval-conditioning parameters and injection-layer parameters $\gamma$ in trainable $\phi$
\FOR{each update}
  \STATE Clear student gradients
  \STATE Sample paired data, clean target tokens $\mathbf x$, and teacher/student conditions $\mathbf q_T,\mathbf q_S$
  \STATE Independently sample $a,b\sim\operatorname{sigmoid}(\mathcal N(\mu,\nu^2))$
  \STATE $\sigma\leftarrow\max(a,b)$; $r\leftarrow\min(a,b)$; set $r=\sigma$ for fraction $p_{\rm diag}$ of the minibatch
  \STATE Sample independent video/audio noise; $\mathbf z_\sigma\leftarrow(1-\sigma)\mathbf x+\sigma\boldsymbol\epsilon$
  \STATE Apply the sampled clean first-frame anchor and its zero-time token mask
  \STATE Evaluate teacher fixed-prompt and null-text velocities with the same retained LQ condition
  \STATE $\mathbf b_T\leftarrow\ovsg[\mathbf v_{T,\emptyset}+w_T(\mathbf v_{T,\rm rest}-\mathbf v_{T,\emptyset})]$
  \STATE $\mathbf b_S\leftarrow\ovsg[\mathbf U_\phi(\mathbf z_\sigma,\sigma,\sigma;\mathbf q_S)]$
  \STATE Set anchored entries of $\mathbf b_T$ and $\mathbf b_S$ to zero
  \STATE Compute $(\mathbf U,\mathbf J)$ with a JVP in direction $(\mathbf b_S,1,0)$, using the FA4 kernel
  \STATE $\mathbf V_\phi\leftarrow\mathbf U+(\sigma-r)\ovsg[\mathbf J]$
  \STATE $e\leftarrow\operatorname{MSE}_{v,\neg\mathbf M}(\mathbf V_\phi,\mathbf b_T)+\lambda_a\operatorname{MSE}_a(\mathbf V_\phi,\mathbf b_T)$
  \STATE $\mathcal L_{\rm iMF}\leftarrow\ovmean_B\!\left[e/\ovsg[(e+\epsilon_\ell)^{p_\ell}]\right]$
  \STATE Obtain $\mathcal L_{\rm G,3D}$ and perform the discriminator update using Algorithm~\ref{alg:ov_gan}
  \STATE Update $\phi$ on $\mathcal L_{\rm iMF}+\lambda_{\rm G}\mathcal L_{\rm G,3D}$
\ENDFOR
\RETURN $\mathbf U_\phi$
\end{algorithmic}
\end{algorithm}

The interval sampler and detached adaptive weighting follow the official iMF construction. Its public reference uses $(\mu,\nu,p_{\rm diag})=(-0.4,1.0,0.5)$ and $(\epsilon_\ell,p_\ell)=(0.01,1.0)$; these remain explicit configuration inputs here. Unlike iMF's from-scratch data-velocity regression, our target is the frozen teacher velocity. We use the boundary formulation without an auxiliary velocity head. Teacher guidance is distilled at $w_T$, so inference needs one student evaluation rather than a second null-text evaluation.

\clearpage
\subsection{Three-Dimensional Adversarial Supervision}
\label{app:ov_gan}
A 3D PatchGAN $C_\omega$ scores spatiotemporal RGB patches, exposing the student to both spatial appearance and short-term temporal artifacts. The adversarial branch evaluates an actual one-step endpoint from pure noise, rather than treating a partial denoising prediction at a random $\sigma$ as the deployed output. Let $\mathcal P$ select a contiguous video crop and let $\operatorname{Dec}_v$ denote the frozen video decoding path used during distillation. Then
\begin{align}
 \hat{\mathbf V}&=\operatorname{Dec}_v\!\left(P_v^{-1}
    [\boldsymbol\epsilon_v-\mathbf U_\phi^v(\boldsymbol\epsilon,1,0;\mathbf q_S)]\right),\notag\\
 \mathcal L_{\rm G,3D}&=-\ovmean C_\omega(\mathcal P(\hat{\mathbf V})),\notag\\
 \mathcal L_{\rm D,3D}&=\tfrac12\ovmean\left[\max(0,1-C_\omega(\mathcal P(\mathbf V^h)))
             +\max(0,1+C_\omega(\ovsg[\mathcal P(\hat{\mathbf V})]))\right].
 \label{eq:ov_gan_losses}
\end{align}
The endpoint expression above is for an unanchored sample; anchored entries are overwritten by the fixed clean boundary as in Algorithm~\ref{alg:ov_gan}.

\begin{algorithm}[H]
\caption{One-step video realism loss and 3D discriminator update}
\label{alg:ov_gan}
\small
\begin{algorithmic}[1]
\REQUIRE Student $\mathbf U_\phi$; frozen video decoder; 3D discriminator $C_\omega$; aligned clean/LQ sample
\REQUIRE Adversarial start/update schedule; spatial crop and temporal crop length
\ENSURE Generator loss $\mathcal L_{\rm G,3D}$ and updated discriminator parameters
\IF{the adversarial branch is inactive at this update}
  \RETURN $0$
\ENDIF
\STATE Draw fresh video/audio Gaussian noise on the target grids
\STATE If anchored, replace the first video latent frame by the clean anchor and set its token time to zero
\STATE Predict the joint one-step endpoint with $\mathbf U_\phi(\cdot,1,0;\mathbf q_S)$
\STATE Restore fixed anchor entries and decode the generated video with gradients to its latent
\STATE Select real and generated crops using the same spatial/temporal sampling rule
\ovphase{Discriminator update: no gradients enter the generator}
\STATE Enable gradients for $\omega$ and clear discriminator gradients; detach the generated crop
\STATE Compute the hinge loss in \eref{eq:ov_gan_losses}
\STATE Backpropagate $\mathcal L_{\rm D,3D}$ and update only $\omega$ on scheduled discriminator steps
\ovphase{Generator update: gradients pass through the discriminator and decoder}
\STATE Hold $\omega$ fixed and clear student gradients; evaluate the undetached generated crop with $C_\omega$
\STATE $\mathcal L_{\rm G,3D}\leftarrow-\ovmean C_\omega(\mathcal P(\hat{\mathbf V}))$
\RETURN $\mathcal L_{\rm G,3D}$ to Algorithm~\ref{alg:ov_imf}
\end{algorithmic}
\end{algorithm}

The video decoder is fixed during DiT distillation. Its input gradient is retained on generator steps, whereas real videos and detached fake videos update only the discriminator. Crops remain contiguous in time; independently sampling unrelated frames would remove the temporal meaning of a 3D discriminator. The anchor, when present, is fixed in both branches and is excluded from crop-only supervision if the selected crop consists solely of the copied boundary.

This adversarial loss supplements joint velocity distillation; it does not replace the audio objective or introduce an audio discriminator. Its coefficient $\lambda_{\rm G}$, activation step, update frequency, and crop size are explicit training inputs. Because the adversarial term also affects the student optimum, one-step learning is an approximation to the restoration flow with an additional video-realism objective.

\clearpage
\subsection{Turbo VAE Distillation}
\label{app:ov_turbo}
Turbo VAE adopts the training method and code of Ultra Flash~\citep{luxury2026ultraflash}, adapted to the 128-channel LTX video-latent interface. Its encoder and decoder have separate targets even when optimized in the same iteration. In particular, the decoder receives a frozen teacher latent, rather than the current student's encoder output; decoder reconstruction gradients therefore do not train the encoder.

\begin{algorithm}[H]
\caption{Distill the lightweight video encoder and decoder}
\label{alg:ov_turbo}
\small
\begin{algorithmic}[1]
\REQUIRE HQ video patches; frozen teacher posterior $q_T$ and channel statistics $(\boldsymbol\mu_T,\boldsymbol\sigma_T)$
\REQUIRE Tiny encoder $E_\eta$, decoder $D_\psi$, separate 3D discriminator $C_\xi$; loss weights and schedules
\ENSURE Turbo encoder $E_\eta$ and decoder $D_\psi$
\FOR{each training iteration}
  \STATE Clear encoder and decoder gradients
  \STATE Sample an HQ spatiotemporal patch $\mathbf X$ and normalize RGB values to $[-1,1]$
  \ovphase{Teacher latent target: frozen and normalized}
  \STATE Sample $\mathbf z^{\rm raw}_T\sim q_T(\cdot\mid\mathbf X)$ without gradients
  \STATE $\mathbf z_T\leftarrow\ovsg[(\mathbf z^{\rm raw}_T-\boldsymbol\mu_T)/\boldsymbol\sigma_T]$
  \ovphase{Independent encoder and decoder distillation branches}
  \STATE $\hat{\mathbf z}\leftarrow E_\eta(\mathbf X)$; align valid latent extents
  \STATE $\mathcal L_E\leftarrow\ell_{\rm lat}(\hat{\mathbf z},\mathbf z_T)$
  \STATE $\hat{\mathbf X}\leftarrow D_\psi(\mathbf z_T)$; remove the decoder's causal warm-up prefix
  \STATE Align decoded and clean valid extents; apply the same comparison crop
  \STATE $\mathcal L_{\rm pix}\leftarrow\ell_{\rm pix}(\hat{\mathbf X},\mathbf X)$
  \STATE $\mathcal L_{\rm per}\leftarrow\ovmean_{b,t}\operatorname{LPIPS}(\hat{\mathbf X}_{b,t},\mathbf X_{b,t})$
  \STATE $\mathcal L_{\rm base}\leftarrow\lambda_E\mathcal L_E+\lambda_{\rm pix}\mathcal L_{\rm pix}+\lambda_{\rm per}\mathcal L_{\rm per}$
  \IF{a discriminator update is scheduled}
    \STATE Enable gradients for $\xi$; clear its gradients and update only $\xi$ on the 3D hinge loss
  \ENDIF
  \IF{a generator adversarial update is scheduled after discriminator warm-up}
    \STATE Hold $\xi$ fixed, retaining input gradients; $\mathcal L_{\rm adv}\leftarrow-\ovmean C_\xi(\hat{\mathbf X})$
    \STATE $\mathcal L_{\rm base}\leftarrow\mathcal L_{\rm base}+\lambda_{\rm VAE,G}\mathcal L_{\rm adv}$
  \ENDIF
  \STATE Update $(\eta,\psi)$ on $\mathcal L_{\rm base}$; keep the teacher and perceptual network fixed
\ENDFOR
\RETURN $(E_\eta,D_\psi)$
\end{algorithmic}
\end{algorithm}

$\ell_{\rm lat}$ and $\ell_{\rm pix}$ are the configured elementwise reconstruction losses (MSE in the LTX reference implementation). The latent term is measured in the teacher's normalized coordinates. LPIPS includes its wrapper's RGB-range preprocessing and is averaged over frames; the reference wrapper maps $[-1,1]$ to clipped $[0,1]$ before evaluation. The adversarial coefficient is separate from any frequency-domain loss. The VAE discriminator $C_\xi$ is distinct from the DiT-distillation discriminator $C_\omega$.

Both distillation branches use the same HQ patch. Downsampling the observed LQ video is a deployment operation before $E_\eta$, not an instruction to train $E_\eta(\mathbf V^l)$ against a high-resolution teacher latent. The teacher's 128-channel normalization convention must also be respected at inference; it must not be applied twice. Audio continues to use its original VAE and vocoder.

\clearpage
\subsection{Causal Memory Blocks and Video-Latent Geometry}
\label{app:ov_memory}
The lightweight video path uses spatial convolutions with causal feature memory. Let $\mathbf H_t$ be a block's input features. With the reference implementation's replicated initial state, its temporal memory is $\mathbf Q_0=\mathbf H_0$ and $\mathbf Q_t=\mathbf H_{t-1}$ for $t>0$. The block computes
\begin{equation}
 \mathbf H'_t=\operatorname{ReLU}\!\left(
       F([\mathbf H_t;\mathbf Q_t])+S(\mathbf H_t)\right),
 \label{eq:ov_memblock}
\end{equation}
where $F$ is a three-convolution spatial stack with intermediate activations and $S$ is the residual projection. A parallel temporal shift and a sequential cache implement the same causal dependency. The cache stores the previous \emph{input} features of each block, not its updated output.

\begin{algorithm}[H]
\caption{Causal-memory video encoding and decoding}
\label{alg:ov_memory}
\small
\begin{algorithmic}[1]
\REQUIRE Video $\mathbf X$ or normalized latent $\mathbf z$; selected encoder/decoder; causal boundary convention
\ENSURE Encoded 128-channel latent or raw reconstructed RGB video with causal warm-up prefix
\IF{encoding}
  \STATE $\mathbf H\leftarrow\operatorname{ReLU}(\operatorname{Conv}_{\rm in}(\operatorname{PixelUnshuffle}_4(\mathbf X)))$
  \FOR{each of three encoder stages}
    \STATE Causally pad the initial feature if needed; pool temporally by factor two
    \STATE Apply stride-two spatial convolution
    \STATE Apply the stage's causal memory blocks using \eref{eq:ov_memblock}
    \STATE If not the final stage, apply the inter-stage channel-conversion convolution
  \ENDFOR
  \STATE Project features to 128 latent channels
\ELSE
  \STATE $\mathbf H\leftarrow\operatorname{ReLU}(\operatorname{Conv}_{\rm in}(3\tanh(\mathbf z/3)))$
  \FOR{each of three decoder stages}
    \STATE Apply the stage's causal memory blocks using \eref{eq:ov_memblock}
    \STATE Upsample spatially by factor two using convolution and PixelShuffle
    \STATE Expand temporal features by factor two and reshape them into successive frames
    \STATE If not the final stage, apply the inter-stage channel-conversion convolution
  \ENDFOR
  \STATE Apply ReLU, the output projection, and final $4\times$ PixelShuffle to produce RGB
\ENDIF
\ovphase{Equivalent execution of each memory block}
\STATE In parallel mode, construct previous-input features by a causal temporal shift
\STATE In sequential mode, read the block's cached previous input, evaluate the block, then cache its current input
\STATE Initialize each block at the sequence boundary using the same replicated-frame convention
\RETURN the raw output tensor; its caller handles temporal alignment
\end{algorithmic}
\end{algorithm}

The LTX-compatible construction has spatial compression $4\times2^3=32$ and temporal compression $2^3=8$. The reference encoder has three stages of width 128; the decoder widths are 512, 256, and 128, with three memory blocks per stage. A 121-frame window therefore has 16 latent video frames. Here $D_\psi$ denotes raw decoder output, including seven warm-up frames; the training or inference wrapper removes this prefix exactly once before comparison or emission.

These feature caches belong to VAE processing. They are separate from the restored-frame boundary used to connect DiT windows, and do not imply persistent DiT key-value caches. Encoder and decoder calls must use the same temporal padding/cropping convention during training and inference. Spatial tiling, when needed for memory, must preserve the selected overlap and valid-region rules.

\clearpage
\subsection{Windowed Base and One-Step Flash Inference}
\label{app:ov_inference}
Algorithm~\ref{alg:ov_inference} distinguishes the unanchored first window from subsequent anchored windows. Neighboring 121-frame windows share one boundary frame. The restored endpoint of one window occupies that boundary in the next window, and is emitted only once. Audio is selected and assembled by timestamps on the same timeline.

\begin{algorithm}[H]
\caption{Restore an audio-video recording with bounded window memory}
\label{alg:ov_inference}
\small
\begin{algorithmic}[1]
\REQUIRE Degraded recording; base or Flash model; frozen decoding modules; internal prompt $\mathbf e_{\rm rest}$
\ENSURE Restored video and audio on the input timeline
\STATE Construct synchronized 121-frame windows with one shared video boundary frame
\STATE $\mathbf a_{\rm prev}\leftarrow\emptyset$; initialize output video and audio timelines
\FOR{each window $j$ in temporal order}
  \STATE Read the LQ video and the audio interval corresponding to that window
  \STATE Set both retention gates to one and select condition-noise amplitude $0.5$
  \STATE Execute only Algorithm~\ref{alg:ov_forward}'s condition-cache phase once, using base or Flash video encoding
  \STATE Draw video/audio Gaussian target states on the output grids
  \IF{$\mathbf a_{\rm prev}\ne\emptyset$}
    \STATE Encode the previous restored frame with the selected video encoder as the new first-frame anchor
    \STATE Overwrite its video latent tokens, set their token times to zero, and fix their update mask
  \ENDIF
  \IF{base mode}
    \STATE Set a descending 15-step Euler schedule $1=\sigma_0>\cdots>\sigma_{15}=0$
    \FOR{$i=0,\ldots,14$}
      \STATE Evaluate fixed-prompt and null-text velocities with identical LQ conditions and anchor
      \STATE $\mathbf v\leftarrow\mathbf v_\emptyset+3(\mathbf v_{\rm rest}-\mathbf v_\emptyset)$
      \STATE $\mathbf z\leftarrow\mathbf z+(\sigma_{i+1}-\sigma_i)\mathbf v$ on unanchored entries
      \STATE Restore the fixed anchor entries
    \ENDFOR
  \ELSE
    \STATE $\mathbf z_0\leftarrow\mathbf z_1-\mathbf U_\phi(\mathbf z_1,1,0;\mathbf q_S)$ on unanchored entries
    \STATE Restore the fixed anchor entries; use the distilled guidance without a second student call
  \ENDIF
  \STATE Decode video using the base decoder or Turbo decoder; decode audio using the original VAE and vocoder
  \STATE Crop padding to the valid video frames and align decoded audio to the window's valid timestamps
  \STATE Save the last restored RGB frame as $\mathbf a_{\rm prev}$
  \STATE Append new video frames once; write only the not-yet-emitted audio interval
\ENDFOR
\STATE Mux restored audio and video using the original frame rate and common timestamps
\RETURN the restored recording
\end{algorithmic}
\end{algorithm}

The first window has no clean boundary and denoises all its video tokens. Subsequent windows use a generated boundary, so the prediction should not be described as conditioned on a ground-truth frame at deployment. A final short window is padded only for model input; the padded frames and waveform samples are removed before output assembly. Audio overlap is handled by timestamps rather than by assuming that one latent audio token equals one video frame.

Base-model guidance retains LQ in both text branches, as in \eref{eq:dual_cfg}. Flash instead uses the fixed-guidance field learned during distillation. The joint DiT always predicts both modalities; Turbo changes the video encoding/decoding path and reduced-resolution condition interface, while the audio encoder, decoder, and vocoder remain unchanged.

\endgroup

\clearpage
\section{Additional Ablations and Controlled Diagnostics}
\label{sec:s12}

This section reports additional ablations and diagnostics that address specific mechanisms of the restoration model and its deployment variant. All numbers are on the Controlled track (71 clips) unless stated otherwise; metric configurations follow \sref{sec:setup}.

\subsection{Severity-weighting interaction term}
\label{sec:s12-weight}
The degradation-aware reweighting in \eref{eq:loss_reweight} includes a multiplicative interaction term $\gamma d_vd_a$ that emphasizes samples degraded in both modalities simultaneously. \tref{tab:weight_interaction} isolates this term by retraining the final $K_3=33{,}000$ AV2AV steps with $\gamma\in\{0,2.0\}$ while holding all other hyperparameters fixed. The interaction term benefits the jointly degraded Controlled subset ($55/71$, approximately $77.5\%$ of clips) substantially more than the single-modality-degraded subset, which is exactly the behavior the term is designed to produce.

\begin{table*}[t]
\centering
\caption{Severity-weighting interaction ablation ($K_3$ retrained, all else fixed). The right-hand columns report the 55 jointly degraded clips; the left-hand columns report all 71 clips. Results for the 16 single-modality clips are summarized below. $\gamma=2.0$ is the default.}
\label{tab:weight_interaction}
\fontsize{7.6}{8.2}\selectfont
\setlength{\tabcolsep}{2pt}
\begin{tabularx}{\textwidth}{@{}l*{3}{Y}*{3}{Y}@{}}
\toprule
& \multicolumn{3}{c}{\textbf{All clips ($n{=}71$)}} & \multicolumn{3}{c}{\textbf{Joint-degradation subset ($n{=}55$)}} \\
\cmidrule(lr){2-4}\cmidrule(lr){5-7}
\textbf{Setting} & MUSIQ$\uparrow$ & DNSMOS$\uparrow$ & FAD$\downarrow$ & MUSIQ$\uparrow$ & DNSMOS$\uparrow$ & FAD$\downarrow$ \\
\midrule
$\gamma=0$ (no interaction) & 70.62 & 2.61 & 6.71 & 70.08 & 2.57 & 6.94 \\
\rowcolor{oursrow}
$\gamma=2.0$ (default) & \textbf{71.17} & \textbf{2.70} & \textbf{6.30} & \textbf{70.81} & \textbf{2.68} & \textbf{6.48} \\
\bottomrule
\end{tabularx}
\end{table*}

\begin{figure}[!htbp]
\centering
\includegraphics[width=0.82\linewidth]{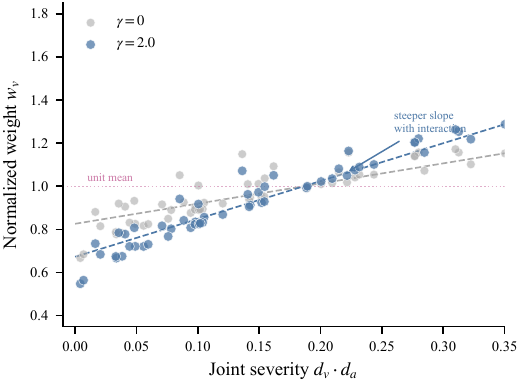}
\caption{Normalized severity weights $w_v$ versus the joint severity $d_vd_a$ over a representative training batch, under $\gamma=0$ (gray) and $\gamma=2.0$ (blue). The interaction term increases the weight on jointly degraded samples without changing the unit-mean constraint.}
\label{fig:weight_scatter}
\end{figure}

\noindent On the jointly degraded subset the interaction term improves MUSIQ by $+0.73$ and DNSMOS by $+0.11$; on the single-modality subset ($n{=}16$) the effect is negligible (MUSIQ $-0.07$, within seed noise per \tref{tab:seed_robustness}; DNSMOS $+0.02$; FAD $-0.24$). The term concentrates its benefit where it should---jointly degraded clips---without harming single-modality cases. \fref{fig:weight_scatter} visualizes the resulting weight distribution.

\subsection{Gate dynamics during adaptation}
\label{sec:s12-gates}
\begin{figure}[!htbp]
\centering
\includegraphics[width=0.82\linewidth]{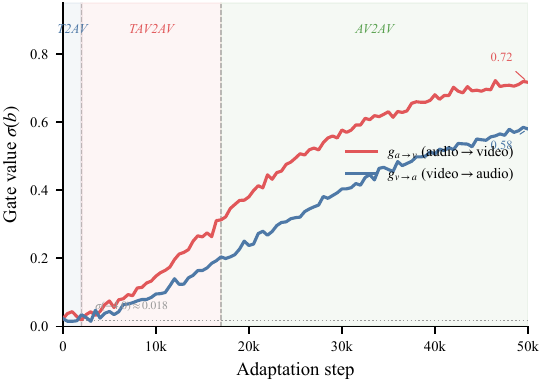}
\caption{Evolution of the learnable cross-modal gates $\sigma(b)$ over the $50{,}000$ adaptation steps. Both gates open monotonically from the near-zero initialization; the audio$\to$video gate ($g_{a\to v}$) reaches a higher plateau than the video$\to$audio gate ($g_{v\to a}$), reflecting the asymmetric dependence measured in \tref{tab:crossmodal_ablation}.}
\label{fig:gate_dynamics}
\end{figure}

\sref{sec:s10-crossmodal} ablates the cross-modal gates at the endpoint. \fref{fig:gate_dynamics} reports how the gate values evolve during the $50{,}000$ adaptation steps, starting from $\sigma(-4.0)\approx0.018$. The audio$\to$video gate grows faster during TAV2AV and reaches a higher plateau ($\approx0.72$) than video$\to$audio ($\approx0.58$), indicating that the video pathway draws more on acoustic evidence than the reverse---consistent with the endpoint ablation, where disabling audio$\to$video costs more MUSIQ.

\subsection{Resolution-dependent VAE timing}
\label{sec:s12-vae}
\begin{figure*}[t]
\centering
\includegraphics[width=0.78\textwidth]{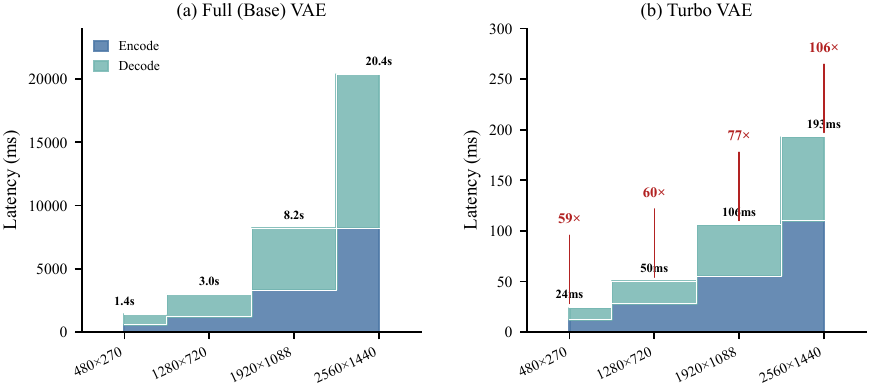}
\caption{Resolution-dependent VAE encoding and decoding times corresponding to \tref{tab:vae_bench}. Left: full VAE. Right: Turbo VAE. Bar totals show combined latency; annotated ratios show the full/Turbo VAE speedup at each resolution.}
\label{fig:resolution_timing}
\end{figure*}

\tref{tab:vae_bench} reports isolated VAE encode$+$decode throughput across four resolutions, and \fref{fig:resolution_timing} separates the encoding and decoding costs. From $480\times270$ to $2560\times1440$, the plotted combined latency increases from approximately $1.4$\,s to $20.4$\,s for the full VAE and from $24$\,ms to $193$\,ms for Turbo VAE. The total latency grows more slowly for Turbo VAE over this tested range, widening the reported speedup from $59.9\times$ to $105.8\times$. These totals describe the measured configurations; they do not determine separate asymptotic scaling laws for the encoder and decoder.

\subsection{Long-sequence chaining beyond the benchmark duration}
\label{sec:s12-longseq}
\begin{figure}[!htbp]
\centering
\includegraphics[width=0.92\linewidth]{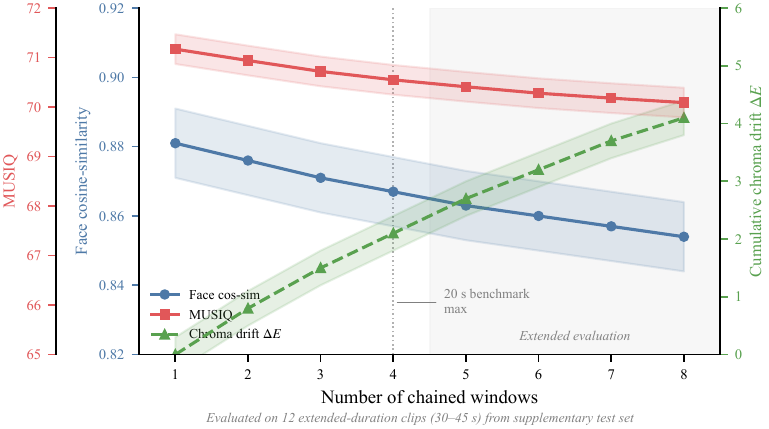}
\caption{Long-sequence chaining metrics on the extended-duration test set ($n{=}12$, $30$--$45$\,s). Face similarity and MUSIQ decline slowly and monotonically; chroma drift grows approximately linearly at $\approx0.55$ $\Delta E$ per window, consistent with the color-drift failure mode in \sref{sec:s2-failure}.}
\label{fig:longseq_metrics}
\end{figure}

The OmniVRBench clips are at most $20$\,s (\tref{tab:bench_stats}), which corresponds to at most four $121$-frame windows at $24$\,fps ($20.17$\,s $\approx 484$ frames). To evaluate whether drift accumulates over longer durations, we construct an additional \textbf{extended-duration test set} of $12$ clips of $30$--$45$\,s, drawn from the same archival sources as the Real track but excluded from all other evaluations. These clips are processed by sliding the $121$-frame window with stride $120$ frames (adjacent windows share exactly one anchor frame), so that $W$ windows cover $120W+1$ frames with no padding and no looping. \fref{fig:longseq_metrics} reports face identity similarity, MUSIQ, and cumulative chroma drift (mean $\Delta E$ in CIE-Lab against the first window's restored frame) as a function of $W$.

\begin{table*}[t]
\centering
\caption{Extended-duration chaining evaluation ($n{=}12$ clips, $30$--$45$\,s). Windows are $121$ frames with stride $120$; $W$ windows cover $120W{+}1$ frames. Face similarity is against the first-window restored face; chroma drift is mean $\Delta E$ (CIE-Lab) against the first-window restored frame.}
\label{tab:longseq}
\fontsize{7.6}{8.2}\selectfont
\begin{tabularx}{\textwidth}{@{}l*{6}{Y}@{}}
\toprule
\textbf{Windows $W$} & 1 & 2 & 3 & 4 & 6 & 8 \\
\textbf{Duration (s)} & 5.0 & 10.0 & 15.0 & 20.0 & 30.0 & 40.0 \\
\midrule
Face cos-sim$\uparrow$ & 0.881 & 0.876 & 0.871 & 0.867 & 0.860 & 0.854 \\
MUSIQ$\uparrow$ & 71.17 & 70.94 & 70.72 & 70.55 & 70.28 & 70.09 \\
Chroma drift $\Delta E\downarrow$ & 0.0 & 0.8 & 1.5 & 2.1 & 3.2 & 4.1 \\
\midrule
\multicolumn{7}{@{}l@{}}{\textit{With per-window anchor reset (upper bound)}} \\
Face cos-sim$\uparrow$ & 0.881 & 0.879 & 0.877 & 0.875 & 0.872 & 0.870 \\
Chroma drift $\Delta E\downarrow$ & 0.0 & 0.3 & 0.5 & 0.7 & 1.0 & 1.3 \\
\bottomrule
\end{tabularx}
\end{table*}

\noindent Over $40$\,s ($8$ windows) face similarity declines by $0.027$ and MUSIQ by $1.08$, both small relative to the gap to the best baseline (\tref{tab:controlled_track}). Chroma drift is the dominant long-horizon effect, growing approximately linearly; resetting the anchor to the first window's restored frame (an oracle upper bound, second block of \tref{tab:longseq}) cuts cumulative drift by ${\sim}70\%$, confirming a chaining artifact rather than a model-level quality regression. This motivates the self-generated-anchor training discussed in \sref{sec:s8} and \sref{sec:limitations}.

\subsection{Audio content preservation beyond intelligibility proxies}
\label{sec:s12-audio}
The closed-set intelligibility pilot in \sref{sec:s1} measures clip-level word accuracy but not content fidelity in an open-vocabulary setting. To complement it, we report an independent open-vocabulary evaluation: two human annotators transcribe the restored audio of $40$ Controlled-track clips (containing speech) with no access to the reference transcript, and we score word error rate (WER) and content-word preservation against the clean reference using the same ASR reference used in the pipeline. \tref{tab:audio_content} reports the results together with insertion/deletion rates, which separate hallucinated content from dropped content.

\begin{table*}[t]
\centering
\caption{Open-vocabulary audio content preservation on the Controlled track ($n{=}40$ speech clips). WER and insertion/deletion rates are computed against the clean reference transcript. ``Numbers/proper nouns'' is a subset analysis of the most fidelity-sensitive tokens.}
\label{tab:audio_content}
\fontsize{7.6}{8.2}\selectfont
\begin{tabularx}{\textwidth}{@{}l*{4}{Y}@{}}
\toprule
\textbf{Method} & WER$\downarrow$ & Insertion rate$\downarrow$ & Deletion rate$\downarrow$ & Numbers/proper nouns acc.$\uparrow$ \\
\midrule
Low-quality input & 0.482 & 0.061 & 0.184 & 0.42 \\
VoiceFixer & 0.311 & 0.058 & 0.092 & 0.61 \\
\rowcolor{oursrow}
OmniVR (ours) & \textbf{0.264} & 0.054 & \textbf{0.071} & \textbf{0.74} \\
\bottomrule
\end{tabularx}
\end{table*}

\noindent OmniVR achieves the lowest WER ($0.264$) and insertion rate, and on the fidelity-sensitive subset of numbers and proper nouns reaches $74\%$ accuracy vs.\ $61\%$ for VoiceFixer. Together with the closed-set result ($94\%$, \sref{sec:s1}), this supports the view that the low STOI/eSTOI reflects envelope-resynthesis sensitivity rather than lost content. The residual WER and the $26\%$ of unrecovered numbers/proper nouns are a genuine limitation, concentrated in clips with $d_a>0.9$, consistent with the over-smoothing failure mode in \sref{sec:s2-failure}.

\subsection{Eligible versus successfully measured sample counts}
\label{sec:s12-counts}
\tref{tab:metric_eligibility} (\sref{sec:s4}) reports eligibility. A small number of eligible clips fail to produce a measurement because a released baseline pipeline runs out of memory or emits too few frames; these are distinct from eligibility. \tref{tab:effective_n} separates the two for every full-reference and distributional metric on the Controlled track.

\begin{table*}[t]
\centering
\caption{Eligible vs.\ successfully measured clip counts on the Controlled track. The baseline is RealBasicVSR for visual rows and VoiceFixer for the audio row. ``Eligible'' is set by the protocol (\tref{tab:metric_eligibility}); ``measured'' is the count that produced a finite metric value. RealBasicVSR shortfalls are due to inference-pipeline OOM on near-4K sources. OmniVR's inference pipeline completes on all 71 clips; its single LPIPS shortfall ($n{=}70$) reflects a metric-computation OOM rather than a model failure (footnote~$c$ in \tref{tab:fr_controlled}).}
\label{tab:effective_n}
\fontsize{7.6}{8.2}\selectfont
\begin{tabularx}{\textwidth}{@{}l*{4}{Y}@{}}
\toprule
\textbf{Metric} & \textbf{Eligible} & \textbf{Baseline measured} & \textbf{OmniVR measured} & \textbf{Common subset} \\
\midrule
LPIPS (visual) & 71 & 70 & 70 & 70 \\
DISTS (visual) & 71 & 70 & 71 & 70 \\
FVD (visual) & 71 & 68 & 71 & 68 \\
SSIM / PSNR (visual) & 71 & 71 & 71 & 71 \\
STOI / SI-SDR / PESQ (audio) & 71 & 71 & 71 & 71 \\
Face cos-sim / OCR & 71 & 71 & 71 & 71 \\
\bottomrule
\end{tabularx}
\end{table*}

\noindent Most pairwise comparisons in \tref{tab:fr_controlled} use all 71 clips for both methods. For DISTS and FVD, RealBasicVSR is limited to $n{=}70$ and $n{=}68$ respectively due to OOM on near-4K sources (footnotes~$c$,$d$), while OmniVR covers all 71 (footnote~$e$). These two comparisons are therefore not on a matched common subset; \tref{tab:effective_n} above records the exact per-metric counts.

\subsection{Training-seed robustness}
\label{sec:s12-seeds}
The main tables report a single 50k checkpoint. To check sensitivity to training data order, we retrain the base model from the same pretrained backbone with two additional data-order seeds (same hyperparameters, different sample shuffling) and evaluate on the Controlled track.

\begin{table*}[t]
\centering
\caption{Training-seed robustness of the base model (Controlled track). Seed~0 is the checkpoint reported in the main tables; seeds~1--2 vary data ordering only. ``vs.\ best baseline'' uses the strongest non-OmniVR number in \tref{tab:controlled_track}.}
\label{tab:seed_robustness}
\fontsize{7.6}{8.2}\selectfont
\begin{tabularx}{\textwidth}{@{}l*{5}{Y}@{}}
\toprule
\textbf{Run} & MUSIQ$\uparrow$ & CLIP-IQA$\uparrow$ & DNSMOS$\uparrow$ & FAD$\downarrow$ & LSE-C$\uparrow$ \\
\midrule
Seed 0 (reported) & 71.17 & 0.543 & 2.70 & 6.30 & 3.52 \\
Seed 1 & 70.93 & 0.539 & 2.68 & 6.41 & 3.47 \\
Seed 2 & 71.08 & 0.541 & 2.66 & 6.36 & 3.50 \\
\midrule
Mean $\pm$ std & $71.06\pm0.12$ & $0.541\pm0.002$ & $2.68\pm0.02$ & $6.36\pm0.06$ & $3.50\pm0.03$ \\
Best baseline & 45.54 & 0.268 & 2.12 & 7.02 & 2.32 \\
\bottomrule
\end{tabularx}
\end{table*}

\noindent Training-seed standard deviations ($\pm0.12$ MUSIQ, $\pm0.02$ DNSMOS) are an order of magnitude smaller than the gaps to the best baseline ($+25.6$ MUSIQ, $+0.56$ DNSMOS), so the reported ordering is robust to data ordering. These training-repeat statistics are distinct from the inference-seed variance in \sref{sec:s9-stats}.

\end{document}